\RequirePackage{fix-cm}
\documentclass[smallextended]{svjour3}       % onecolumn (second format)
\smartqed  % flush right qed marks, e.g. at end of proof

\usepackage{cite}
\usepackage{graphicx}
\usepackage{amsmath}
\usepackage{stfloats}
\usepackage{url}

\usepackage[gray]{xcolor}
\usepackage{colortbl, booktabs}
\usepackage{enumitem}
\usepackage{tabularx}
\usepackage{hyperref}
\usepackage{multicol}

\usepackage{subcaption}

\usepackage[title]{appendix}

\usepackage[export]{adjustbox}

\usepackage{csquotes}

\begin{document}

%\title{\textcolor{blue}{A Comprehensive Study} on AI Algorithms to Implement Safety Using Communication
\title{\textcolor{blue}{A Comprehensive Study} on Artificial Intelligence Algorithms to Implement Safety Using Communication Technologies%\thanks{Grants or other notes
}
% \subtitle{Do you have a subtitle?\\ If so, write it here}


\author{Rafia Inam \and 
        Alberto Yukinobu Hata \and 
        Vlasjov Prifti \and
        Sara Abbaspour Asadollah
}


\institute{Rafia Inam \at
              Ericsson AB, Stockholm, Sweden \\
              \email{rafia.inam@ericsson.com}           %  \\
           \and
           Alberto Yukinobu Hata \at
              Ericsson Telecomunica\c{c}\~{o}es S/A, Indaiatuba, Brazil \\
              \email{alberto.hata@ericsson.com}
            \and
            Vlasjov Prifti \at
              M\"{a}lardalen University, V\"{a}ster\r{a}s, Sweden \\
              \email{vlasiov@info.al}
            \and
            Sara Abbaspour Asadollah \at
              M\"{a}lardalen University, V\"{a}ster\r{a}s, Sweden \\
              \email{sara.abbaspour@mdh.se}
}

\date{Received: date / Accepted: date}

\maketitle

\begin{abstract}
%The recent development of Artificial Intelligence (AI) has increased a lot the interest of researchers and practitioners towards applying its techniques into everyday life. Domains like automotive, health care and air space are benefiting from the use of AI. One of the most important aspects that technology aims to deliver is catering safety. Thus, the attempt to use AI techniques into carrying out safety issues is momentarily at a progressive state. Communication technologies have been around for many years as an essential part of society. Cellular and non-cellular communication are embracing the use of AI as well, sharing the same issues such as observing the environmental variations, learning or planning. Furthermore, multiple AI algorithms are trained in the cloud, where these communication technologies play a vital role.
%This article investigates the existing research performed on the use of AI to implement safety using different communication technologies. For this purpose, a systematic mapping study was conducted to summarize the recent publication trends and the current gaps in this field. The outcomes of this study contribute to the researchers by presenting new challenges for further extensive research and to the practitioners to find new methods or tools in order to apply them to the industry.
The recent development of artificial intelligence has increased the interest of researchers and practitioners towards applying its techniques into multiple domains like automotive, health care and air space to achieve automation. Combined to these applications, the attempt to use artificial intelligence techniques into carrying out safety issues is momentarily at a progressive state. As the artificial intelligence problems are getting even more complex, large processing power is demanded for safety-critical systems to fulfill real-time requirements. These challenges can be solved through edge or cloud computing, which makes the communication an integral part of the solution. 
% Safety   is   an   important   aspect   and   the   attempt   to   use   artificial intelligence techniques  into  carrying  out  safety  issues  is  momentarily  at a  progressive  state.  The generic use of artificial intelligence faces challenges of high processing devices to train the algorithms and if used in safety-critical systems there are further real-time requirements to execute them. These challenges make use of edge or cloud use and  consequently  the  communication  plays  an  integral  part  int he whole system
% We aim at providing a comprehensive picture of the state of the art solution focusing on the use of artificial intelligence to implement  safety  using  different  communication  technologies  in diverse  application  domains.
This study aims at providing a comprehensive picture of the state of the art artificial intelligence based safety solutions that uses different communication technologies in diverse application domains. To achieve this, a systematic mapping study is conducted and \textcolor{blue}{565 relevant papers} are shortlisted through a multi-stage selection process, which are then analyzed according to a systematically defined classification framework. The results of the study are based on these main objectives: to clarify current research gaps in the field, to identify the possibility of increased usage of cellular communication in multiple domains, to identify the mostly used artificial intelligence algorithms and to summarize the  emerging future research trends on the topic. The results demonstrate that \textcolor{blue}{automotive domain is the one applying artificial intelligence and communication the most to implement safety and the most used artificial intelligence in this domain is neural networks, clustering and computer vision; applying cellular communication to automotive domain is highest; the use of non-cellular communication technologies is dominant however a clear trend of a rapid increase in the use of cellular communication is observed specially from 2020 with the roll-out of 5G technology.}

%Our results demonstrate: 1. Automotive domain is the one applying artificial intelligence and communicating most to implement safety and the most used artificial intelligence in this domain is  computer vision and clustering and mostly non-cellular. Applying cellular communication to Automotive domain is highest; 2. Non-cellular communication is dominant, however we see opportunities of increased cellular communication with 5G; 3. we observe a clear trend of a rapid increase in the use of communication from 2011 mostly from academia, 5G is increasing from last 2 years; we see artificial intelligence techniques start increasing around the same time.

\keywords{Artificial Intelligence \and safety \and cellular communication \and 5G \and non-cellular communication \and systematic mapping study}

\end{abstract}

\section{Introduction}
\label{sec:introduction}

The advances in the \textit{Artificial Intelligence (AI)} field enabled different software systems to achieve higher levels of automation. This evolution moves towards a less dependence on human intervention for an intelligent system to make decisions~\cite{Alsamhi2019}. %, Wollschlaeger2017}. 
This also brings the possibility to have systems that interacts with humans in different manners. However, autonomy and interaction mechanisms must be designed in a way to ensure safety of the whole ecosystem. Consequently, \textit{safety} becomes a key component of AI-based autonomous systems~\cite{%Huang2020,
Everitt2018}. 
%and specially for those systems where humans are directly interacting with. 
Some examples of these systems include in which autonomous agents are involved that can operate in a heterogeneous environment, intermixed with humans, and perform a broad spectrum of tasks (for example general purpose robots~\cite{InamETFA18}, autonomous vehicles~\cite{Bila2017}, Internet-of-Things (IoT) in industry 4.0~\cite{Wollschlaeger2017}, and drones~\cite{Ullah2019})
%~\cite{FordBookRobots, Guiochet2017, Bila2017, InamETFA18}. 
These smart autonomous agents are equipped with sensors and AI to operate in an automated or semi-automated fashion.

%and cooperate with each other and humans in a safe, autonomous, and reliable manner.  
%and destine to bring a new industrial revolution. 4th industrial revolution that is based on autonomous  and smart Cyber Physical Systems (CPSs) [2]
%AI is an integral part of 

The smart autonomous agents are part of a big and complex smart \textit{Cyber Physical System (CPS)} in which they cooperate with each other and humans in a safe, autonomous, and reliable and trustworthy manner ~\cite{Huang2020}. These agents are used in a vast set of application domains ranging from automotive industry~\cite{Bila2017}, robotics~\cite{Guiochet2017}, %mining~\cite{}, 
air space using unmanned aerial vehicles (UAV)~\cite{Fellan2018}, agriculture~\cite{Jha2019}, healthcare~\cite{Kun-Hsing2018}, smart manufacturing~\cite{Wollschlaeger2017}, telecommunication~\cite{Cayamcela2018}, Internet-of-Things (IoT)~\cite{Islam2015}, and more.
%~\cite{Bila2017, Guiochet2017, Kun-Hsing2018, Jha2019, Cayamcela2018, Islam2015}. 
AI algorithms usually require efficient hardware processing capabilities to train and execute, which is not always possible to do on the agent or device which usually have limited resources. Additionally, for safety-critical systems there are mostly real-time requirements to execute the algorithms. As an example, \textcolor{blue}{the processing of large amount of sensor data generated by autonomous vehicle must be made in real-time to avoid any accident. Thus, there is a need for AI-based safe operations and, at the same time, for powerful processing hardware.} %Thus, the need is of not only AI-based safe operations, but the access to 
%vast knowledge base and 
%powerful processing hardware is also essential. 

% These challenges make the \textit{communication} of the agents with each other and with powerful remote machines \textcolor{red}{Rafia: remote machines or remote algorithms or remote functions?} (could be at edge and/or cloud infrastructures) an integral part of the whole system. 

\textcolor{blue}{All these challenges make the \textit{communication} to remote agents and remote processing infrastructures (e.g. edge or cloud computing) an integral part of the whole system.}
Based on the system’s demands, the communication needs could be very different. Two main types of communication technologies that play a significant role in such systems are \textit{cellular}~\cite{Ullah2019} and \textit{non-cellular communications}~\cite{Saleh2018}. %, Aceto2018}. 
These technologies must fulfil system requirements, otherwise, communication can introduce safety risks by not making the system able to respond in a precise and timely manner. 

%Because of these challenges,  these systems need communications between agents and processing units. are both safety and mission critical. Safety
%AI is present in diverse applications, ranging from autonomous vehicles to personal assistants. In these devices, access to a vast knowledge base and powerful processing hardware is essential. This is possible by communicating with remote servers, such as edge and cloud infrastructures. Therefore, communication technologies have an increasing importance in the automation process. 
%However, the communication technologies must fulfil established requirements, such as low latency and high throughput. Otherwise, communication can introduce safety risks by not making the system able to respond in a precise and timely manner. 

Notably, \textit{a strong correlation between AI, safety and communication} technologies exist and the progresses made in one field have a direct impact on the others~\cite{Sharma2020}. Traditionally, these fields were treated separately, but more recently a plenty of work covers the integration of AI using communication technologies to implement safety in different application domains. This is reflected in the publications made in recent years, which supports the convergence of these areas~\cite{sirohi2020convolutional}.
%{Ullah2019, Wollschlaeger2017, Fellan2018, sirohi2020convolutional}.

% Traditionally, these fields were treated separately, but more recently a plenty of works that covers the integration of AI with communication or safety have been proposed [REF]. 
%%% TODO: briefly explain systematic mapping study, present differences and advantages.
%There were efforts to track the evolution in these field, such as ..., however none of them covered them at the same time....
\textbf{ Paper Contributions:} The main goal of this study is to identify, classify, analyze and to identify gaps in the state-of-the-art research publications that utilizes AI algorithm(s) using communication to achieve safety in diverse domains. The study exclusively focuses on software aspects. To target this goal, a structured map of the available research literature is constructed by conducting a well-established methodology from Software Engineering research community called systematic  mapping~\cite{petersen2015guidelines}. %{kitchenham2007guidelines, petersen2008systematic}.
Using the systematic mapping process, we searched and selected relevant studies, defined a classification framework, applied the classification on the shortlisted studies in relation to various AI algorithms and communication technologies used in different domains to implement safety, and analysed and discussed the obtained results. 
%We also identified the  gaps  in  the  current  research  on  the use  of  AI  algorithms  and  communication  technologies in  different  application  domains;  and  to investigate in  different  domains  which  of  the  non-cellular  communication  technologies  can  potentially  be  replaced  by  the cellular ones to implement AI-based safety. 

The main contributions of this study are: 
\begin{itemize}
    \item A systematic review of current methods or techniques of AI and communication technologies applied in different domains to implement safety and to identify gaps in it;
 \item An investigation on the potential possibility of replacement of non-cellular communication technologies by the cellular ones to implement AI-based safety;
 \item A discussion of the current challenges and emerging future research trends on the topic, from both industry and academia perspectives, and to identify the interest of the industry on the topic;
 \item  Investigate the maturity of the performed research and the research approaches in general by considering the research type;
 \item Observe trends in the community’s
research focus by considering the research contribution for finding the increased focus on developing new methods, models or tools;
\end{itemize}

To the best of our knowledge, this paper presents the first systematic investigation into the state of the art on safety using AI and communication technologies in different domains.

\textbf{Paper outline:} The paper is organized as follows: \autoref{sec:related_works} presents an overview of related studies that covers AI for safety or communication; \autoref{sec:study_method} describes each step of the process of the systematic mapping used in the current study; \autoref{sec:classscheme} presents the classification schemes developed
in this study; \autoref{sec:results} analyses the mapping results and provides the plots generated in the mapping process; \textcolor{blue}{\autoref{sec:validity} describes the threats to the validity of this work; and finally \autoref{sec:conclusions} concludes this systematic mapping.}

\section{Related Works}
\label{sec:related_works}

% mention surveys in the communication+AI, AI+safety fields

There are many efforts in the literature to provide a comprehensive study that cover AI to implement safety.
%or to communication. Regarding AI to implement safety, there are different works that make an extensive review in this field.
% \textcolor{red}{I think we mention clustering, image processing also here. Alberto can you find something on these? Image processing is a technique ... Clustering is heavily used in automotive domain \cite{}}
Neural networks (NNs) and more specifically, deep learning (DL) is a technique that is widely adopted in a variety of safety related problems. 
More recently, Huang et al. \cite{Huang2020} presents the growing interest to train safe DL models for critical applications (e.g. autonomous vehicles and autonomous surgery). The work of Huang explores a variety of strategies to ensure model safety, such as interpretability, verification and testing. Tran et al. \cite{Tran2020} presents an overview of verification methods to ensure the correctness of AI in safety systems. 
\textcolor{blue}{Fayyad et al. \cite{Fayyad2020} presents a study on the use of deep learning based algorithms for perception, localization, and mapping used in autonomous vehicle systems in short-range or local vehicle environments.}
Reinforcement learning (RL) is an AI technique that is getting more popular in safety applications. For instance, in \cite{Garcia2015} is made an overview of studies that employ a particular RL method that incorporates safety, named safe RL, which basically investigates techniques to ensure safe exploration. 
The combination of computer vision and safety also brought the interest of researchers to elaborate survey works as in \cite{Al-Kaff2018}. Clustering is also another AI algorithm that is getting attention, specially inside vehicular communication problems \cite{Cooper2017}.
There are also works that make analysis of safety in a more general context of AI. To exemplify, \cite{Leslie2019} presents a comprehensive study of AI ethics and safety, and compiles papers related to this field. While \cite{Everitt2018} reunites studies that discusses the safety of Artificial General Intelligence (AGI) algorithms. It covers studies that bring ideas of how AGI can be designed in a safe manner.

Additionally, there are literature review studies that analyse safety in different application domains. %Robotics is one such field, specially used in industries 4.0. 
\cite{Guiochet2017} discusses different solutions proposed by the scientific community to guarantee safety when interacting with robots. Similarly, \cite{Bila2017} presents strategies to avoid accidents in intelligent transportation systems (ITS) when dealing with unforeseen situations. AI in healthcare is a topic that is covered in the review article of \cite{Kun-Hsing2018}, which brings the usage of different AI methods and data infrastructure for medicine related problems. \textcolor{blue}{\cite{Pouyan2020} examines the perceived benefits and risks of AI medical devices with clinical decision support features from consumers' perspectives using an online survey based data from 307 individuals, and reveals that technological, ethical (trust factors), and regulatory concerns significantly contribute to the perceived risks of using AI applications in healthcare.} Agriculture is yet another area that has large application of AI, specially in problems related to automation as raised in the review study of \cite{Jha2019}. This work also highlights the use of communication technologies in the agricultural automation, specially in those solutions that involve Internet of Things (IoT).

Besides not having survey works covering safety in communication domain, there are plenty of studies that explore AI in communication domain. \cite{Tong2019}  covers different AI solutions employed together with vehicle-to-everything (V2X) communication to leverage traffic safety and efficiency. While in \cite{Alsamhi2019} AI is targeted for robot communication to enable complex decisions in a efficient way. In the cognitive radio field, which deals with dynamic reconfiguration of telecommunication networks, there are surveys studies that discusses the application of different AI algorithms as in \cite{Bkassiny2013} and \cite{Abbas2015}. Self-organizing network (SON) is a topic related to automation of telecommunication tasks that have derived literature review studies investigating the usage of different AI techniques \cite{Klaine2017}. The recent advent of 5G also brings survey that involves the employment of AI \cite{Cayamcela2018}. 
\textcolor{blue}{\cite{sirohi2020convolutional} study the use of convolutional neural networks (CNN) and their variants for object detection, localization, and classification in modern 5G enabled ITS. }
Moreover, there are also studies that narrows down to specific machine learning techniques, mainly deep learning %\cite{Mao2018}  
\cite{Erpek2020} and reinforcement learning \cite{Luong2019}.

Another perspective explored by the studies is the relationship between the communication technologies and the different application domains. For example, there are survey studies that investigate different communication architectures and strategies used in the health care domain as presented in \cite{Islam2015} and \cite{Aceto2018}. There are also studies that make a wider analysis of a particular communication technology in different domains. In \cite{Ullah2019}, the impact of 5G communication is addressed in intelligent transportation, drones and healthcare. With the rise of Industry 4.0, the communication in manufacturing field also has been investigated in \cite{Wollschlaeger2017} and \cite{Marcon2017}. Other studies cover communication needs for specific machines, such as automated grounded vehicles (UAV) \cite{Fellan2018} and drones \cite{Sharma2020}.

The current study takes a different perspective from the previous works by providing analysis of studies that make use of AI and communication technology to provide safety in diverse domains. As shown in this section, these three key terms (AI, communication and safety) have been used separated or combined, but their simultaneous usage is neglected and is not covered so far to the best of our knowledge. As AI is getting more and more used in different fields, the safety of its application have a proportional importance. Likewise, communication and AI is even more related as both depends on each other either to provide an efficient communication or ensure the real-time execution through cloud computing.

\section{Study Method} 
\label{sec:study_method}

\begin{figure}[h]
    \centering
    \includegraphics[width=0.8\textwidth]{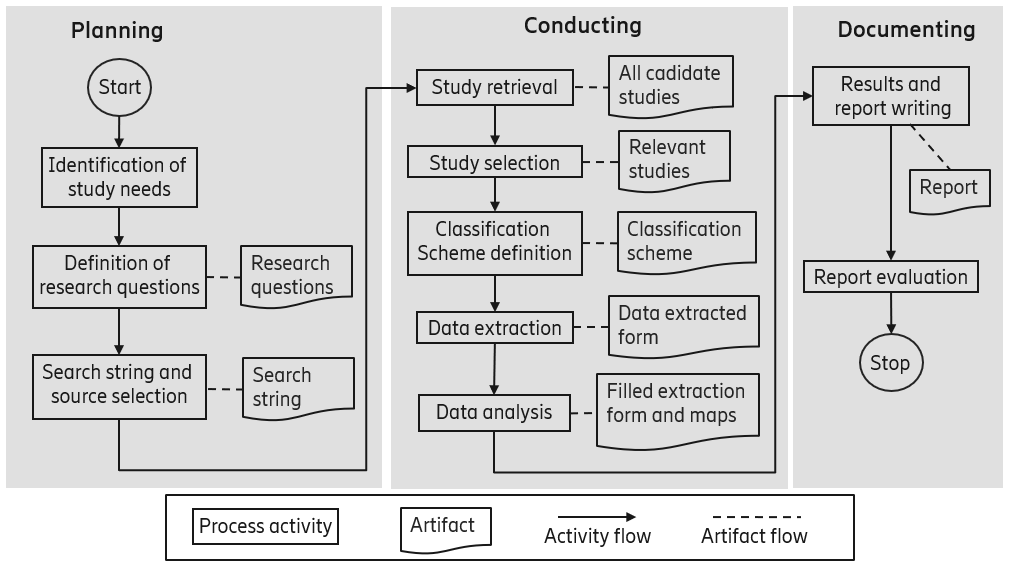}
    \caption{Overview of the research method process and its main activities. Diagram adapted from \cite{Bozhinoski2019}.}
    \label{fig:process}
\end{figure}

The systematic mapping (SM) method is employed in this study to understand how the field of AI-based safety enabled by communication technology is evolving. Essentially, SM performs an overview of research works to answer key research questions of this field. The SM analysis is also supported by visual maps that bring temporal and quantitative perspectives of different aspects of the study field. The SM process can be performed in different manners, like systematic literature review \cite{kitchenham2007guidelines}, or systematic mapping study \cite{petersen2008systematic}. This study follows the guidelines from Peterson et. al. for conducting systematic mapping studies, first presented in 2008 \cite{petersen2008systematic} and then improved in 2015 \cite{petersen2015guidelines} %, wohlin2012experimentation}  
and divide the overall process into three main classical phases for systematic mapping studies: planning, conducting, and documenting, and then further divide each phase into multiple steps as depicted in Figure \ref{fig:process}. %and described below.
%[8] B. A. Kitchenham, S. Charters, Guidelines for performing systematic literature reviews in software engineering, Tech. Rep. EBSE-2007-01, Keele University and University of Durham (2007).
%[9] C. Wohlin, P. Runeson, M. H¨ost, M. Ohlsson, B. Regnell, A. Wessl´en, Experimentation in Software Engineering, Computer Science, Springer, 2012.
%[38]. Petersen, K., Feldt, R., Mujtaba, S., Mattsson, M.: Systematic mapping studies in software engineering. In: 12th International Conference on Evaluation and Assessment in Software Engineering (EASE) 12, pp. 1{10 (2008)
%[39]. Petersen, K., Vakkalanka, S., Kuzniarz, L.: Guidelines for conducting systematic mapping studies in software engineering: An update. Information and Software Technology 64, 1{18 (2015)

%In this study we adopted the guidelines from [REF], which is comprised by five steps: Definition of research questions; Search string and source selection; Definition of study selection criteria; Data extraction; and data analysis. Each of these steps are detailed in the next sections.
\textbf{Planning: } The first phase in the study process and it aims at (i) demonstrating the need for performing a mapping study on the use of AI algorithms using communication in diverse domains to implement safety;
%on safety for MRSs; indeed, as discussed also in Section 9, secondary studies exist on topics related to robotics safety like mechanical and controller design [20] and human-robot interaction [21, 22, 23], but none of them takes into consideration safety from a software engineering point of view; 
(ii) defining the main research questions (see Section \ref{sec:rqs}); and (iii) defining the search string and selecting the scientific online digital libraries to perform the systematic mapping study (see Section \ref{sec:stringSource}).

\textbf{Conducting: } This phase consists of the following steps: 

\begin{enumerate}[label=(\roman*)]
\item \textit{Study retrieval}: The final search string is applied to the short-listed well-known academic search databases (see Section \ref{sec:studyretrieval}). The output of this activity is a comprehensive list of all candidate studies resulting from the search in the digital libraries.

\item \textit{Study selection}: Duplicates are removed from the candidate entries. Then entries are filtered first using the inclusion and exclusion criteria, and second using the Title-Abstract-Keywords (T-A-K) criteria (see Section \ref{sec:studySelection}) to obtain the final list of relevant studies to be considered in later steps of the study.

\item \textit{Classification scheme definition}: The next step in the systematic mapping study is to establish how the relevant studies are going to be classified. The classification scheme are designed to collect data for addressing the research questions \cite{wohlin2012experimentation} as formulated in the planning phase. The process of defining the classification scheme is described in Section~\ref{sec:classification}, and the obtained classifications are explained in Section~\ref{sec:classscheme}. %including the clusters for keywording method.

\item \textit{Data extraction}: Each selected relevant study is analysed and then information are extracted. For this purpose, a data extraction form is generated and filled with the extracted information in order to be analyzed during the next step. More details are presented in Section \ref{sec:data_extraction}.

\item \textit{Data analysis}: Comprises the last step of conducting phase. A detailed and comprehensive analysis of the extracted information (performed in the previous step) is made in the form of different maps (i.e. plotting charts such as bubble charts, pie charts and graphs) and an analysis is made, with the aim to address each research question of the study. The details about this activity are in Section~\ref{sec:dataAnalysis}.

\end{enumerate}

%  \textcolor{SaraColor}{Sara: I suggest, if we used "we did ...or ... for (i) and (iv), it would be better to keep the text more harmonize and apply it for (ii), (iii), and (v) as well!!}
 
\textbf{ Documenting: } The last phase in the study process aims at (i) generating a detailed analysis of the extracted information in the previous phase and thorough explanations of the obtained results as presented in Section~\ref{sec:results}, (ii) the analysis of possible threats to validity see Section~\ref{sec:validity}.

\subsection{Definition of Research Questions} 
\label{sec:rqs}

The main goal of this systematic mapping study is to observe and analyze the use of AI algorithms implementing safety using communication technologies in different application domains. As part of analyzing, we aim to investigate research gaps by highlighting topics where research results are lacking, and identify the potential for cellular communication adoption for industrial use. This goal is refined into the set of the following research questions (RQs):

\begin{itemize}[itemindent=1em]
    \item[\textbf{RQ1a}]: What types of AI algorithms are used to implement safety? 
    \item[\textbf{RQ1b}]: What kind of communication technologies are mostly used to implement safety? %modify graph
    \item[\textbf{RQ1c}]: Which are the main application domains where these AI algorithms find applicability in order to provide safety?
    \setlength{\itemindent}{-.7em}
    \item[--]\textbf{Objective}: This three-part question aims to obtain from the current state-of-the-art the types of AI algorithms, communication technologies and the domains where they are applied to implement safety.
    \setlength{\itemindent}{1em} \\
    \item[\textbf{RQ2a}]: What are the current research gaps in the use of AI to implement safety using communication technologies? %modify graphs 
%\textcolor{red}{  \item[\textbf{RQ2b}]: What are the current research gaps in the use of AI to implement safety using cellular and non-cellular communication technologies in the most popular application domains?  % Please note that we explain we chose only 5 popular application domains that are more than 5% in the previous graph. %%we develop 2 pie in bubble graphs one for cellular and second for non-cellular
%        \begin{itemize}
%            \item \textbf{Objective}: The aim of this research question is to identify the gaps in the current research area on the use of AI algorithms and communication technologies in different application domains.  
%        \end{itemize}   
%}
    % \item[\textbf{RQ2b}]: What is potential for using cellular communication to implement safety using AI in the most popular application domains?\textcolor{red}{OR}  \\
    \item[\textbf{RQ2b}]: Which application domains bring the opportunity to use cellular communication to implement AI-based safety?
    \setlength{\itemindent}{-.7em}
    \item[--]\textbf{Objective}: The aim of this two-fold research question is to (i) identify the gaps in the current research area on the use of AI algorithms and communication technologies in different application domains; and to (ii) investigate in different domains which of the non-cellular communication technologies can potentially be replaced by the cellular ones to implement AI-based safety.\\
    \setlength{\itemindent}{1em}
    \item[\textbf{RQ3a}]: What is the publication trend with respect of time in terms of cellular and non-cellular communications technologies? %modify graphs
    \item[\textbf{RQ3b}]: What is the publication trend with respect of time in terms of AI-based safety? 
    \setlength{\itemindent}{-.7em}
    \item[--]\textbf{Objective}: This two-part question aims to illustrate the current state of the topic and establishes a better understanding regarding the future trends.
    \setlength{\itemindent}{1em}\\
    \item[\textbf{RQ4}]: What type of research contributions are mainly presented in the studies?
    \setlength{\itemindent}{-.7em}
    \item[--]\textbf{Objective}: This research question aims to reflect new methods, tools or techniques, that might be further developed by researchers in this area. \\
    \setlength{\itemindent}{1em}
    \item[\textbf{RQ5}]: Which are the main research types being employed in the studies?
    \setlength{\itemindent}{-.7em}
    \item[--]\textbf{Objective}: This research question aims to understand the importance of the research approach and what it represents. 
    \setlength{\itemindent}{1em}\\
    \item[\textbf{RQ6}]: What is the distribution of publications in terms of academic and industrial affiliation?
    \setlength{\itemindent}{-.7em}
   \item[--] \textbf{Objective}: The goal is to know at what extent the industry is interested on the topic.
    \setlength{\itemindent}{1em}
\end{itemize}
% \noindent
% The answers to these research questions provide the outcomes of this master thesis discussed in section \ref{sec:results}. 
Research questions determine the main objectives of the systematic mapping study, and thus this step delineates how the remaining steps of the study process will be performed, specially the search process, the data extraction process, and the data analysis process.

\subsection{Search String and Source Selection} \label{sec:stringSource}
This section aims at the identification of the search string and the selection of database sources to apply the search in order to achieve a good coverage of existing research on the topic and have a manageable number of studies to be analysed as required to perform systematic review \cite{kitchenham2007guidelines} and sytematic mapping study \cite{petersen2015guidelines}. %[7,8]. 
%[7] K. Petersen, S. Vakkalanka, L. Kuzniarz, Guidelines for conducting systematic mapping studies in software engineering: An update, Information and Software Technology 64 (2015) 1–18.
%[8] B. A. Kitchenham, S. Charters, Guidelines for performing systematic literature reviews in software engineering, Tech. Rep. EBSE-2007-01, Keele University and University of Durham (2007).
The search string is a logical expression formed by a combination of keywords that is used to query the research databases. A relevant search string should return research works that address the study's RQs. The proper selection of the research database sources is also primordial for a good coverage, and for this, database sources related to both engineering and computer science were selected. The following sections detail the search string formulation and the database source selection:

\subsubsection{Search string}
\label{subsub:string}

In this work, the Population, Intervention, Comparison and Outcomes (PICO) method proposed by Kitchenham and Charters~\cite{kitchenham2007guidelines} is used to define the search string. PICO is based on the following set of criteria:
% In order to create the search string, the method called Population, Intervention, Comparison and Outcomes (PICO) proposed by Kitchenham and Charters in \cite{kitchenham2007guidelines} is applied. PICO criteria used for systematic mapping studies in software engineering is defined as follows:

\begin{itemize}
    \item \textbf{Population}: The category, application area or a specific role that covers this systematic mapping study.
    \item \textbf{Intervention}: Methodology, tool or technology that is applied in the study field.
    \item \textbf{Comparison}:  Methodology, tool or  technology that is used to compare in the study field.
    \item \textbf{Outcome}: The expected consequences (i.e. outcomes) of this study.
\end{itemize}

For this systematic mapping study, PICO was defined as follows: 

\begin{itemize}
    \item \textbf{Population}: AI-based safety algorithms.
    \item \textbf{Intervention}: Communication technologies.
    \item \textbf{Comparison}:  No empirical comparison is made, therefore not applicable.
    \item \textbf{Outcome}: A classification of the primary studies, which indicates the role of AI algorithms to implement safety using communication technologies.
\end{itemize}

Analysing the \textbf{Population} and \textbf{Intervention} categories, relevant keywords are identified: ``\textit{AI algorithm}'', ``\textit{safety}'' and ``\textit{communication technology}''. From that, the following sets are derived, which groups synonyms and similar terms of the keywords:

\begin{itemize}
    \item \textbf{Set 1}: Terms related to ``\textit{safety}'' field.
    \item \textbf{Set 2}: Terms related to \textbf{Intervention} are searched for.
    \item \textbf{Set 3}: Terms related to the classification of AI algorithms are searched for.
\end{itemize}

\begin{table}[t!]
    \centering
    \begin{tabular}{p{0.4\linewidth}|p{0.5\linewidth}}
        \toprule
        \textbf{Population}: AI-based safety algorithms &  (safety) AND (\enquote{artificial intelligen*} OR \enquote{Machine learning} OR \enquote{regression analysis} OR \enquote{supervis* learning} OR \enquote{unsupervis* learning} OR \enquote{clustering algorithm} OR \enquote{fuzzy logic} OR \enquote{image process*} OR \enquote{deep learning} OR \enquote{computer vision} OR \enquote{neural network} OR \enquote{augmented reality} OR \enquote{virtual reality} OR \enquote{speech recognition} OR \enquote{image recognition} OR ontology OR \enquote{state machine} OR \enquote{cognitive architecture} OR \enquote{language processing})\\
        \midrule
        \textbf{Intervention}: Cellular networks & (communicat* OR cellular OR 2G OR 3G OR 4G OR 5G OR \textcolor{blue}{LTE OR} GSM )  \\
        \bottomrule
    \end{tabular}
    \caption{Composition of search string.}
    \label{tab:tabela1}
\end{table}

Based on PICO and research questions, the following search string shown in \autoref{tab:tabela1} was composed. In this search string composition, logical operators (\textbf{AND} and \textbf{OR}), keywords, synonyms of the main keywords and terms related to the respective field, where the classification process will be based on, were taken into consideration to be used. It has to be noted that, the use of the search string may not be precisely the same for every digital library depending on their query syntax.

\begin{table}[t!]
\centering
\begin{tabularx}{0.49\textwidth} { 
  | >{\arraybackslash}X |}
 \hline
safety \textbf{AND} 
( \enquote{artificial intelligen*} OR \enquote{Machine learning} OR \enquote{regression analysis} OR \enquote{supervis* learning} OR \enquote{unsupervis* learning} OR \enquote{clustering algorithm} OR \enquote{fuzzy logic} OR \enquote{image process*} OR \enquote{deep learning} OR \enquote{computer vision} OR \enquote{neural network} OR \enquote{augmented reality} OR \enquote{virtual reality} OR \enquote{speech recognition} OR \enquote{image recognition} OR ontology OR \enquote{state machine} OR \enquote{cognitive architecture} OR \enquote{language processing} )
\textbf{AND} 
( communicat* OR cellular OR 2G OR 3G OR 4G OR 5G OR \textcolor{blue}{LTE OR} GSM )
  \\
 \hline
\end{tabularx}
\caption{The search string is formed by three logical expressions combined by \textbf{AND} operator.}
\label{tab:tabela2}
\end{table}

The final search string that is used to retrieve the studies is derived by the combination of \textbf{Population} and \textbf{Intervention} search strings obtained during the PICO process. The resulting string, presented in the \autoref{tab:tabela2}, is formed by the association of three logical expressions separated by \textbf{AND} operator. Each expression represents ``\textit{safety}'', ``\textit{AI}'' and ``\textit{communication}'' fields. 
% Derived from PICO method as explained in Section~\ref{subsub:string}, \autoref{tab:tabela2}  presents the final form of the search string. It consists of three main parts related by the logical \textbf{AND} with each other and logical \textbf{OR} within. 

\subsubsection{Source selection}
In order to find the existing relevant occurrences for this topic, four scientific online digital libraries were chosen: \textit{IEEE Xplore Digital Library}\footnote{IEEE Xplore Digital Library [Online]. Available: \url{https://ieeexplore.ieee.org/Xplore/home.jsp}},
\textit{ACM Digital Library}\footnote{ACM Digital Library [Online]. Available: \url{https://dl.acm.org/}},
\textit{Scopus}\footnote{Scopus [Online]. Available: \url{https://www.scopus.com/}}, and %\sara{and } 
\textit{Web of Science}\footnote{Web of Science [Online]. Available: \url{https://www.webofknowledge.com/}}. 
According to P. Brereton et al.~\cite{brereton2007lessons}, these libraries are known to be valuable in the field of software engineering when performing literature reviews or systematic mapping studies. Additionally, these libraries exhibit a high accessibility and support the export of search results to computationally manageable formats.

\subsection{Study Retrieval} \label{sec:studyretrieval}
%The final search string that will be used to retrieve the studies is derived by the combination of \textbf{Population} and \textbf{Intervention} search strings obtained during the PICO process. The resulting string, presented in the \autoref{tab:tabela2}, is formed by the association of three logical expressions separated by \textbf{AND} operator. Each expression represents ``safety'', ``AI'' and ``communication'' fields. 

%The search string is used to query the studies from the sources with the necessary adaptations in the syntax is made. \autoref{tab:tabela3} describes the adaptations made for each online digital library. The query resulted a total amount of 6473 primary studies, mostly from the Scopus and IEEE databases. In \autoref{tab:tabela4} the number of studies for each database is shown.
This is the first step in the conducting phase, in which the finalized search string is used to query the studies from the short-listed libraries with some small necessary adaptations in its syntax. \autoref{tab:tabela3} describes the adaptations made in the search string for each online digital library. The query resulted in a total amount of \textcolor{blue}{8760} potentially candidate studies, mostly from the Scopus and IEEE databases. \autoref{tab:tabela4} presents the respective number of studies for each database.
% This search string is used as the input in the online digital libraries to perform the search. Typically, a small derivation from the main search string, adapting it to the syntax of each library was needed. \autoref{tab:tabela3} describes the details on how the search string is applied on these digital libraries.

\begin{table}[t!]
    \centering
    \begin{tabular}{p{0.4\linewidth}|p{0.5\linewidth}}
        \toprule
        \textbf{Digital Library} &  \textbf{Description}  \\
        \midrule
        IEEE Xplore Digital Library & The search string is applied on text box of ``\textit{Command Search}" found in the ``\textit{Advanced Search}". \\
        \midrule
        ACM Digital Library & The search string is performed in the ``\textit{Advanced Search}". Since the number of occurrences was very large, the search was divided into three parts and applied within Title, Abstract and Author Keywords. In ``\textit{View Query Syntax}" $\rightarrow$ ``\textit{Edit Search}" the parts are connected with operator \textbf{AND} by default; it is changed to \textbf{OR}. \\
        \midrule
        Scopus &   In ``\textit{Advanced Search}" of Document Search, the search string is applied as three sub-strings.\\
        \midrule
        Web of Science &  In the text box of ``\textit{Advanced Search}" the string is applied. The string ``\textbf{TS=}"
        is added in the beginning of the string, where TS=Topic. \\
        \bottomrule
    \end{tabular}
    \caption{Query process of each digital library.}
    \label{tab:tabela3}
\end{table}

% In the end of this step, from performing the search in four different libraries, a considerable amount of primary studies, 6473, were found. In \autoref{tab:tabela4} the respective number of studies for every database is shown.
% \begin{table}[t!]
%     \centering
%     % \begin{tabular}{p{0.4\linewidth} p{0.3\linewidth}}
%     \begin{tabular}{l >{\color{blue}}c}
%         \toprule
%         \textbf{Digital Library}    & \textbf{Search Results}  \\
%         \midrule
%         ACM Digital Library         & 184  \\
%         Web of Science              & 959  \\
%         IEEE Xplore Digital Library & 2205 \\
%         Scopus                      & 3125 \\
%         \midrule
%         \textbf{Total}              & \textbf{8760} \\
%         \bottomrule
%     \end{tabular}
%     \caption{Amount of studies retrieved from each database and the total number of potentially candidate studies.}
%     \label{tab:tabela4}
% \end{table}
\begin{table}[t!]
    \centering
    % \begin{tabular}{p{0.4\linewidth} p{0.3\linewidth}}
    \begin{tabular}{l >{\color{blue}}c}
        \toprule
        \textbf{Digital Library}    & \textbf{Search Results}  \\
        \midrule
        ACM Digital Library         & 413  \\
        Web of Science              & 1361 \\
        Scopus                      & 2885 \\
        IEEE Xplore Digital Library & 4101 \\
        \midrule
        \textbf{Total}              & \textbf{8760} \\
        \bottomrule
    \end{tabular}
    \caption{Amount of studies retrieved from each database and the total number of potentially candidate studies.}
    \label{tab:tabela4}
\end{table}

\subsection{Study Selection} \label{sec:studySelection}
This step is used to determine the relevant studies matching the goal of the systematic mapping study. The automatic search results from digital library might have inconsistencies, due to presence of ambiguities and lack of details in the retrieved studies. Therefore, relevant studies must be extracted by following a selection process. The first step is the removal of duplicates as same studies can be retrieved from different sources. To identify these duplicates and remove them, the EndNote tool\footnote{EndNote [Online]. Available: \url{http://endnote.com}}, was applied. \textcolor{blue}{This step removed 2219 studies and leaving 8760 studies after this step.}

The selection process continues with the criteria definition and selection of relevant studies. Details are provided below.

% Performing the selection process happens to be the most challenging part of the systematic mapping study. After obtaining the total amount of studies from the automatic search, the selection towards finding the relevant studies begins with the removal of the duplicates. To identify these duplicates and remove them, considering a large number of the studies, a free software provided by the university was used. This tool called \textit{EndNote} \footnote{EndNote [Online]. Available: \url{http://endnote.com}} is widely used in research. By exporting the citations from each digital library to EndNote, the removal of the duplicates was performed. The number of studies being removed was 1669, leaving the process with 4804 studies after this step.\\[0.2cm]
% The selection process continues with the selection criteria. Inclusion and exclusion criteria as described in \ref{4.3} are applied over the remaining 4804 studies. 

\subsubsection{Criteria for Study Selection} % Inclusion and Exclusion criteria
The study selection criteria step is used to determine the relevant studies that match the goal of the systematic mapping study according to a set of well-defined inclusion (IC) and exclusion criteria (EC). 
% According to \cite{keele2007guidelines}, the selection criteria is one of the longest and time-consuming steps of the process. These criteria are divided into inclusion and exclusion. 
To classify a study as relevant, it should satisfy all inclusion criteria at once, and none of the exclusion ones. The applied inclusion and exclusion criteria are presented in the \autoref{tab:criteria}.

\begin{table}[h]
    \centering
    \begin{tabular}{c|p{0.85\linewidth}}
        \toprule
        \textbf{ID} & \textbf{Criteria} \\
        \midrule
        \textbf{IC 1} & Studies implementing safety. \\
        \textbf{IC 2} & Studies presenting AI algorithms in different application domains. \\
        \textbf{IC 3} & Studies proposing the use of any type of communication technology. \\
        \midrule
        \textbf{EC 1} & Studies that are duplicate of other studies. \\
        \textbf{EC 2} & Studies that are considered secondary studies to other ones. \\
        \textbf{EC 3} & Studies that are not peer-reviewed, short papers (4 pages and less). \\
        \textbf{EC 4} & Studies that are not written in English language. \\
        \textbf{EC 5} & Studies that are not available in full-text. \\
        \bottomrule
    \end{tabular}
    \caption{Inclusion (IC) and exclusion (EC) criteria.}
    \label{tab:criteria}
\end{table}

\subsubsection{Evaluation of Study Relevance}
%The criteria is used to determine whether a study is relevant or not. For that, T-A-K of each study is evaluated according to inclusion and exclusion criteria. The relevant studies were classified according to the following terminology: 
These inclusion and exclusion criteria were firstly applied on the Title-Abstract-Keywords (T-A-K) tactic of each study to determine its relevance to the RQs. The relevant studies were classified according to the following terminology: 
\begin{itemize}
    \item A study is marked as \textit{relevant} (\textbf{R}) if it meets all the inclusion criteria and none of the exclusion criteria.  
    \item A study is marked as \textit{not-relevant} (\textbf{NR}) if it does not fulfil one of the inclusion criteria or meets at least one of the exclusion criteria. 
    \item A study is marked as \textit{not-clear} (\textbf{NC}) if there are uncertainties from T-A-K analysis.
\end{itemize}

\begin{figure*}[h]
     \centering
     \includegraphics[cfbox=blue,width=\textwidth]{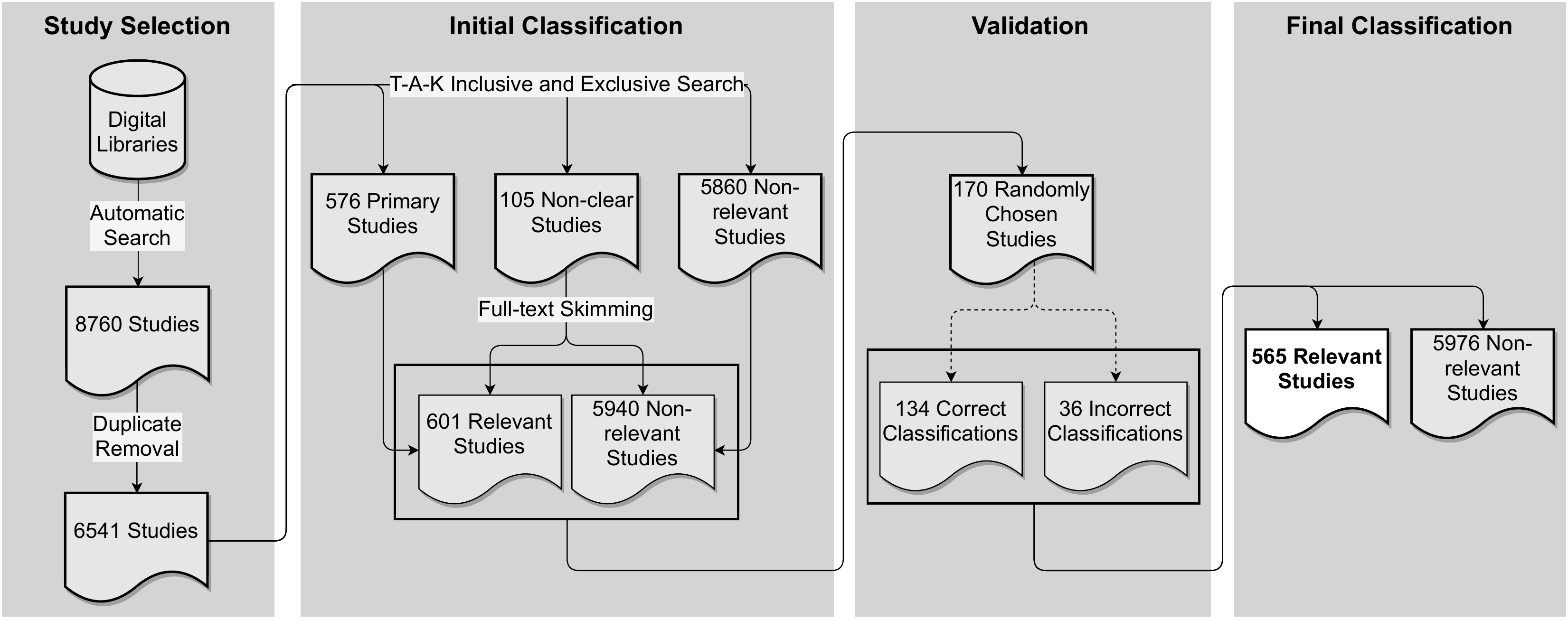}
     \caption{Workflow of the search and selection process. Diagram adapted from \cite{Bozhinoski2019}.} %\textcolor{red}{it shows that the random selection is applied to only relevant studies. 238+34 =272. NonRelevant numbers should be different}} %\textcolor{red}{Alberto please check is it the correct diagram?}} %\sara{Please add the number of Non relevant studies also we can have one more box for Non relevant studies for NC ones after full skimming OR we can remove the non-relevant studies box.}
     \label{fig:selection}
 \end{figure*}

If a study is labeled as \textbf{NC}, which could be due to the lack of information in T-A-K fields, the full-text (title, abstract, keywords, all sections and appendices, if any) skimming is conducted to decide about its inclusion in our set of relevant studies. 
% Three researchers have been involved during those phases. 
\textcolor{blue}{From a total of 105 \textbf{NC} studies, 25 studies were evaluated as \textbf{R} after full-text skimming, bringing the mapping study to 601 relevant \textbf{R} studies. To better visualize the selection process, \autoref{fig:selection} illustrates the detailed workflow of the whole process with the amount of studies obtained in each phase.}

%\textcolor{red}{Rafia: I ll add 1 para about reviewer info for data mapping}
% \textcolor{blue}{add second round of search and selection process here.}

One of the sub-steps of the selection process described in the guidelines is to perform snowballing sampling. According to \cite{jalali2012systematic}, snowballing is a method of identifying additional relevant studies after obtaining the primary ones. 
%There are two types of snowballing: forward snowballing, where the new studies are found in the citations of the paper, and backward snowballing that looks into the reference lists. 
Considering the slightly large number of papers obtained from the criteria described above, it was decided not to adopt snowballing in this systematic mapping study.

\subsection{Classification Schemes Definition} \label{sec:classification}

This step in the systematic mapping study establishes how the relevant studies are being classified to address the research questions presented in the previous section. For that, we first need to define a classification scheme, and for this purpose \textbf{six main facets} are added (which are associated with research questions). 
% In addition to the general publication metadata (e.g. title, abstract, keywords, author(s), author’s affiliation, publication year), some of which are dedicated to research questions and others for demographics purposes (see Section~\ref{sec:classscheme}). 

The first two of the main facets are \textit{type of research contribution} and \textit{research type} which are adopted unaltered from categorization presented in \cite{petersen2008systematic} and \cite{wieringa2006requirements}. 
The two used methods to develop more facets for our classification are \textit{initial categorization} and \textit{keywording}. The initial categorization is based on the discussions with experts from their respective fields and have an initial set of facets based on acquired knowledge from the experts. Few regular meetings were held in order to refine the categories further. The last four facets, \textit{AI algorithm}, \textit{cellular communication technology}, \textit{non-cellular communication technology} and \textit{application domain} were chosen after an initial categorization. 
%To develop the facet for our classification, we used two methods: \textit{initial categorization} and \textit{keywording}. The initial categorization is based on the discussions with experts from their respective fields and tried to have an initial set of facets based on acquired knowledge from the experts. Few regular meetings were held in order to refine the categories further. 

After that, a systematic \textit{keywording} method proposed in \cite{petersen2008systematic} is applied in order to refine the categories further. The keywording method is applied as follows:

\begin{itemize}
    \item \textit{Reading}: The abstracts and keywords of the selected relevant studies are read again. The intention is to look for a particular set of keywords representing the application domains, cellular and non-cellular communication types, and AI algorithms used to implement safety. Regarding \textit{application domain}, the keywording was mostly applied in the keywords of the studies rather than in the abstract. 
    \item \textit{Clustering}: Obtained keywords are grouped into clusters to have a set of representative clusters of keywords, which are used during the study classification.
\end{itemize}

% The clusters for \textit{keywording} method are found in Appendix \ref{keyy}.
% After performing the \textit{keywording} method %\sara{the \textit{keywording} method}
% the following classifications embodied below were obtained.

Both an initial categorization and keywording methods resulted in a set of classes (clusters) for each of the targeted facet. The obtained classes for each facet are described in Section \ref{sec:classscheme}.

\subsection{Data extraction}
\label{sec:data_extraction}

To extract the data from the selected relevant studies and to store it, a well-structured data extraction form is generated. This form is based on the classification scheme and the research questions. The data extraction form is shown in \autoref{tab:extraction}. It contains the \textit{data items} and their respective \textit{values} and the RQs which they are derived from.    
%Usually in systematic mapping studies, where two or more researchers are involved, the Cohen Kappa coefficient %{S. Keele, “Guidelines for performing systematic literature reviews in software engineering,” Technical report, EBSE Technical Report EBSE-2007-01, Tech. Rep., 2007.}  
%is used. This method is a good a practice to establish a strong agreement between the researchers. In this case this coefficient will not be used. \\
% For this step of the systematic mapping study process, Microsoft Excel is used to organize the data into spreadsheets. 

\begin{table*}[t!]
    \centering
    \begin{tabular}{l|llr}
        \toprule
        & \textbf{Data item}          & \textbf{Value}                              & \textbf{RQ}\\
        \midrule
        & Study ID                    & Number                                     & \\
        & Title                       & Name of the study                           & \\ 
        & Abstract                    & Abstract of the study                       & \\
        & Keywords                    & Set of keywords                             & \\
        \textbf{General} & Author       & Set of authors                            & \\
        & Author affiliation          & Type of affiliation                         & RQ6 \\
        & Pages                       & Number                                     & \\
        & Reference type              & Type of reference                           & \\
        & Year of publication         & Calendar year                               & RQ3a, RQ3b \\
        & Database                    & Name of the database                        & \\
        \midrule
        & AI algorithm                  & Type of AI algorithm            & RQ1a, RQ2a, RQ2b \\
        & Communication technology      & Type of comm. tech.       & RQ1b, RQ2a, RQ2b, RQ3a \\
        \textbf{Specific} & Application domains          & Type of application domain  & RQ1c \\
        & Research contribution       & Type of contribution                        & RQ4 \\
        & Research type               & Type of research                            & RQ5 \\
        \bottomrule
    \end{tabular}
    \caption{Data extraction form.}% \textcolor{red}{We need to divide communication into 2 rows: cellular and non-cellular. Sara.this table needs to double check(the RQ column) after we finalize the RQs.}}
    \label{tab:extraction}
\end{table*}

\begin{table}[t!]
    \centering
    \begin{tabular}{l c}
        \toprule
        \textbf{Classification}    & \textbf{Amount}  \\
        \midrule
        Correct         & 134 out of 170 (78.82 \%)  \\
        Incorrect       & 36 out of 170 (21.18 \%) \\
        \midrule
        \textbf{Total}              & \textbf{170} \\
        \bottomrule
    \end{tabular}
    \caption{\textcolor{blue}{Validation results from the 
    total studies. 170 (3.54 \%) studies were randomly reviewed from 6541 studies.}}
    \label{tab:validation}
\end{table}

For this step of the systematic mapping study process, Microsoft Excel spreadsheet was used to organize and store the extracted information from each relevant study for subsequent analysis. During this step, we found that some studies could be classified into more than one category according to the classification schema defined in the previous step. These studies were thus classified into all relevant categories. 

%\textcolor{red}{Sara please cross-ckeck this the following para in bold, please improve the language if possible}
%\textbf{In order to validate our data extraction process, 170 (from the total 4808) (3.54 \%) studies have been randomly selected and one researcher checked whether the results were consistent, independently from the researcher who performed the initial extracted data. 34 out of the randomly chosen 170 studies were found as non-Relevant which were incorrectly marked as Relevant. None of the non-relevant studies was found as relevant.}

\textcolor{blue}{In order to validate our data extraction process, 170 studies from  6541 studies (601 \textbf{R} + 5940 \textbf{NR}) of the initial classification have been randomly selected. One author checked whether the results were consistent, independently who performed the initial extracted data.} 
We  calculate  the  overall  agreement  over the  170  studies  as  80\% (\textcolor{blue}{36} %\textcolor{red}{Sara: Fig. 2 and table 7 show 36, which one is correct? 34 or 36?} 
studies out  of 170 studies were found as non-Relevant which were marked as Relevant by the other author).  This  means,  if  a  randomly selected  subject  is  rated  by  a  randomly  selected  rater  and then the process is repeated, there is 80\% chance to get the same rating decision the second time.

The output of this step was a filled data-extraction form consisting of all relevant studies that were categorized and validated in identified classes (see \autoref{tab:validation} for the total numbers).

% \textcolor{blue}{Alberto to provide results of validation: how much was modified during validation process, maybe the percentage of accuracy}

\subsection{Data analysis} \label{sec:dataAnalysis}
Data analysis is the last step of the systematic mapping study process. In this step, the map of the field is produced from the classified studies. Through the maps a comprehensive analysis of the studies is performed to finally answer the research questions presented in Section~\ref{sec:rqs}. Different methods to produce the map can be used, such as bubble charts, pie charts and line graphs. Produced maps and the derived analysis are presented in Section~\ref{sec:results}. The spreadsheet of the classified relevant studies can be found at \url{https://bit.ly/39cBNAy}.%\textcolor{blue}{We also upload the final version of classified relevant studies in form of an excel file at \textcolor{red}{https://......}.}

\section{Classification Scheme} 
\label{sec:classscheme}

This section presents the classification schemes developed in this study to address the research questions. 

\subsection{Artificial intelligence algorithm classification}
% Based on the initial categorization %\sara{here we need to explain to the readers what do we mean by initial categorization } 
% and the later \textit{keywording}, %\sara{\textit{keywording}},
% the classification was shaped as follows. For each {data item} of this facet a short description is presented .  
% \textcolor{red}{Link to appendix B}
Classes were obtained from an initial categorization and \textit{keywording} (see Appendix~\ref{sec:appendix_B}). For each {data item} of this facet a short description is presented as follows:

{\color{blue}
\begin{itemize}
    \item \textit{Clustering} - An unsupervised learning method that groups similar data without having the target features in the training data \cite{xu2008clustering}. Outputs might be less accurate compared to supervised methods due to the absence of expert knowledge input during the training. Examples of methods are K-means, DBSCAN and MeanShift.
    \item \textit{Cognitive Architecture} - Aims to use the research of cognitive psychology in order to create artificial computational system processes reasoning like humans \cite{lieto2018role}. Although complex behaviors can be modeled, the development of such architectures might require some effort. Soar, COGNET and CLARION are examples of cognitive architectures.
    \item \textit{Computer Vision} - It is a class of problem that relates to the process of automating the visual perception by including tasks from low-level vision as noise removal or edge sharpening to high-level vision by segmenting the images or interpreting the scene \cite{bezdek1999image}.
    This class includes primary studies in the field of image processing, virtual reality, or augmented reality, but did not specified the specific AI algorithm. Although computer vision can be implemented using different AI methods, neural networks (which includes deep learning) are popularly used.
    % \textcolor{red}{Good to mention that this is not exactly AI algorithm but it can be implemented using AI algorithm and thats why we think that it is importnt to keep this here.}
    \item \textit{Fuzzy logic system} - Solves a problem through a set of If-Then rules that takes into account the uncertainties in the inputs and outputs by imitating human decision-making \cite{dadios2012fuzzy}. As these rules are written in natural language, this method provides explainability of the decisions. However, it demands a manual process to model the rules and the models might have lower accuracy as compared to other AI algorithms.
    \item \textit{Natural Language Processing (NLP)} - It is a class of problems that studies mechanisms to make the computer program understand the human language. Most of the NLP solutions rely on deep learning or rule-based systems \cite{Socher2012}. This category include studies that mention the usage of NLP solution, but did not make clear about the specific AI algorithm used.
    % Aims at improving the ability of a computer program in understanding the human language. 
    \item \textit{Neural Network (NN)} - Refers to a set of algorithms designed to simulate the connection between the neurons \cite{ripley2007pattern}. This method has large popularity given the ability to solve complex problems in different fields. As a drawback, it demands NN architectures with thousands or millions of neurons, which makes necessary large dataset to train the model and lead to high inference times. Examples of NN include multilayer perceptron (MLP), Long short-term memory (LSTM) and Deep Learning (DL).
    \item \textit{Ontology} - Refers to a formal and structural way of representing concepts, relations or attributes in a domain \cite{fensel2001ontologies}. This requires a domain expert knowledge to design the ontology and the usage of a specific language.
    \item \textit{Optimization} - Class of an iterative procedures that aims at achieving ideally the optimal solution of a problem beginning from a guessed solution \cite{ruder2016overview}.  
    \item \textit{Regression Analysis} - Studies the relationship between a dependent and independent variable to models a function that makes predictions. The model is obtained by iteratively adjusting the function parameters to map a given set of input and outputs \cite{bishop2006pattern}. Usually it is applied to problems that are linearly separable.
    \item \textit{Reinforcement Learning (RL)} - Class of methods that determines optimal actions that maximizes the reward of an agent that is continuously exploring the state space and taking observations \cite{Sutton2018}. It is widely applied in problems that involves automating the decision taken by an intelligent agent. Q-learning and its DNN variant, DQN, are examples of RL methods.
    \item \textit{Finite State Machine (FSM)} - Represents a computational model that performs predefined actions depending on a series of events \cite{gladyshev2004finite}. It requires knowledge from the domain expert to model the states and the transitions between them. Commonly, it is used in tasks that involves simple decisions. Hierarchical FSM and fuzzy state machine are examples of FSM algorithms.
    \item \textit{Non-specified} - Relates to primary studies that mention the use of AI or ML, but do not elaborate into a particular technique.   
    \item \textit{Other} - Relates to primary studies that use a AI technique that is not part of the above-mentioned ones. This includes random forests and gradient boosting.
\end{itemize}
}

\subsection{Communication technology classification}
\label{sec:comm_tec_class}
% \textcolor{red}{Link to appendix B}
Based on the initial categorization for communication and the process of keywording, the following classes were extracted (see Appendix~\ref{sec:appendix_B}): 
% \textcolor{red}{do we need references in this section ? Maybe a ref of a book on communication technologies}

\begin{multicols}{2}
    \begin{itemize}
        \item \textit{Cellular  
        \item Non-cellular 
       % \item Satellite
        \item Not-mentioned}
    \end{itemize}
\end{multicols}

To address the first and second research questions (RQ1, RQ2), cellular and non-cellular communication technologies are further divided into subcategories. This is necessary to investigate which of the communication technologies are mostly used to implement safety in different application domains and which of the non-cellular communication technologies can be potentially replaced with the cellular ones \cite{Naik2020}: 

\subsubsection{Cellular communication technology classification}

\begin{itemize}
    \item \textit{2G} - Second generation of cellular communication technology that incorporated digital speech and medium-rate data transfer. GSM, GPRS and EDGE are also other terms used for 2G.
    \item \textit{3G} - The third generation brought higher speed data transfer and improved quality in the services compared to 2G. WCDMA is also referred to 3G technology.
    \item \textit{4G} - A higher data transfer rate leveraged new use-cases such as IoT and industry automation. LTE is another term usually associated to 4G.
    \item \textit{5G} - Low latency communication with high data rate made possible to expand the use-cases to diverse domains.
    \item \textit{Cellular (not-specified)}: Category used to group studies that do not mention which cellular technology used in the T-A-K fields.
\end{itemize}

\subsubsection{Non-Cellular communication technology classification}

\begin{itemize}
    \item \textit{WiFi} - Includes communication technologies based on IEEE 802.11 standard, such as vehicular ad hoc networks (VANET) and dedicated short range communication (DSRC).
    \item \textit{Radio-frequency identification (RFID)}: A short distance communication technology that uses electromagnetic field to transfer data. \textcolor{blue}{This class also includes Near Field Communication (NFC), which is a technology derived from RFID.}
    \item \textit{Bluetooth} - Communication technology intended for low energy consumption, short range communication and low data transfer rate.
    \item \textit{Satellite} - Long range communication technology, however sensitive to noises and dependent on line of sight.  
    \textcolor{blue}{\item \textit{Zigbee} - Radio-based communication that has lower cost and energy consumption compared to WiFi and Bluetooth. Although it can achieve similar coverage as these technologies, it supports a much lower bandwidth.}
    \item \textit{Non-cellular (not-specified)} - Comprises studies that do not specify the employed non-cellular technology in the T-A-K fields. 
\end{itemize}

\subsection{Application domain classification}
\label{sec:application_domain}

From keywording, the following classification scheme was derived for the domains where safety is provided by applying AI:

%\begin{multicols}{2}
    \begin{itemize}
        % \item \textit{Agriculture} - Covers agricultural automation for cropping fields.
        \item \textit{Air space} - Includes unmanned aerial vehicles (UAV) and airport management systems.
        % \textcolor{red}{we need to explain three of them. others are self-explanatory}
        \item \textit{Automotive} - Encompasses intelligent transportation systems, with predominance of advanced driver-assistance systems (ADAS), autonomous vehicles and truck platooning.
        \item \textit{Construction} - Monitoring of building infrastructure and monitoring of construction workers.
        \item \textit{Education} - Includes personal training and academic learning applications.
        \item \textit{Factory} - Mainly covers warehouses, chemical industries and nuclear plants.
        \item \textit{Health care} - Referees to automation of patient care, remote surgeries \textcolor{blue}{and monitoring of elderly people}.
        \item \textit{Marine} - Underwater applications, unmanned surface vehicles (USVs) \textcolor{blue}{and vessels}.
        \item \textit{Mining} - Covers monitoring of mining workers and autonomous underground machinery.
        \item \textit{Robotics} - Comprises utilization of robots for process automation.
        \item \textit{Surveillance} - Environment monitoring for security \textcolor{blue}{and dangerous areas}.
        \item \textit{Telecommunication} - Applications in the communication domain, such as enhancements in the reliability and traffic capacity.
        \item \textit{Other} - Specifies studies that do not fit in the aforementioned application domains. 
    \end{itemize}
%\end{multicols}

\subsection{Research contribution classification}
\label{sec:research_contribution}

By defining a classification framework for \textit{types of research contribution}, new methods, techniques or tools can be developed in the future by the researchers to implement safety. In order to categorize this facet, the classification scheme presented and defined by Petersen et al. in \cite{petersen2008systematic} is used. This scheme is highly adopted by researchers when conducting a systematic mapping study~\cite{asadollah201710}. This facet was formed with the following \textit{data items}:

\begin{itemize}
  \item  \textit{Model} - Refers to studies that present information and abstractions to be used in AI-based safety.    
  \item  \textit{Method} - Refers to general concepts and detailed working procedures that address specific concerns about implementing AI-based safety.  
  \item  \textit{Metric} - Refers to specific  measurements and metrics to assess certain properties of AI-based safety.
  \item  \textit{Tool} - Refers to any kind of tool or prototype that can be employed in the current models.
  \item  \textit{Open Item} - Refers to the remaining studies that do not belong to any of the above mentioned categories.
\end{itemize}

\subsection{Research types classification}
\label{sec:research_type}

% According to \cite{asadollah201710}, the \textit{research types} classification is chosen, because of the importance on understanding the research approach and the value it represents. 
To categorize the \textit{research types}, the classification scheme proposed by Wieringa et al. in \cite{wieringa2006requirements} was applied, which includes the following \textit{data items}: 

\begin{itemize}
  \item  \textit{Validation Research} - Investigates the novel techniques, that are yet to be implemented in the practice. Examples of investigations include experiments, prototyping and simulations.
  \item  \textit{Evaluation Research} - Evaluates an implemented solution in practice considering the benefits and drawbacks of this solution. It includes case studies, field experiments and related studies.    
  \item  \textit{Solution Proposal} - Refers to a novel solution for a problem or a significant extension to an existing one.
  \item  \textit{Conceptual Proposal} - Gives a new approach at looking topics by structuring them in conceptual frameworks or taxonomy.
  \item  \textit{Experience Paper} - Reflects the experience of an author explaining at what extent something is performed in practice.  
  \item  \textit{Opinion Paper} - Refers to studies that mostly express the opinion of the authors on certain methods.  
\end{itemize}

\section{Results of the field map and discussion}
\label{sec:results}

This section presents the analysis made on the relevant studies, which is organized as follows:

% In this section of the thesis, the results obtained from analyzing the primary studies are presented. %The extensive list of the primary studies is found in Appendix \ref{AppendA}. 
% A better understating on how this section is assembled is shown below: 

\begin{itemize}
    %\item Each subsection coincides with a research question taken into account to be answered.
    \item Each subsection covers one or a set of research questions from Section~\ref{sec:rqs}.
    \item Outcomes of the research questions are presented in chart format.
    %\item For each plotted chart or graph the related discussion is described accordingly.
    \item An analysis supported by a chart is provided to answer the research question.
\end{itemize}

\subsection{Results of \textbf{Research Question 1(a-c)}}

%\sara{Vlasiov, please do not forget to have a big table with all 272 papers. Similar to table 1 in "10 Years of research on debugging concurrent and multicore software: a systematic mapping study" paper or TABLE II in "Management of Service Level Agreements for
%Cloud Services in IoT: A Systematic Mapping Study" }

The first research question is a three-part question. It aims at making an analysis towards AI algorithms, communication technologies and application domains. %For each sub-question, a single figure is plotted to present the amount of the primary studies discussing these classes.

\subsubsection*{\textbf{RQ1a.} What types of AI algorithm are used to implement safety?} %\sara{It is better to copy-past the RQs, otherwise you need another section ("discussion") with RQs and the answers one by one.}}
% \hfill

\begin{figure}[h]
    \centering
    \includegraphics[
 %   trim={0.5cm 0.8cm 0 1.2cm},clip,
 width=0.85\textwidth, cfbox=blue]
    {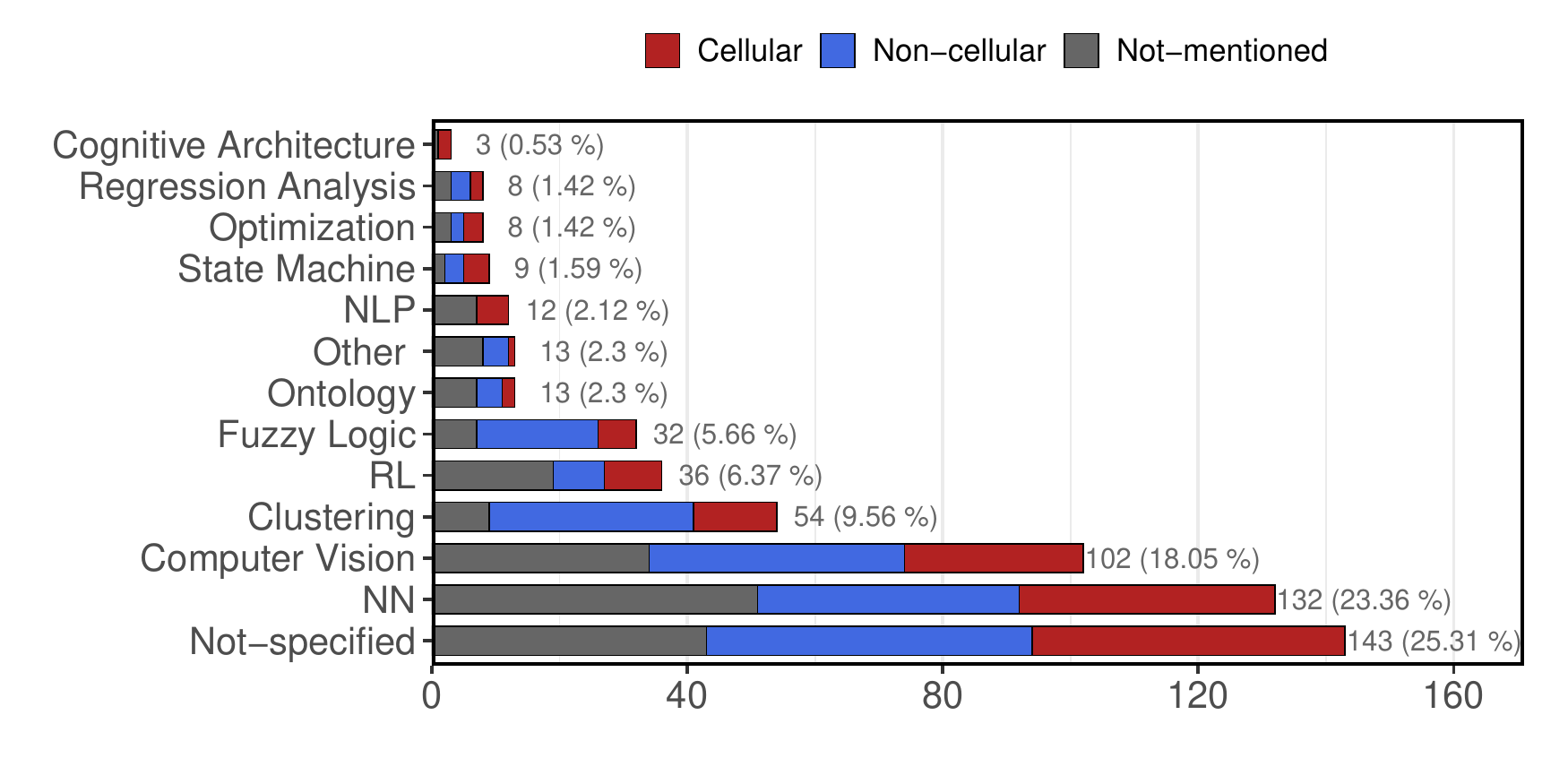}
    \caption{Distribution of studies according to AI method.}
    %\sara{Please change the label to percentage and not the number.}
    %\sara{I suggest to change the position of "Other" to the end of the graph(the last right place)}
   
    \label{fig:RQ1a}
\end{figure}

The AI algorithms used to implement the safety was summarized in a bar chart presented in \autoref{fig:RQ1a}. \textcolor{blue}{From this chart, we analyze that neural networks was the largest adopted method, with 132 studies (23.36\%). Alongside, computer vision and clustering corresponds to 18.05\% and 9.56\% of studies, respectively. Cognitive architecture was the least used algorithms with only 3 mentions. However, most of the studies (25.31\%) did not specify the employed AI method. A detailed classification of the studies in terms of AI algorithm can be checked in \autoref{tab:ai_class} of Appendix~\ref{sec:appendix_B}.}

\textcolor{blue}{\textbf{Discussion: } The larger use of NN, computer vision and clustering as compared to the other AI techniques is obvious from the results. It is mainly due to their use in the field of automotive domains as discussed in \textit{RQ1c}. }

Overall, a slight predominance of the non-cellular communication can be seen. This reveals that so-far the cellular technology is not the first option to be used together with the AI-based safety problems. However, when analysed the AI-algorithms individually, in some cases there are a larger adoption of cellular communication especially in reinforcement learning, NLP, state machine, optimization and cognitive architectures. This demonstrates that cellular communication has large opportunities to be used as will be discussed in RQ2.

% From this chart, we analyze that computer vision was the largest adopted method, with 56 studies (23.73\%). Alongside, clustering and neural network corresponds to 14.41\% and 12.71\% of studies, respectively. Cognitive architecture was the least used algorithms with only 2 mentions. However, most of the studies (25.42\%) did not specify the employed AI method. A detailed classification of the studies in terms of AI algorithm can be checked in \autoref{tab:ai_class} of Appendix~\ref{sec:appendix_B}.

% It also can be seen a predominance of non-cellular communication in which the majority of the AI algorithms are applied. This suggests that so-far the cellular technology is not the first option to be used together with the AI-based safety problems. However, the presence of cellular communication is still notable in most of the algorithms. This demonstrates that cellular communication has large opportunities as will be discussed in RQ2. 

% In \autoref{fig:RQ1a}, the number of studies using AI to provide safety is illustrated. This bar chart shows that the most discussed algorithm with 63 studies is \textit{computer vision}. However, this cannot be declared for sure, because almost 26\% of the studies do not clearly specify the type of algorithm that they use. Furthermore a lot of studies present \textit{clustering} and \textit{neural networks} with 14.4\% and 12.9\%, respectively. On the other hand, cognitive architecture and optimization are the least used algorithms with only 2 mentions each.       

\subsubsection*{\textbf{RQ1b.} What kind of communication technologies are mostly used to implement safety?} %\sara{I think this graph only shows "What kind of communication technologies are mostly used to implement safety". To show the data in different application domains, you can have another RQ which the answer can present with a table with 2 columns. The first column contains different domains (each domain in 1 row) and the second column contains a similar pie chart for each domain.}
% \hfill

\begin{figure}[h]
    \centering
    \includegraphics[trim={0.5cm 1.2cm 2.5cm 1.5cm},clip,width=1.0\textwidth, cfbox=blue]{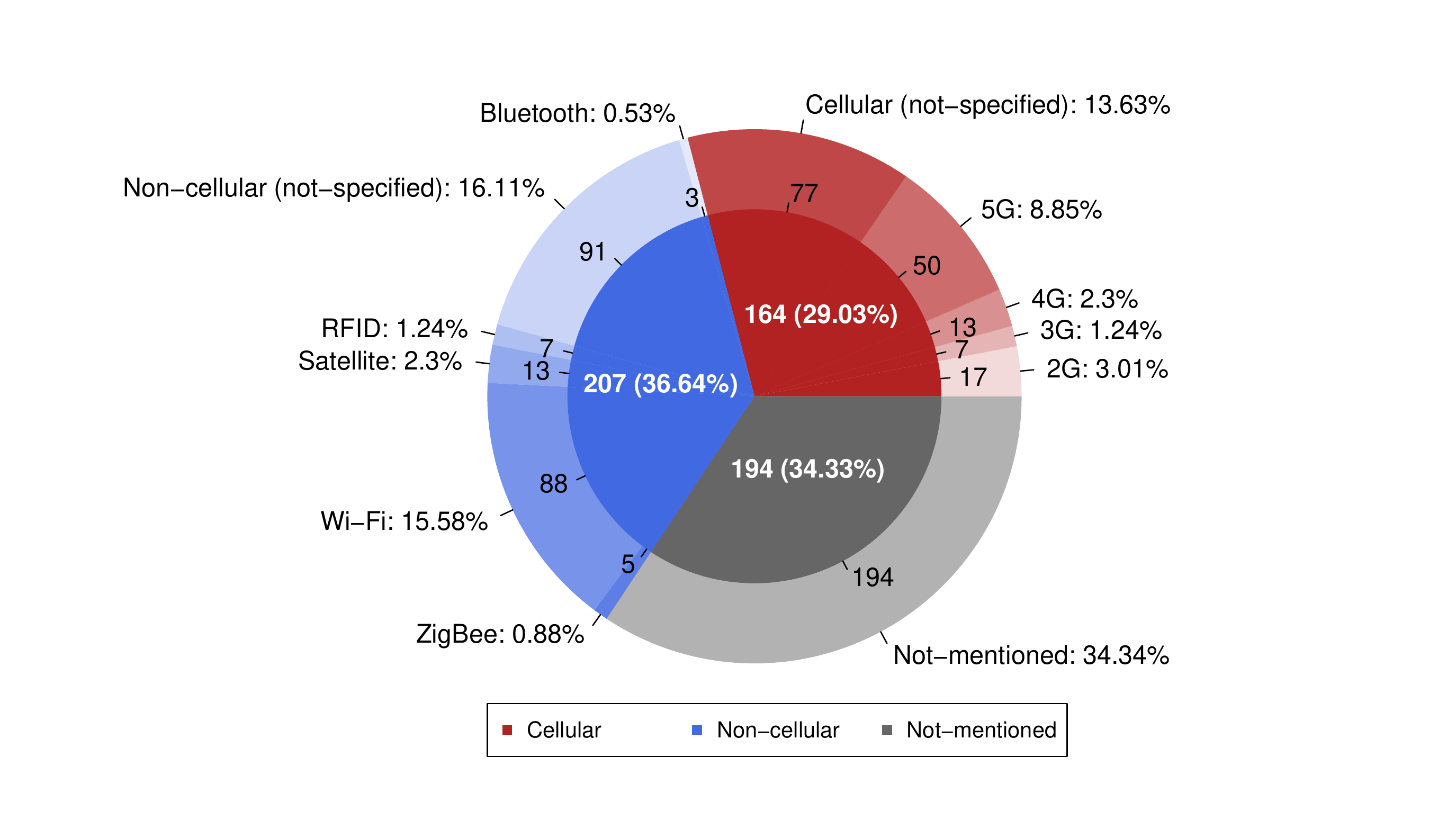}
    \caption{Distribution of studies for communication technologies.} %\textcolor{red}{we modify this graph with cellular, non-cellular, satellite, not-mentioned. Then we make 2 new graphs for cellular and non-cellular}}
    \label{fig:RQ1b}
\end{figure}

This research question depicts the analysis of the publications with respect to the communication technologies. In \autoref{sec:comm_tec_class}, a classification scheme focusing on communication technologies was discussed, which includes 2G, 3G, 4G, 5G, cellular (not-specified), non-cellular, satellite and not-mentioned. 
The pie chart in \autoref{fig:RQ1b} presents the outcomes for each category addressing the number of relevant primary studies and their respective percentages. \textcolor{blue}{From the results, it can be inferred that the dominant technology with 207 studies (36.64\%) is the non-cellular communication. From that, WiFi is the most employed non-cellular technology, with 88 studies (15.58\%).}

\textcolor{blue}{On the other hand, cellular communication is has slightly lower mentions, resulting in a total of 164 (29.03\%) publications. In this category, 5G is the most used communication with 8.85\%. Behind that, 2G and 4G technologies appear with 3.01\% and 2.3\%, respectively. 3G is the least addressed communication with 1.24\% of the studies. There are also 194 papers (34.33\%) that do not specify the communication technology used.}

\textbf{Discussion: }We believe there could be three reasons for a lower usage of cellular technologies as compared to the non-cellular ones: First, the usage of non-cellular technologies is easy for research purposes, specially in academic sector due to its non-propriety nature; second, cellular technology's focus was initially limited to provide connectivity to only cellphones. Another reason might be the higher bandwidth and lower latency of the WiFi compared to most of the cellular technologies (i.e. 2G -- 4G).
Last but not the least, due to the propriety nature of cellular technologies, most of the research done in industry is published  as patents or standards, which are not part of this study. 

\textcolor{blue}{Among the cellular technologies, 5G is the most used one, although it is the newest among all the network generations. We think the reason is that 5G supports the connectivity of industry verticals through its network slicing concept \cite{InamSOCNE2015}, especially for automotive domain \cite{InamETFA2016}, which was missing in its predecessor network generations. And this enables a big number of new
%Now with the advent of 5G, its 
use cases which are extended to a diversity of domains, resulting in its larger adoption as compared to the previous generations of mobile communication technologies}. 

% From the results, it can be inferred that the dominant technology with 119 studies, with half of the total primary publications (50\%) are the non-cellular communication. From that, WiFi is the most employed non-cellular technology, with 32 studies (13.56\%). 
% On the other hand, cellular communication is mentioned slightly more than a third of studies with a total of 81 publications. The GSM technology in 2G and the recent boost of 5G are among the most used cellular communication with 5.08\% and 5.93\%, respectively. \autoref{fig:RQ1b} shows that the least addressed communication category is 3G with almost 2.97\% of the studies. There are also 38 (16.1 \%) studies that do not specify the communication technology used.

% We believe there could be three reasons for a lower usage of cellular technologies as compared to the non-cellular ones: First, the usage of non-cellular technologies is easy for research purposes, specially in academic sector due to its non-propriety nature; second, cellular technology's focus was initially limited to provide connectivity on cellphones. Now with the advent of 5G, its use case was extended also and we observe an increase in its use as well (5.93\%). The last reason might be the higher bandwidth and lower latency of the WiFi compared to most of the cellular technologies (i.e. 2G -- 4G).
The communication technology classification of each relevant study can be found in \autoref{tab:comm_class} of Appendix~\ref{sec:appendix_B}.

\subsubsection*{\textbf{RQ1c.} Which are the main application domains where these AI algorithms find applicability in order to provide safety?}
\label{subsec:RQ1c}
% \hfill

\begin{figure}[h]
    \centering
    \includegraphics[trim={0.5cm 0.5cm 0 0},clip,width=0.85\textwidth, cfbox=blue]{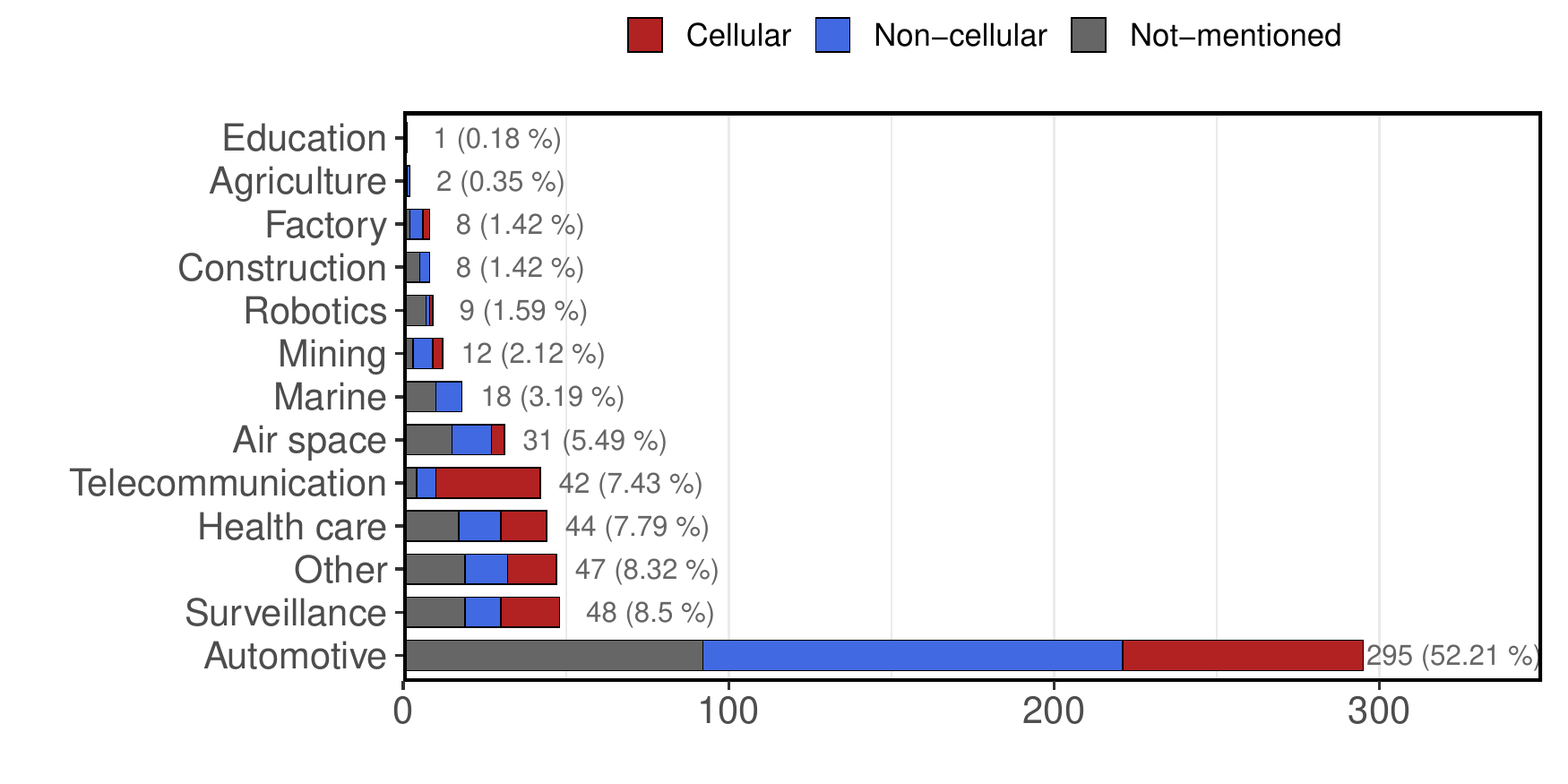}
    \caption{Distribution of studies for application domains.} %\sara{Please change the label to percentage and not the number.} 
    %\sara{I suggest to change the position of "Other" to the end/top of the graph(the last top/bottom place)}
    
    \label{fig:RQ1c}
\end{figure}

\autoref{fig:RQ1c} depicts a bar chart presenting the number of publications and the percentage of the pool of the relevant studies addressing the application domains in which the AI algorithms are applied. From the chart, an unequal distribution of the publications can be seen. The majority of the publications focus on the \textit{Automotive} domain, mostly within the field of Intelligent Transportation Systems (ITS) with \textcolor{blue}{52.21\% of studies (295 publications). 
The second most discussed domain is \textit{Surveillance}, which is mostly dominated with the use of cellular technology. \textit{Health care} and \textit{telecommunication} comes after, with 7.79\% (44 publications) and 7.43\% (42 publications). In both domains, cellular communication has a larger presence. The application domains with the least recognition are \textit{Education} and \textit{Agriculture} with 0.42\% (1 publication) and 0.85\% (2 publications), respectively.}

\textbf{Discussion: }We believe that the reason for   \textit{Automotive} being the most used domain is due to the inherent mobile nature of the autonomous vehicles and its need for a reliable cellular communication to connect to the other infrastructures (e.g. an intelligent traffic light) and to access services residing in the edge or cloud. Although cellular communication promotes higher mobility, non-cellular technology is predominant in this scenario.

\textcolor{blue}{The analysis of the communication aspect indicates that most of the non-cellular technology is used in the automotive domain. This reveals a gap and a potential opportunity to increase cellular technology usage in the automotive related problems as will be discussed in more detail in \autoref{subsec:RQ2}.}

% \autoref{fig:RQ1c} depicts a bar chart presenting the number of publications and the percentage of the pool of the relevant studies addressing the application domains in which the AI algorithms are applied. From the chart, an unequal distribution of the publications can be seen. The majority of the publications focus on the \textit{Automotive} domain, mostly within the field of Intelligent Transportation Systems (ITS) with 58.47\% of studies (140 publications). We believe that this is due to the inherent mobile nature of the autonomous vehicles moving on the roads and its need for a reliable cellular communication in order 1) to connect to the other infrastructure (like an intelligent traffic light etc.) and 2) to access services residing in the edge or cloud.
% Even though it was expected that the most discussed domain in the studies to be \textit{Automotive}, considering that in the application domain classification there are 12 more different classes, this number is quite high and the gap with the other domains as well. 

\textcolor{blue}{The reason for the least used \textit{Agriculture} and \textit{Education} domains does not mean that AI and communication are ignored, but that safety is not currently a key required aspect in these domains. The complete classification of the studies according to the domain is presented in \autoref{tab:dom_class} of Appendix~\ref{sec:appendix_B}.}

% The second most discussed domain is \textit{Telecommunication}, which is mostly dominated with the use of cellular technology, and \textit{Air Space}, both with 8.05\% (19 publications).  \textit{Health care}, which corresponds to 7.20\% of the studies (17 publications), is another domain that grabs the attention. This means that AI is getting a lot of attention even in medicine field. The application domains with the least recognition are \textit{Education} and \textit{Construction} with 0.42\% (1 publication) and 0.85\% (2 publications), respectively. The complete classification of the studies according to the domain is presented in \autoref{tab:dom_class} of Appendix~\ref{sec:appendix_B}.

% The analysis of the communication technology adopted in the domains indicates a large preference of non-cellular communication. Domains that cellular communication has prevailed are \textit{Telecommunication} and \textit{Surveillance}. \textit{Automotive}, which is the most mentioned domain, is also the domain with the largest presence of cellular communication, though non-cellular having the largest contribution. This shows a potential increase of cellular technology usage exploring more automotive related problems as will be discussed in the results of RQ2 below.

\subsection{Results of \textbf{Research Question 2(a--b)}}
\label{subsec:RQ2}

This section presents the results of the RQ2, which is divided into two research questions. The RQ2 covers the relationship between AI-based safety and two classification schemes: communication technology and application domain.
% To better picture this relation, in \autoref{fig:alg_dom_comm} a bubble chart is presented in a 2-axis scatter pie plot. The horizontal axis represents  the AI algorithms, while the vertical axis represents the application domains. The scatter pie presents the distribution of communication technologies where the AI algorithm was employed to implement safety in an application domain.\\

\subsubsection*{\textbf{RQ2a.} What are the current research gaps in the use of AI to implement safety using communication technologies?}
% \hfill

\begin{figure}[!b]
    \centering
    \includegraphics[trim={0.8cm 0.9cm 0.1cm 0.4cm},clip,width=0.85\textwidth, cfbox=blue]{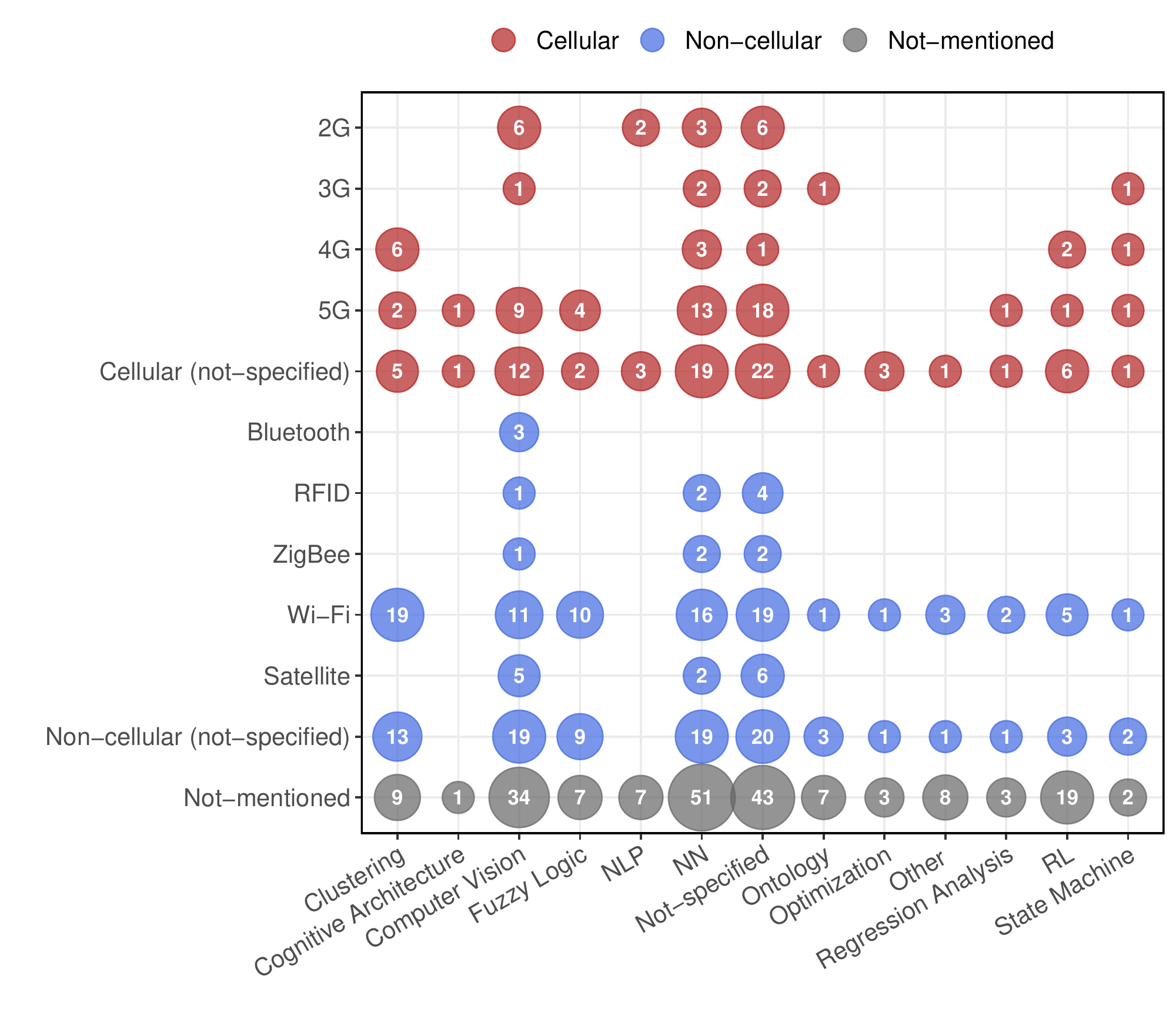}
    \caption{Bubble chart presenting the amount of publications according to the AI algorithm and the communication technology used to implement safety.}
    \label{fig:alg_com}
\end{figure}

This research question investigates the existing gaps and the future opportunities in the AI-based safety using communication technologies. The bubble chart on \autoref{fig:alg_com} presents the amount of publications for each pair of AI algorithm and communication technology. The bubble size is proportional to the number of publications. There are no bubbles for the pairs that have no publications. 

\textcolor{blue}{The chart shows that the gaps mostly belong to \textit{Cognitive Architecture}, \textit{Optimization}, \textit{Regression Analysis}, and \textit{State Machine} have the lowest amount of publications (less than 10). A possible explanation is that these AI algorithms have low applicability on problems that combine safety and communication.}
% The chart shows that the gaps mostly belong to  \textit{Cognitive Architecture}, \textit{Optimization}, \textit{Regression Analysis}, and \textit{NLP} have the lowest amount of publications (less than 5). All these publications are applied on cellular communication. \textit{Regression Analysis} also have few publications (4 in total), however having  presence in both cellular and non-cellular technologies. 
% Possible explanation is that although the above mentioned AI algorithms have been applied for different problems, they were not used explicitly to implement safety for those applications where the communication requirement was vital. 

%Recently, with the advent of new hardware-software technologies from last decade like edge- or cloud-based services and to access the services online, the AI-communication pair is made feasible to be utilized in real-time and now the researchers in this field are realizing this potential.}

\textcolor{blue}{Examining the largest bubbles (excluding the not-specified and not-mentioned ones), for the non-cellular communication, the pair ``\textit{Clustering}'' -- ``\textit{Wi-Fi}'' has 19 publications, while for the cellular communication, the pair ``\textit{NN}'' -- ``\textit{5G}'' has 13 publications. Neural networks also have a relatively large adoption in the non-cellular communication, which has 16 studies using WiFi. }

\textbf{Discussion: }This suggests that neural networks are widely used to implement safety using either WiFi or 5G. Still in neural networks, it is also noticed that the bubbles are spread over different communication technologies, which indicates high popularity of this AI algorithm. A similar scenario is present in \textit{Computer Vision}, which has a considerable presence on both WiFi (10 studies) and 5G (7 studies), and the bubbles are distributed on other communication technologies.

\textcolor{blue}{\textit{Reinforcement learning}, has lower numbers compared to the aforementioned methods, but as it is getting increasingly popular, there might be some opportunity to invest on this algorithm to implement safety using communication. It can also be noticed that besides \textit{cognitive architectures} and \textit{NLP} being popular in other areas, they are not commonly chosen in the safety field. From the communication perspective, it is observed that 5G and Wi-Fi are the common choice for most of the AI algorithms. Therefore, there is a large gap in the usage of other communication technologies, such as 4G and ZigBee together with the AI algorithms. }

\subsubsection*{\textbf{RQ2b.} What is potential for using cellular communication to implement safety using AI in the most popular application domains?}
% \hfill

%\textcolor{red}{highlight use of cellular communication in other RQs, which non-cellular can be replaced, and in which domain, e.g. DSRC for automotive, WIFI for IoT, smart manufacturing}

\begin{figure}[!t]
    \centering
    \includegraphics[trim={1.4cm 0.9cm 0.5cm 0.4cm},clip,width=0.85\textwidth, cfbox=blue]{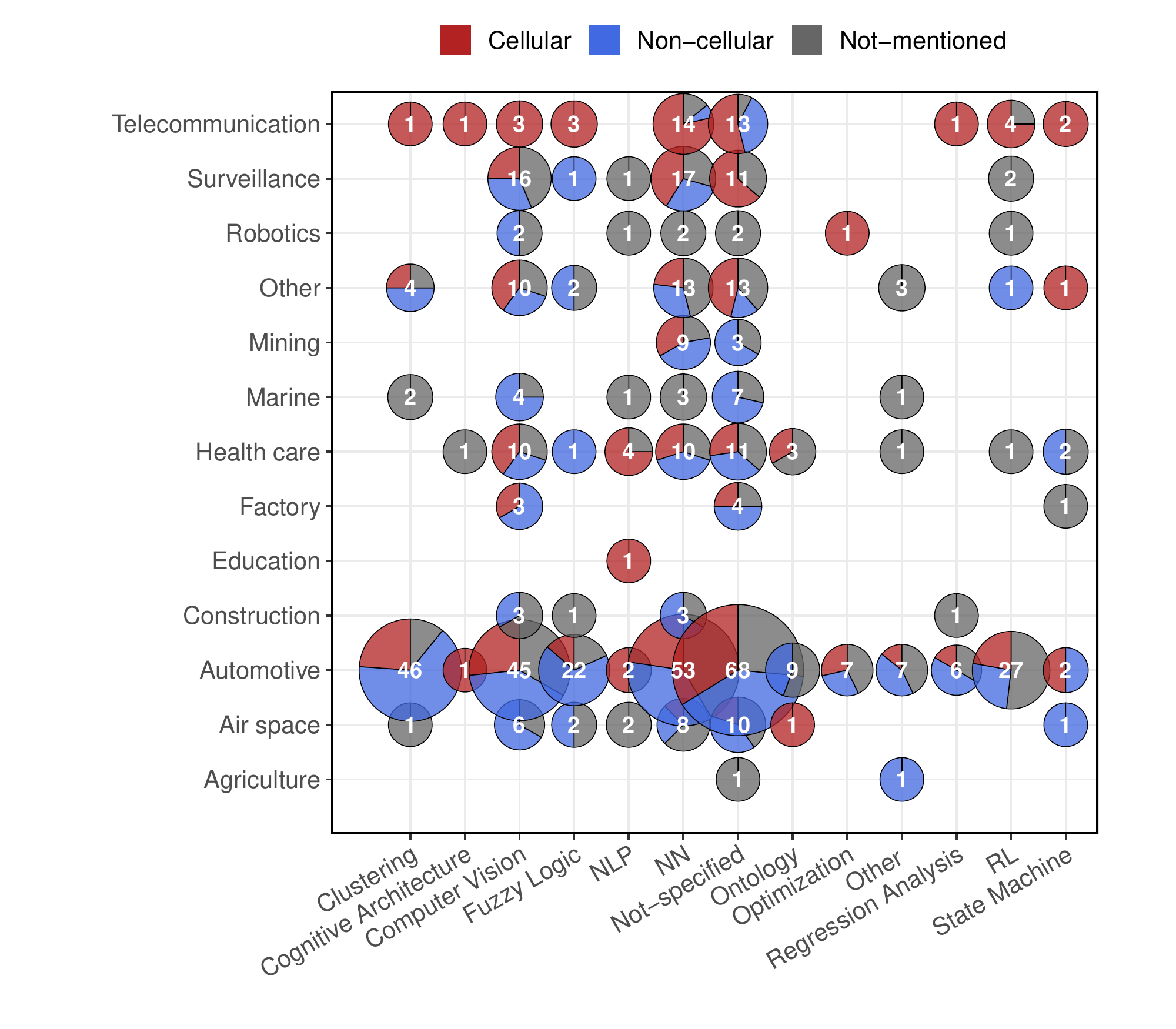}
    \caption{Bubble pie char illustrating the relationship between communication technologies, AI algorithms and number of relevant publications.}%\textcolor{red}{ We merge all cellular technologies, only 4 values on y-axis. we create 2 new bubble charts, 1 for cellular and 1 non-cellular}}  
    \label{fig:alg_dom_comm}
\end{figure}

This research question addresses the possibility of using cellular communication to implement AI-based safety in application domains that does not have large presence of this communication technology, but have brought lot of attention of non-cellular communication.
\autoref{fig:alg_dom_comm} presents a similar chart as used in RQ2a, but with pie charts placed in each bubble. Each pie chart presents the amount of publications that correlates to a certain AI algorithm and application domain, and the distribution of communication technology. 

\textcolor{blue}{In this chart, the top five largest bubbles are predominant in the automotive domain. This includes the intersections with ``\textit{NN}'' (53), ``\textit{Clustering}'' (46), ``\textit{Computer Vision}'' (42), ``\textit{RL}'' (27) and ``\textit{Fuzzy Logic}'' (22). The corresponding pie charts from this list show a bigger portion of non-cellular communication than cellular one in all cases. This indicates that cellular communication has some opportunity to embrace automotive problems as already discussed in RQ1c (\autoref{subsec:RQ1c}).} 
\textcolor{blue}{The subsequent largest bubbles are located in ``\textit{Surveillance}'' domain, where ``\textit{NN}'' and ``\textit{Computer Vision}'' have 16 and 17 studies, respectively. }

\textbf{Discussion: }The analysis of the pie charts show a large portion of cellular communication in ``\textit{NN}'' and an opposite situation in ``\textit{Computer Vision}''. This shows an opportunity to address safety using computer vision in surveillance combined with cellular technology. There are few large bubbles in ``\textit{Heath Care}'' domain. In this, the pair with ``\textit{NN}'' (10 studies) shows a pie chart that has non-cellular communication as the main option. Therefore, an opportunity exists to use cellular communication together with health care and neural networks. It is also noticed that some domains have a low usage or even absence of cellular communication as observed in ``\textit{Marine}'', ``\textit{Mining}'' and ``\textit{Robotics}'' domains. In case of marine and mining, the coverage of cellular networks is still limited in these areas. Similarly, as robotic applications are usually targeted to indoor spaces, cellular communication becomes a natural choice.

\textcolor{blue}{In summary, we can interpret that in many domains, there is a potential to replace non-cellular communication by cellular one. In particular, the advent of 5G could promote this transition as higher coverage, capacity, speed and lower latency are expected.}

\subsection{Results of \textbf{Research Question 3(a--b)}}

This section presents results for RQ3, which addresses the publication trends in point of view of communication technology and AI algorithms to implement safety.

\subsubsection*{\textbf{RQ3a.} What is the publication trend with respect of time in terms of cellular and non-cellular communications?} %\sara{communications???}}
\label{RQ3a}
% \hfill

\begin{figure}[!h]
    \centering
    \includegraphics[width=0.85\textwidth, cfbox=blue]{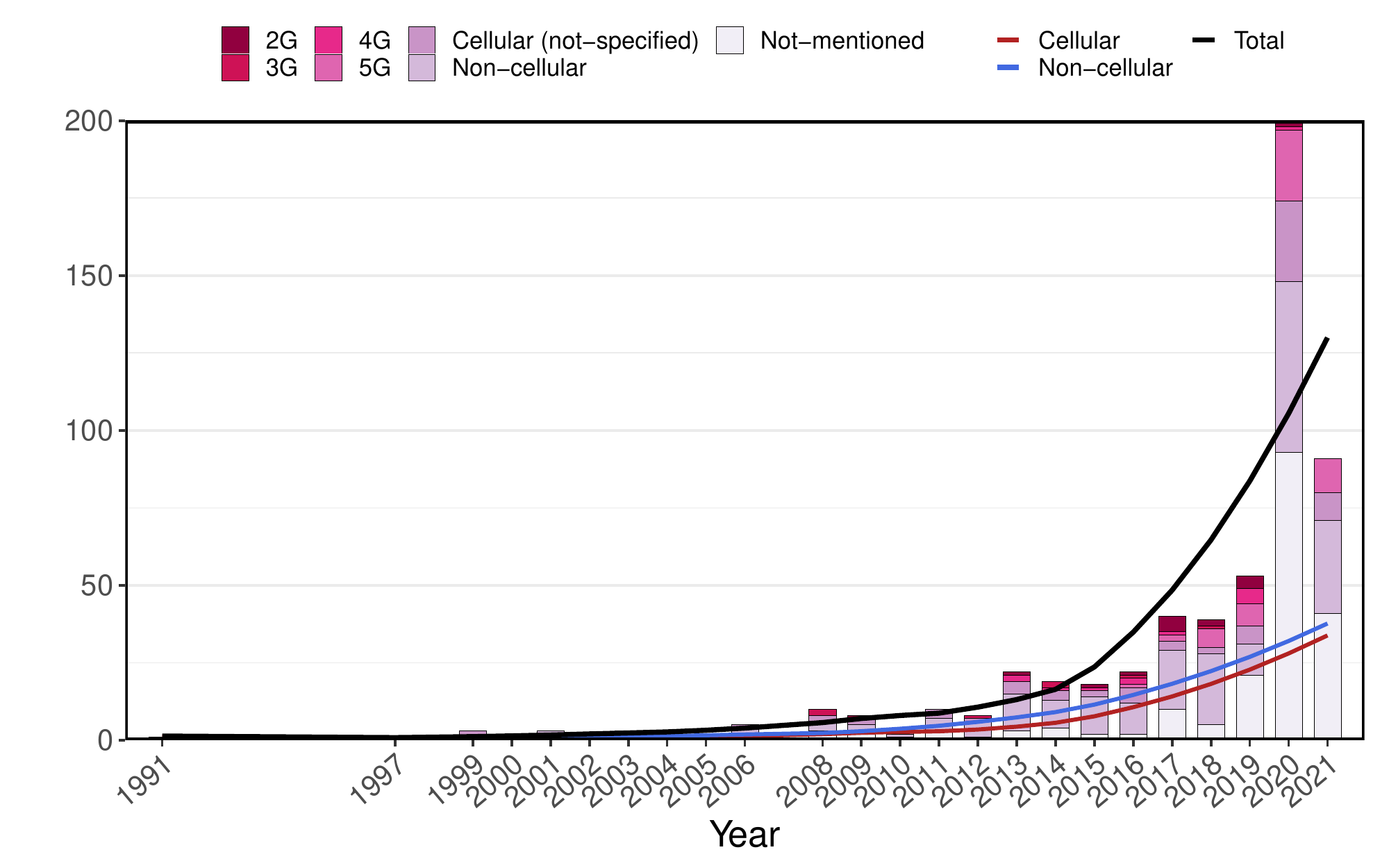}
    \caption{Distribution of publications with respect of time regarding the communication technologies. The fitted curves represent the publication trend in respect to cellular, non-cellular and the combination of both communication technologies. Values of each stacked bar can be checked in \autoref{tab:publication_year}.}%\textcolor{red}{Like Fig 6 of Trustworthiness paper} }%\sara{you can also add one line maybe for the "total number" and make a "trend line" base on that (I guess you know that the excel can calculate and make trend line for charts)}  
    % \label{fig:yyy}
    \label{fig:comm_year}
\end{figure}

This research question targets the publication trends of the  communication technologies with respect to time. The total number of relevant publications of communication technology was plotted against the publication year. 
The outcome is displayed in the bar chart of \autoref{fig:comm_year}. Each stacked bar represents the amount of publication in a certain year and the contribution made by a particular communication technology in that year. \textcolor{blue}{The three curves represent the fitted line of cellular (red line), non-cellular (blue line) and total (black line) publications over the years. From these lines, a trend in the increase of publications in both cellular and non-cellular technologies is observed. Likewise, the ``\textit{Total}'' curve follows the same tendency. By visually comparing the cellular and non-cellular trend curves, it is noticed that there are slightly more studies on the non-cellular communication.} The amount of publication by year for each communication technology is presented in the \autoref{tab:publication_year}.

% The outcome is displayed in the bar chart of \autoref{fig:yyy}. Each stacked bar represents the amount of publication in a certain year and the contribution made by a particular communication technology in that year. The curves represent the fitted line of cellular, non-cellular and total publications over the years. From these lines, a trend in the increase of publications in both cellular and non-cellular technologies is observed. Likewise, the total curve follows the same tendency. The amount of publication by year for each communication technology is presented in the \autoref{tab:publication_year}.

% The outcomes are then displayed by the line chart in figure \ref{fig:yyy}. Each marked line represents a different communication technology according to the classification in \ref{classification}, while the dotted line, derived from the total number of primary studies (in the figure: "Grand Total"), represents the trend of the whole publications over the years. 

\begin{table}[!ht]
\centering
\begin{tabular}{>{\color{blue}}r|>{\color{blue}}r>{\color{blue}}r>{\color{blue}}r>{\color{blue}}r>{\color{blue}}r>{\color{blue}}r>{\color{blue}}r|>{\color{blue}}r}
  \toprule
 \textbf{Year} & 2G & 3G & 4G & 5G & NS & NC & NM & \textbf{Total} \\ 
  \midrule
  \rowcolor{lightgray}1991 &  &  &  &  &   1 &  &  & 1 \\ 
  1997 &  &  &  &  &  &   1 &  & 1 \\ 
  \rowcolor{lightgray}1999 &  &  &  &  &   2 &  &   1 & 3 \\ 
  2000 &  &  &  &  &   1 &  &  & 1 \\ 
  \rowcolor{lightgray}2001 &  &  &  &  &  &   1 &   2 & 3 \\ 
  2002 &  &  &  &  &  &   1 &  & 1 \\ 
  \rowcolor{lightgray}2003 &  &  &  &  &   2 &   1 &  & 3 \\ 
  2004 &  &  &  &  &  &   2 &  & 2 \\ 
  \rowcolor{lightgray}2005 &  &  &  &  &  &   1 &  & 1 \\ 
  2006 &  &  &  &  &   1 &   4 &  & 5 \\ 
  \rowcolor{lightgray}2008 &  &   2 &  &  &   2 &   3 &   3 & 10 \\ 
  2009 &   1 &  &  &  &   2 &   3 &   2 & 8 \\ 
  \rowcolor{lightgray}2010 &  &  &  &  &   2 &   1 &   1 & 4 \\ 
  2011 &  &  &  &  &   3 &   4 &   3 & 10 \\ 
  \rowcolor{lightgray}2012 &  &   1 &  &  &   1 &   5 &   1 & 8 \\ 
  2013 &   1 &  &   2 &  &   4 &  12 &   3 & 22 \\ 
  \rowcolor{lightgray}2014 &  &   2 &   1 &  &   3 &   9 &   4 & 19 \\ 
  2015 &   1 &  &   1 &  &   2 &  12 &   2 & 18 \\ 
  \rowcolor{lightgray}2016 &   1 &   1 &   2 &   1 &   5 &  10 &   2 & 22 \\ 
  2017 &   5 &  &   1 &   2 &   3 &  19 &  10 & 40 \\ 
  \rowcolor{lightgray}2018 &   2 &   1 &  &   6 &   2 &  23 &   5 & 39 \\ 
  2019 &   4 &  &   5 &   7 &   6 &  10 &  21 & 53 \\ 
  \rowcolor{lightgray}2020 &   2 &  &   1 &  23 &  26 &  55 &  93 & 200 \\ 
  2021 &  &  &  &  11 &   9 &  30 &  41 & 91 \\ 
   \bottomrule
\end{tabular}
\caption{Publication amount with respect to the year and communication technology. NS, NC and NM denote cellular not-specified, non-cellular and not-mentioned, respectively.}
\label{tab:publication_year}
\end{table}

\textcolor{blue}{From the \autoref{fig:comm_year} (and \autoref{tab:publication_year}), it can be observed that the first mention of particular communication technology, a cellular one, was far back in 1991. Between 1991 and 2008 the number of publications is mostly rare and not significant. In 2008 however, a total number of 10 studies were registered. In this year the first publication mentioning the use of cellular technology (3G) in AI to implement safety was registered. After a couple of years that followed a boost. In 2013 a big step had been taken, when the number of publications became more than twice larger than the one in 2008, resulting in 22 publications. After that year, the trend only moves upwards, reaching its peak in 2017 with 40 relevant studies. The first year where each category had at least 1 publication was in 2016. Further we observe an increased use of 5G from 2016 on-wards for safety implementations.}

\textcolor{blue}{\textbf{Discussion: }In 2020 the amount of publications increased almost four times compared to the previous year, recording 200 publications. From this, 5G had a substantial contribution, with 23 studies, probably due to the rollout of this technology. At the end of the timeline, in 2021 is shown a drop in the amount of the total studies as not all publications were accounted. However, this number is already larger than the total amount found in 2019.}
% In the end, the significant drop from 2018 to 2019 happened given that the search in the digital libraries was made in early 2020, therefore we think that not all publications from 2019 are accounted.}
% This happens because a lot of studies take quite a considerable amount of time to be published, considering their issues in proceedings.

\subsubsection*{\textbf{RQ3b.} What is the publication trend with respect of time in terms of AI algorithms?}% and application domains?}
% \hfill

\begin{figure}[!h]
    \centering
    \includegraphics[trim={0cm 0.8cm 0 0cm},clip,width=0.80\textwidth, cfbox=blue]{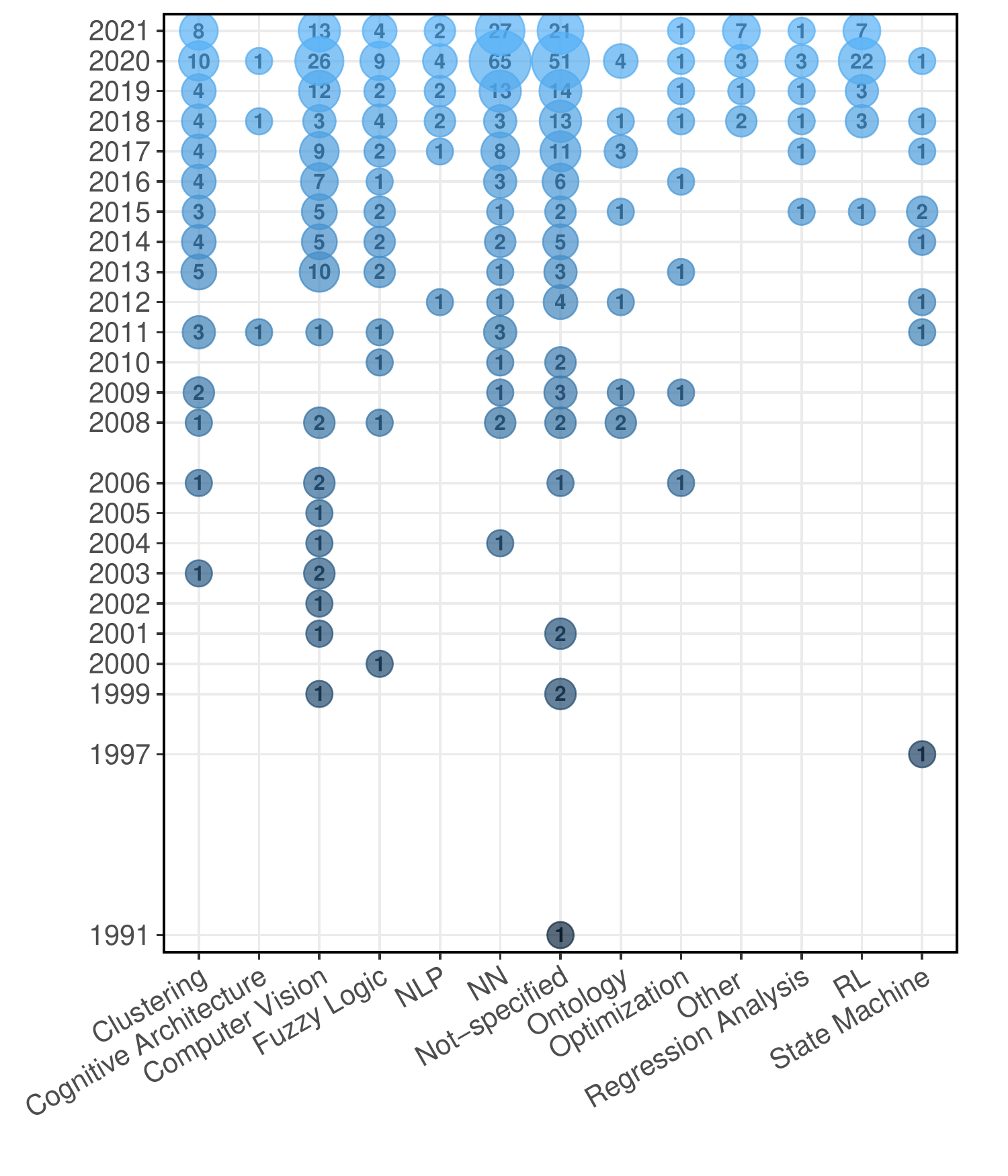}
    \caption{Bubble char presenting the relationship between AI algorithm, year of publication and number of relevant studies.}  
    \label{fig:bubblechart2}
\end{figure}

The essence of this research question is to analyse the publication trend in terms of AI algorithms to implement safety. 
% In \autoref{fig:bubblechart2}, a bubble chart addressing the relationship between AI algorithms, year of publications and number of relevant papers is depicted. 
\autoref{fig:bubblechart2} shows a the amount of publications related to AI algorithm to implement safety that were published in each year and are represented by the bubble size. 
% The bubble chart is combined by the years in the y-axis, algorithms in the x-axis, and the bubbles in the middle representing the number of relevant studies.\\
\textcolor{blue}{The biggest bubble belongs to the pair ``\textit{NN -- 2020}'' (65 studies), reasoning that the popularity of neural networks is getting high even in recent years. }

\textbf{Discussion: }While considering the whole timeline, it is verified that computer vision has the largest bubbles spread along the years. Meanwhile, fuzzy logic has the lowest bubbles spread over the timeline. A considerable change is present from 2019 to 2020 in reinforcement learning (from 3 to 22),  which suggests that recently this method is getting more attention. Overall, a rapid increase is observed in 2020, almost all algorithms presented an increase in its usage, and is ongoing in 2021 also. A drop in the amount of the total studies in 2021 is because not all publications were accounted in this study. However, as mentioned previously, this number is already much larger than the total amount found in 2019.

%\textcolor{blue}{We observe a relative boost in AI-based safety from the year 2018 and on-wards.
%When considering the year %``\textit{2018}'', a boost of AI-based safety was relatively recent. 
%To be noted that the majority of the studies that explicitly mention the AI algorithm are published after 2013 (22 studies) , and we observe a boost from year 2017 (38 studies) and on-wards. 
%In overall, the outcomes present an increasing trend of the publications. 
%However, it seems as for particular algorithms such as \textit{computer vision}, the interest of the community has slightly decreased very recently, after peaking in 2010. 
%The drop between 2018 and 2019 was previously discussed in \textbf{RQ3a}. 
%On the contrary, the Neural Network presents an increase tendency, mostly increasing from 2017.}

\subsection{Results of \textbf{Research Questing 4-6}}

This section groups research questions related to research contributions, research types and author's affiliation.\\

\subsubsection*{\textbf{RQ4.} What type of research contributions are mainly presented in the studies?}
% \hfill

\begin{figure}[!b]
    \centering
    \includegraphics[trim={3cm 0.6cm 0 0.7cm},clip, width=0.80\textwidth, cfbox=blue]{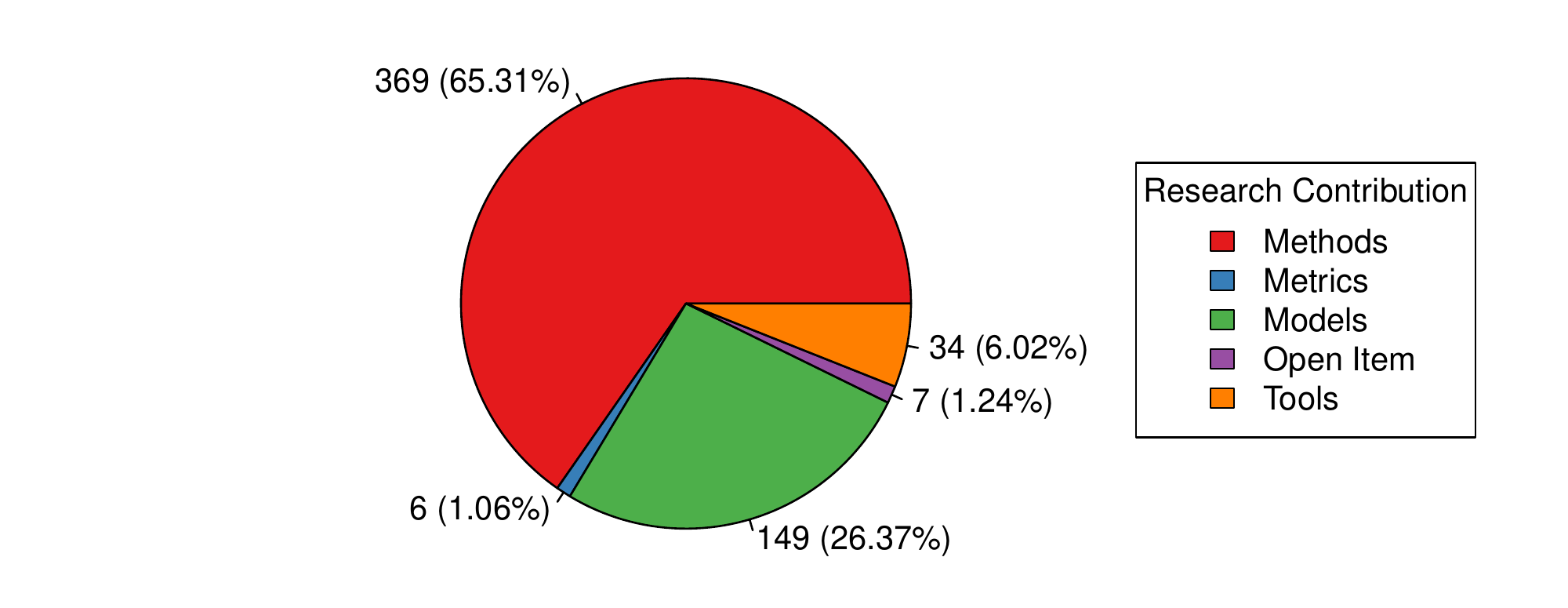}
    \caption{Distribution of studies regarding the types of research contribution.} %\sara{Please change the label to percentage and not the number.}
     %\sara{I suggest to change the position of "Open Item" to the end of the graph(the last right place)}
    
    \label{fig:RQ4}
\end{figure}

% In this section, the \textbf{RQ4} results are presented. 
\textcolor{blue}{\autoref{fig:RQ4} displays a pie chart, where every slice represents the number of publications according to the research contribution category defined in classification scheme (details in Section~\ref{sec:classscheme}). The contribution categories are \textit{Methods, Metrics, Models, Tools} and \textit{Open Items}. The chart shows that the researchers are mostly focused on presenting \textit{Methods} as a research contribution, comprising 65.31\%, which corresponds to 369 publications. \textit{Models} and \textit{Tools} come after with 26.37\% and 6.02\% respectively. The least amount of studies, with only 6 publications (1.06\%) discuss \textit{Metrics}. There were 7 studies (1.24\%) that was not possible to determine the contribution class.} 

\textbf{Discussion: }From the outcomes, it can be said that the spotlight from the research community is towards giving solutions in form of methods or approaches, and lacking especially metrics for providing safety using AI and communication technology. One reason for this lack could be that the metrics to measure safety are usually provided in safety standards only, which could only be implied or followed by the software implementing AI algorithms using communication.

\subsubsection*{\textbf{RQ5.} Which are the main research types being employed in the studies?}
\hfill

\begin{figure}[h]
    \centering
    \includegraphics[trim={0 1.9cm 0 1.9cm},clip,width=0.85\textwidth, cfbox=blue]{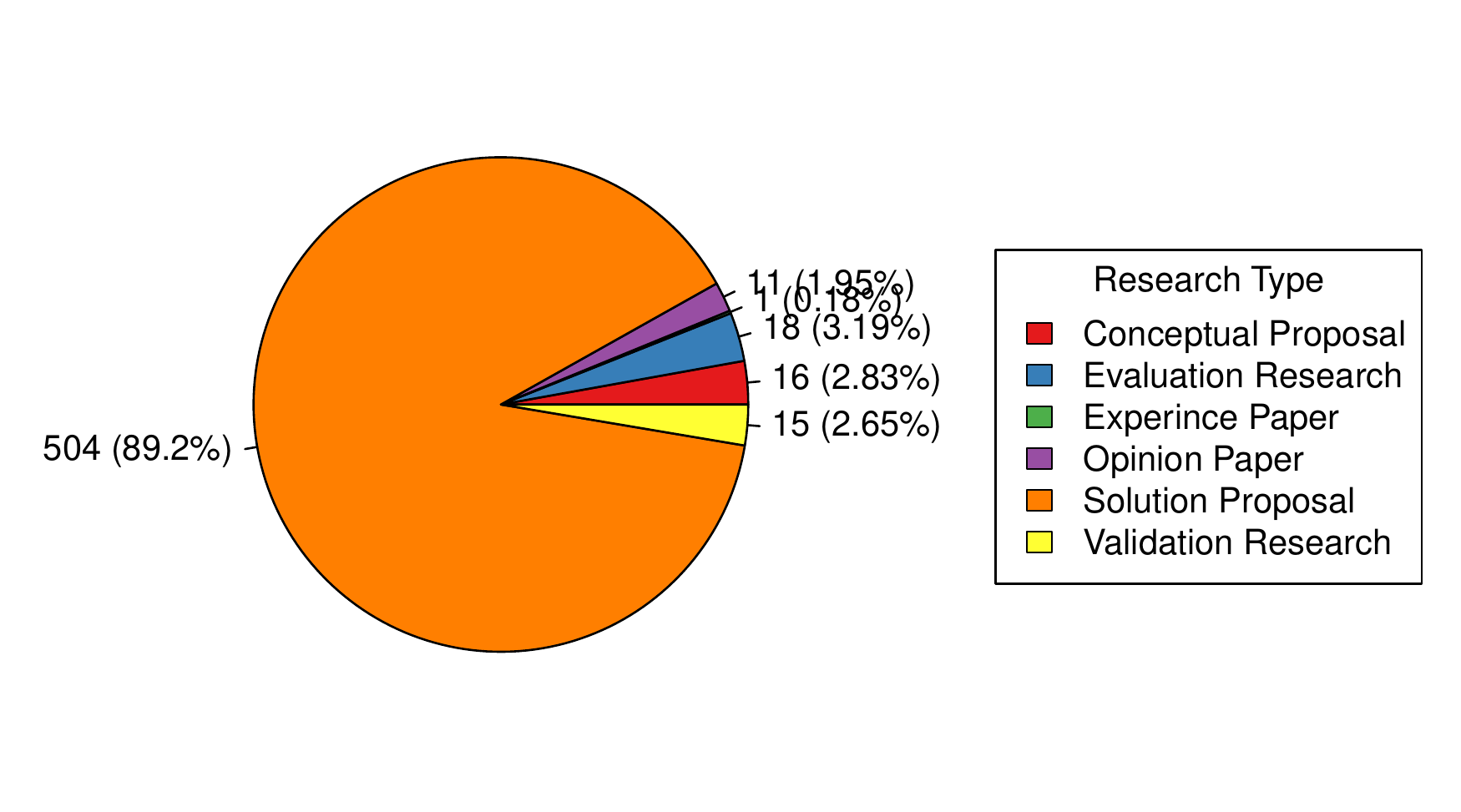}
    \caption{Distribution of studies regarding the research types.} %\sara{Please change the label to percentage and not the number.}
    \label{fig:RQ5}
\end{figure}

\autoref{fig:RQ5} depicts a pie chart with the number of publications and their percentage grouped by research type classification. In Section~\ref{sec:research_type} the research type classification was acknowledged as \textit{Validation Research}, \textit{Evaluation Research}, \textit{Solution Proposal}, \textit{Conceptual Proposal}, \textit{Experience Paper} and \textit{Opinion Paper}. \textcolor{blue}{The chart shows that the vast majority of the publications embrace the \textit{Solution Proposal}, which composes 89.20\% of the total number of relevant studies (504 studies). 
% It was found 3.19\% of publications employed \textit{Evaluation Research} and \textit{Validation Research} each. 
Few studies fall in the remaining research types: \textit{Evaluation Research} -- 3.19\%, \textit{Conceptual Proposal} -- 2.83\%, \textit{Validation Research} -- 2.65\%, \textit{Opinion Paper} -- 1.95\% and \textit{Experience Paper} -- 0.18\%.} 

\textbf{Discussion: }From these outcomes we could conclude that a large number of solutions are yet to be evaluated and validated by the industry. One reason for this could be the novelty of the topic and its recent boost.         

% \textcolor{red}{In order to be consistent with our classification, we classified 150 papers together at the start of the review process, and marked as R, NR, or NC. and then the same number of papers we classified together.}

\subsubsection*{\textbf{RQ6.} What is the distribution of publications in terms of academic and industrial affiliation?}
% \hfill

In order to have a better understanding of publication origin, whether they come from academia or the industry, the affiliations of the authors were taken into account. The categorization was created based on the one in \cite{asadollah201710}. 
It contains three data items: \textit{academia}, \textit{industry} and \textit{both}, as follows:
%. The following categorization was applied: 

\begin{itemize}
    \item \textit{Academia} - The study is written only by academic authors.
    \item \textit{Industry} - The study is written only by industrial authors.
    \item \textit{Both} - The study is conducted as a consequence of a collaboration between academic and industrial authors. 
\end{itemize}
\noindent

\begin{figure}[h]
    \centering
    \includegraphics[trim={3cm 0.9cm 0 0.9cm},clip, width=0.70\textwidth, cfbox=blue]{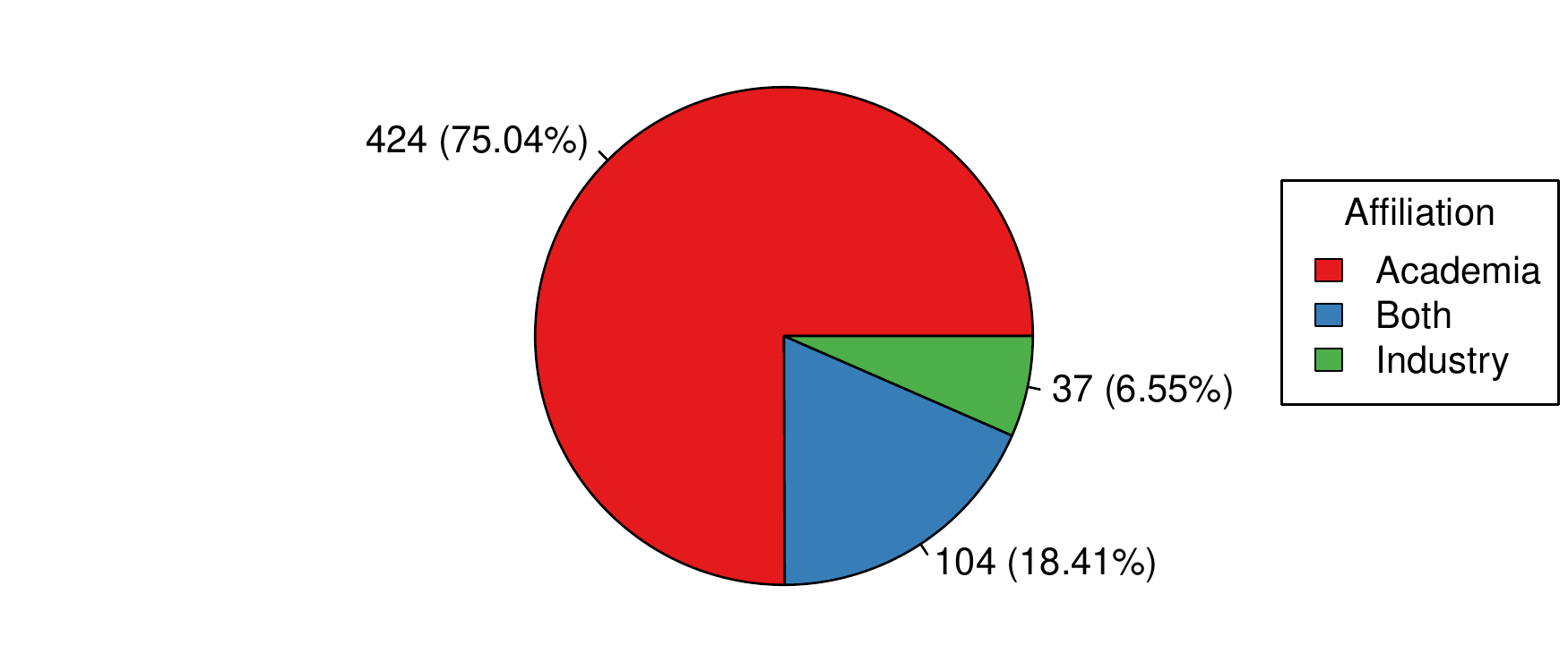}
    \caption{Distribution of studies regarding the affiliation of the authors.}
    \label{fig:RQ6}
\end{figure}

\textcolor{blue}{In \autoref{fig:RQ6}, a pie chart depicts the distribution of the publications addressing the affiliation of the authors. The chart shows that majority of the studies, 75.04\% (424 publications), correspond to \textit{Academic} publications. The remaining is distributed into 6.55\% (37 publications) from \textit{Industry} and 18.41\%  (104 publications) of the studies are derived from \textit{Both}.} The lack of industrial publications was confirmed by the low number of industrial authors participating in the research area.

\textbf{Discussion: }The main reason for this low number of industrial publications is that usually the industrial research is not published in the form of research papers. Even if it does, most of industrial publications consist in a format of patents or white papers that might reveal less information and consequently not being considered by this study (as stated in the exclusion criteria). That is why, we think that this low number of papers is not a realistic indicator, considering the fact that AI techniques are already adopted in many application domains. However, there is a tentative to increase the joint studies, bringing the academia and industry to work together as we can see in the chart.          

\section{Threats to validity}
\label{sec:validity}

When conducting a systematic mapping study, there are usually some issues that might jeopardize the validity of the work. In this subsection, the method to mitigate the threats of validity is described, which follows the the guidelines of Wohlin et al. in \cite{wohlin2012experimentation}. 

%\begin{itemize}
\subsection{Construct validity} 

Construct validity in systematic mapping studies stresses the validity of the extracted data concerning the research questions. For this systematic mapping study, the guidelines in \cite{petersen2015guidelines} were followed. The selection of the primary studies represents the research questions in a confident way. Starting with the search string, it is well-formulated using the PICO method. Then, the most common digital libraries in software engineering were chosen to perform the automatic search, which gives more depth to the study as mentioned in Section \ref{sec:stringSource}. The publications collected by performing the search were reviewed with respect to their titles and abstracts. If a publication could not be judged for relevance based on its title and abstract, full-text skimming was performed to decide about its relevance in the research area. 

Furthermore, the studies were screened under rigorous inclusion and exclusion criteria. To be noted is that not using ``snowballing sampling'' might reduce the credibility of the study, but the big number of primary studies covers the desired scope. Considering all the above-mentioned, it is believed that no relevant studies were left out of the research.  
Moreover, the selection team members classified a certain number of common studies (170 out of 4808 studies have been randomly selected) to check the overall agreement, achieving 0.80 which shows almost perfect agreement.

\subsection{Internal validity} 
Internal validity represents those threats that seek to establish a causal relationship, where some conditions may influence the study. To mitigate this threat, a predefined protocol with an exact data extraction form was used. Regarding the data analysis, since descriptive statistics would be employed, the threats are minimal.  

Another threat is related to the classification schema for mapping. To mitigate,we partially based our classification of the primary studies on  initial  categorization, which is based on the discussions with experts from their respective fields and refined the categories further using few iterative meetings. Further, we systematically applied the keywording method to define the classification schemes. We believe that this process of classification would have refined our mapping to mitigate this threat.

\subsection{External validity} 

Threats to external validity define the extent of the generalizability of the outcomes of systematic mapping study. In this research, the most drastic threat related to external validity is not covering the whole scope of AI using communication technologies to provide safety. This threat is mitigated by a well-established search string and the automatic search. Having a good set of inclusion and exclusion criteria helped as well in the external validity of the study. However, since the T-A-K skimming was performed, due to abstracts with incomplete information, some relevant studies might have been wrongly excluded.
However, as it is known that abstracts do not always reveal the true content of papers, there is a risk that we might have excluded a paper with poor abstract but valid content.
%\textcolor{red}{Sara do we move your last para from Construct validity here?}

To further minimize the external threat, only the publications written in English (widely used language in qualitative scientific studies) were chosen and additionally only the peer-reviewed studies were accepted.
    
\subsection{Conclusion validity} 

Conclusion validity demonstrates that the study can be repeated in the future and yet have the same results. The relationship between the extracted and synthesized data and the outcomes of the study may influence the repetitiveness.

In order to mitigate this threat, all the steps leading to data analysis including search string, automatic search, selection criteria. are well-documented. They can be used by researchers to replicate the study. 
Furthermore, the data extraction form is well-documented, decreasing the biases of that process. The same reasoning is followed with the classification scheme as well, where the framework is well-defined and has the respective references.

\section{Conclusions and future directions}
\label{sec:conclusions}
The main goal of this paper is first, to grasp an overview of the current state-of-the-art research publications that utilizes AI to implement safety using different communication technologies, second, to investigate gaps in it, and third, to identify emerging future research trends on the topic. To achieve this, a systematic mapping analysis is conducted. From an initial total of \textcolor{blue}{8760} studies obtained after an automatic search, a number of \textcolor{blue}{565} relevant studies were identified using structured selection process, covering three main aspects: AI algorithm, specific communication technology and safety as a property. By classification of these \textcolor{blue}{565} studies, we have identified different application domains using AI and communications to implement safety aspects, the current challenges and emerging future research trends on the topic from both industry and academia. 

%We have also investigated the potential increased use of the cellular communication specially 5G to implement AI-based safety in multiple domains and an increasing interest of the industry on the topic. 
The study results indicate the start of an increased use of the cellular communication, specially 5G, to implement AI-based safety in multiple domains and an increasing interest of the industry on the topic.
The outcomes of the study also suggests that regarding AI algorithms, the attention of researchers is focused more towards applying \textcolor{blue}{\textit{neural networks} and} \textit{computer vision} techniques, even though a high percentage of studies do not specify the technique, threatening the validation of the previous statement. Meanwhile, the use of techniques like \textit{optimization} or \textit{cognitive architecture} are less preferred. \textcolor{blue}{Despite \textit{reinforcement learning} not being the mostly adopted algorithm, it is one of the algorithm that showed the highest increase in 2020.} Concerning the other categories, \textit{non-cellular} is clearly dominant in communication, however we observe \textcolor{blue}{a rapidly increasing recent interest in one of the latest cellular networks i.e. 5G specially from 2020}. The topic is generally gaining the attention of the researchers in the past years and the trend seems to remain increasing. \textcolor{blue}{The analysis further reveals that currently \textit{automotive} outnumbers all the other application domains (52.21\%) followed by surveillance and health care (8.5\% and 7.79\%, respectively) . Moreover, it can be concluded that the majority of studies contribute with \textit{methods} and \textit{solution proposals}.}

%\subsection{Future work}
Concerning the future work, the results of this systematic mapping study benefit the researchers and the industry at the same time. Researchers can use this work, to further deepen their knowledge considering new challenges in this topic. The various classifications provided in this work may serve researchers as initial categorizations or a base foundation with the will to further extend them in conducting other systematic mapping studies or systematic literature reviews. 

\textcolor{blue}{The study pinpoints current gaps in research that may represent opportunities for further research on different AI algorithms using communication in multiple domains. These gap areas lack the research which can be explored further. Few such research areas could be in the fields of robotics, mining, factory which are currently behind but can leverage on the results from the automotive area by applying and testing AI algorithms to autonomous agents like robots in smart factories, logistics etc.}
Further, different industries can not only benefit from this gap analysis in their respective areas, but can also start research in new required AI fields. Such as \textit{Autonomous vehicles} are currently doing a lot of research on \textit{Computer vision, neural networks, clustering}. One future direction could be to explore new areas like trustworthy AI aspects, health industry and surveillance also need this heavily. 

%. A lot of automotive domain work on AI algorithms can also be applied and tested on autonomous agents like robots in smart factories, logistics etc. 

Further, the study presents currently increasing research trend, such as one increasing trend is about the autonomous vehicle domain and the effect of the 5G on the studies.  
With the roll-out of 5G in 2020, an increased research trend is to use it in different domains. One future direction could be to research on the use 5G instead of WiFi technologies for safety. 

The lack of evaluation researches indicates that the industry is either not putting the research into practice or not publishing research in the form of research papers. Therefore, an heightened interest in evaluating the proposed solutions like case studies or field experiments is expected. In the near future, more efforts could be put into providing new tools and defining metrics, as it, for now, the research area is missing it.

%One of the directions that was included is about the autonomous vehicle domain and the effect of the 5G on the studies.
%\item A lot of automotive domain work on AI algos can also be applied and tested on autonomous agents like robots in factories, logistics etc.
    %     \item We can make some future directions from the gaps.
    %     \item Separate conclusions form future directions.
    %     \item We extract future directions form the results discussions under RQs.

%To summarize, the study provides a good state-of-the-art for the topic and solid foundations for developing new tools and performing more evaluation researches in the use of software based AI algorithms to provide safety using communication technologies.  

\section{Declarations}

\subsection*{Funding}

The research of the last author for this work is supported by Swedish Foundation for Strategic Research(SSF) via the Serendipity project, and KKS SACSys Synergy project (Safe and Secure Adaptive Collaborative Systems).

\subsection*{Conflicts of interest/Competing interests}

The authors declare that they have no conflict of interest.

\subsection*{Availability of data and material}

Not applicable.

\subsection{Code availability}

Not applicable.

% \begin{acknowledgements}
% The research of the last author for this work is supported by the Serendipity project funded by the Swedish Foundation for Strategic Research (SSF).
% \end{acknowledgements}

\bibliographystyle{spmpsci}      % mathematics and physical sciences

\bibliography{References}

\begin{thebibliography}{10}
\providecommand{\url}[1]{{#1}}
\providecommand{\urlprefix}{URL }
\expandafter\ifx\csname urlstyle\endcsname\relax
  \providecommand{\doi}[1]{DOI~\discretionary{}{}{}#1}\else
  \providecommand{\doi}{DOI~\discretionary{}{}{}\begingroup
  \urlstyle{rm}\Url}\fi

\bibitem{Abbas2015}
Abbas, N., Nasser, Y., Ahmad, K.E.: Recent advances on artificial intelligence
  and learning techniques in cognitive radio networks.
\newblock EURASIP Journal on Wireless Communications and Networking
  \textbf{2015}(1), 174 (2015).
\newblock \doi{10.1186/s13638-015-0381-7}.
\newblock \urlprefix\url{https://doi.org/10.1186/s13638-015-0381-7}

\bibitem{Aceto2018}
Aceto, G., Persico, V., Pescapé, A.: The role of information and communication
  technologies in healthcare: taxonomies, perspectives, and challenges.
\newblock Journal of Network and Computer Applications \textbf{107}, 125 -- 154
  (2018).
\newblock \doi{https://doi.org/10.1016/j.jnca.2018.02.008}.
\newblock
  \urlprefix\url{http://www.sciencedirect.com/science/article/pii/S1084804518300456}

\bibitem{Al-Kaff2018}
Al-Kaff, A., Martín, D., García, F., de~la Escalera, A., {María Armingol},
  J.: Survey of computer vision algorithms and applications for unmanned aerial
  vehicles.
\newblock Expert Systems with Applications \textbf{92}, 447 -- 463 (2018).
\newblock \doi{https://doi.org/10.1016/j.eswa.2017.09.033}.
\newblock
  \urlprefix\url{http://www.sciencedirect.com/science/article/pii/S0957417417306395}

\bibitem{Alsamhi2019}
Alsamhi, S.H., Ma, O., Ansari, M.S.: Survey on artificial intelligence based
  techniques for emerging robotic communication.
\newblock Telecommunication Systems \textbf{72}(3), 483--503 (2019).
\newblock \doi{10.1007/s11235-019-00561-z}.
\newblock \urlprefix\url{https://doi.org/10.1007/s11235-019-00561-z}

\bibitem{asadollah201710}
Asadollah, S.A., Sundmark, D., Eldh, S., Hansson, H., Afzal, W.: 10 years of
  research on debugging concurrent and multicore software: a systematic mapping
  study.
\newblock Software quality journal \textbf{25}(1), 49--82 (2017)

\bibitem{bezdek1999image}
Bezdek, J.C., Keller, J., Krisnapuram, R., Pal, N.R.: Image processing and
  computer vision.
\newblock In: Fuzzy Models and Algorithms for Pattern Recognition and Image
  Processing, pp. 547--678. Springer (1999)

\bibitem{Bila2017}
{Bila}, C., {Sivrikaya}, F., {Khan}, M.A., {Albayrak}, S.: Vehicles of the
  future: A survey of research on safety issues.
\newblock IEEE Transactions on Intelligent Transportation Systems
  \textbf{18}(5), 1046--1065 (2017).
\newblock \doi{10.1109/TITS.2016.2600300}

\bibitem{bishop2006pattern}
Bishop, C.M.: Pattern recognition and machine learning.
\newblock springer (2006)

\bibitem{Bkassiny2013}
{Bkassiny}, M., {Li}, Y., {Jayaweera}, S.K.: A survey on machine-learning
  techniques in cognitive radios.
\newblock IEEE Communications Surveys Tutorials \textbf{15}(3), 1136--1159
  (2013)

\bibitem{Bozhinoski2019}
Bozhinoski, D., {Di Ruscio}, D., Malavolta, I., Pelliccione, P., Crnkovic, I.:
  Safety for mobile robotic systems: A systematic mapping study from a software
  engineering perspective.
\newblock Journal of Systems and Software \textbf{151}, 150--179 (2019).
\newblock \doi{https://doi.org/10.1016/j.jss.2019.02.021}

\bibitem{brereton2007lessons}
Brereton, P., Kitchenham, B.A., Budgen, D., Turner, M., Khalil, M.: Lessons
  from applying the systematic literature review process within the software
  engineering domain.
\newblock Journal of systems and software \textbf{80}(4), 571--583 (2007)

\bibitem{Cooper2017}
{Cooper}, C., {Franklin}, D., {Ros}, M., {Safaei}, F., {Abolhasan}, M.: A
  comparative survey of vanet clustering techniques.
\newblock IEEE Communications Surveys Tutorials \textbf{19}(1), 657--681 (2017)

\bibitem{dadios2012fuzzy}
Dadios, E.: Fuzzy logic: algorithms, techniques and implementations.
\newblock BoD--Books on Demand (2012)

\bibitem{Erpek2020}
Erpek, T., O'Shea, T.J., Sagduyu, Y.E., Shi, Y., Clancy, T.C.: Deep Learning
  for Wireless Communications, pp. 223--266.
\newblock Springer International Publishing (2020).
\newblock \doi{10.1007/978-3-030-31764-5_9}.
\newblock \urlprefix\url{https://doi.org/10.1007/978-3-030-31764-5_9}

\bibitem{Pouyan2020}
Esmaeilzadeh, P.: Use of {AI}-based tools for healthcare purposes: a survey
  study from consumers' perspectives.
\newblock BMC medical informatics and decision making \textbf{20}(1), 170--170
  (2020).
\newblock \doi{10.1186/s12911-020-01191-1}.
\newblock Publisher: BioMed Central

\bibitem{Everitt2018}
Everitt, T., Lea, G., Hutter, M.: Artificial general intelligence {(AGI)}
  safety literature review.
\newblock In: Proceedings of the Twenty-Seventh International Joint Conference
  on Artificial Intelligence, {IJCAI-18}, pp. 5441--5449. International Joint
  Conferences on Artificial Intelligence Organization (2018).
\newblock \doi{10.24963/ijcai.2018/768}.
\newblock \urlprefix\url{https://doi.org/10.24963/ijcai.2018/768}

\bibitem{Fayyad2020}
Fayyad, J., Jaradat, M.A., Gruyer, D., Najjaran, H.: Deep learning sensor
  fusion for autonomous vehicle perception and localization: A review.
\newblock Sensors \textbf{20}(15) (2020).
\newblock \doi{10.3390/s20154220}

\bibitem{Fellan2018}
{Fellan}, A., {Schellenberger}, C., {Zimmermann}, M., {Schotten}, H.D.:
  Enabling communication technologies for automated unmanned vehicles in
  industry 4.0.
\newblock In: 2018 International Conference on Information and Communication
  Technology Convergence (ICTC), pp. 171--176 (2018)

\bibitem{fensel2001ontologies}
Fensel, D.: Ontologies.
\newblock In: Ontologies, pp. 11--18. Springer (2001)

\bibitem{Garcia2015}
Garc{{\'i}}a, J., Fern, o~Fern{{\'a}}ndez: A comprehensive survey on safe
  reinforcement learning.
\newblock Journal of Machine Learning Research \textbf{16}(42), 1437--1480
  (2015).
\newblock \urlprefix\url{http://jmlr.org/papers/v16/garcia15a.html}

\bibitem{gladyshev2004finite}
Gladyshev, P., Patel, A.: Finite state machine approach to digital event
  reconstruction.
\newblock Digital Investigation \textbf{1}(2), 130--149 (2004)

\bibitem{Guiochet2017}
Guiochet, J., Machin, M., Waeselynck, H.: {Safety-critical advanced robots: A
  survey}.
\newblock {Robotics and Autonomous Systems} \textbf{94}, 43--52 (2017).
\newblock \doi{10.1016/j.robot.2017.04.004}.
\newblock \urlprefix\url{https://hal.archives-ouvertes.fr/hal-01394136}

\bibitem{Huang2020}
Huang, X., Kroening, D., Ruan, W., Sharp, J., Sun, Y., Thamo, E., Wu, M., Yi,
  X.: A survey of safety and trustworthiness of deep neural networks:
  Verification, testing, adversarial attack and defence, and interpretability.
\newblock Computer Science Review \textbf{37}, 100270 (2020).
\newblock \doi{https://doi.org/10.1016/j.cosrev.2020.100270}.
\newblock
  \urlprefix\url{http://www.sciencedirect.com/science/article/pii/S1574013719302527}

\bibitem{InamSOCNE2015}
Inam, R., Karapantelakis, A., Vandikas, K., Mokrushin, L., Vulgarakis~Feljan,
  A., Fersman, E.: Towards automated service-oriented lifecycle management for
  {5G} networks.
\newblock In: 2015 IEEE 20th Conference on Emerging Technologies Factory
  Automation (ETFA), pp. 1--8 (2015)

\bibitem{InamETFA18}
{Inam}, R., {Raizer}, K., {Hata}, A., {Souza}, R., {Forsman}, E., {Cao}, E.,
  {Wang}, S.: Risk assessment for human-robot collaboration in an automated
  warehouse scenario.
\newblock In: IEEE 23rd International Conference on Emerging Technologies and
  Factory Automation (ETFA), pp. 743--751 (2018)

\bibitem{InamETFA2016}
Inam, R., Schrammar, N., Wang, K., Karapantelakis, A., Mokrushin, L., Feljan,
  A.V., Fersman, E.: Feasibility assessment to realise vehicle teleoperation
  using cellular networks.
\newblock In: 2016 IEEE 19th International Conference on Intelligent
  Transportation Systems (ITSC), pp. 2254--2260 (2016)

\bibitem{Islam2015}
{Islam}, S.M.R., {Kwak}, D., {Kabir}, M.H., {Hossain}, M., {Kwak}, K.: The
  internet of things for health care: A comprehensive survey.
\newblock IEEE Access \textbf{3}, 678--708 (2015)

\bibitem{jalali2012systematic}
Jalali, S., Wohlin, C.: Systematic literature studies: database searches vs.
  backward snowballing.
\newblock In: Proceedings of the 2012 ACM-IEEE international symposium on
  empirical software engineering and measurement, pp. 29--38. IEEE (2012)

\bibitem{Jha2019}
Jha, K., Doshi, A., Patel, P., Shah, M.: A comprehensive review on automation
  in agriculture using artificial intelligence.
\newblock Artificial Intelligence in Agriculture \textbf{2}, 1 -- 12 (2019).
\newblock \doi{https://doi.org/10.1016/j.aiia.2019.05.004}.
\newblock
  \urlprefix\url{http://www.sciencedirect.com/science/article/pii/S2589721719300182}

\bibitem{kitchenham2007guidelines}
Kitchenham, B., Charters, S.: Guidelines for performing systematic literature
  reviews in software engineering.
\newblock Tech. rep., Technical report EBSE-2007-01, Keele University and
  University of Durham (2007)

\bibitem{Klaine2017}
{Klaine}, P.V., {Imran}, M.A., {Onireti}, O., {Souza}, R.D.: A survey of
  machine learning techniques applied to self-organizing cellular networks.
\newblock IEEE Communications Surveys Tutorials \textbf{19}(4), 2392--2431
  (2017)

\bibitem{Leslie2019}
Leslie, D.: {Understanding artificial intelligence ethics and safety: A guide
  for the responsible design and implementation of AI systems in the public
  sector} (2019).
\newblock \doi{10.5281/zenodo.3240529}.
\newblock \urlprefix\url{https://doi.org/10.5281/zenodo.3240529}

\bibitem{lieto2018role}
Lieto, A., Bhatt, M., Oltramari, A., Vernon, D.: The role of cognitive
  architectures in general artificial intelligence (2018)

\bibitem{Luong2019}
{Luong}, N.C., {Hoang}, D.T., {Gong}, S., {Niyato}, D., {Wang}, P., {Liang},
  Y., {Kim}, D.I.: Applications of deep reinforcement learning in
  communications and networking: A survey.
\newblock IEEE Communications Surveys Tutorials \textbf{21}(4), 3133--3174
  (2019)

\bibitem{Marcon2017}
{Marcon}, P., {Zezulka}, F., {Vesely}, I., {Szabo}, Z., {Roubal}, Z., {Sajdl},
  O., {Gescheidtova}, E., {Dohnal}, P.: Communication technology for industry
  4.0.
\newblock In: 2017 Progress In Electromagnetics Research Symposium - Spring
  (PIERS), pp. 1694--1697 (2017)

\bibitem{Cayamcela2018}
{Morocho Cayamcela}, M.E., {Lim}, W.: Artificial intelligence in {5G}
  technology: A survey.
\newblock In: 2018 International Conference on Information and Communication
  Technology Convergence (ICTC), pp. 860--865 (2018)

\bibitem{Naik2020}
{Naik}, G., {Park}, J.M., {Ashdown}, J., {Lehr}, W.: Next generation {Wi-Fi}
  and {5G NR-U in the 6 GHz} bands: Opportunities and challenges.
\newblock IEEE Access \textbf{8}, 153027--153056 (2020).
\newblock \doi{10.1109/ACCESS.2020.3016036}

\bibitem{petersen2008systematic}
Petersen, K., Feldt, R., Mujtaba, S., Mattsson, M.: Systematic mapping studies
  in software engineering.
\newblock In: 12th International Conference on Evaluation and Assessment in
  Software Engineering (EASE) 12, pp. 1--10 (2008)

\bibitem{petersen2015guidelines}
Petersen, K., Vakkalanka, S., Kuzniarz, L.: Guidelines for conducting
  systematic mapping studies in software engineering: An update.
\newblock Information and Software Technology \textbf{64}, 1--18 (2015)

\bibitem{ripley2007pattern}
Ripley, B.D.: Pattern recognition and neural networks.
\newblock Cambridge university press (2007)

\bibitem{ruder2016overview}
Ruder, S.: An overview of gradient descent optimization algorithms.
\newblock arXiv preprint arXiv:1609.04747  (2016)

\bibitem{Saleh2018}
{Saleh}, N., {Kassem}, A., {Haidar}, A.M.: Energy-efficient architecture for
  wireless sensor networks in healthcare applications.
\newblock IEEE Access \textbf{6}, 6478--6486 (2018)

\bibitem{Sharma2020}
Sharma, A., Vanjani, P., Paliwal, N., Basnayaka, C.M., Jayakody, D.N.K., Wang,
  H.C., Muthuchidambaranathan, P.: Communication and networking technologies
  for {UAVs}: A survey.
\newblock Journal of Network and Computer Applications \textbf{168}, 102739
  (2020).
\newblock \doi{https://doi.org/10.1016/j.jnca.2020.102739}.
\newblock
  \urlprefix\url{http://www.sciencedirect.com/science/article/pii/S1084804520302137}

\bibitem{sirohi2020convolutional}
Sirohi, D., Kumar, N., Rana, P.S.: Convolutional neural networks for 5g-enabled
  intelligent transportation system: A systematic review.
\newblock Computer Communications  (2020)

\bibitem{Socher2012}
Socher, R., Bengio, Y., Manning, C.D.: Deep learning for nlp (without magic).
\newblock In: Tutorial Abstracts of ACL 2012, ACL '12, p.~5. Association for
  Computational Linguistics, USA (2012)

\bibitem{Sutton2018}
Sutton, R.S., Barto, A.G.: Reinforcement Learning: An Introduction.
\newblock A Bradford Book, Cambridge, MA, USA (2018)

\bibitem{Tong2019}
{Tong}, W., {Hussain}, A., {Bo}, W.X., {Maharjan}, S.: Artificial intelligence
  for vehicle-to-everything: A survey.
\newblock IEEE Access \textbf{7}, 10823--10843 (2019)

\bibitem{Tran2020}
{Tran}, H., {Xiang}, W., {Johnson}, T.T.: Verification approaches for
  learning-enabled autonomous cyber-physical systems.
\newblock IEEE Design Test pp. 1--1 (2020)

\bibitem{Ullah2019}
{Ullah}, H., {Gopalakrishnan Nair}, N., {Moore}, A., {Nugent}, C., {Muschamp},
  P., {Cuevas}, M.: {5G} communication: An overview of vehicle-to-everything,
  drones, and healthcare use-cases.
\newblock IEEE Access \textbf{7}, 37251--37268 (2019)

\bibitem{wieringa2006requirements}
Wieringa, R., Maiden, N., Mead, N., Rolland, C.: Requirements engineering paper
  classification and evaluation criteria: a proposal and a discussion.
\newblock Requirements engineering \textbf{11}(1), 102--107 (2006)

\bibitem{wohlin2012experimentation}
Wohlin, C., Runeson, P., H{\"o}st, M., Ohlsson, M.C., Regnell, B., Wessl{\'e}n,
  A.: Experimentation in software engineering.
\newblock Springer Science \& Business Media (2012)

\bibitem{Wollschlaeger2017}
{Wollschlaeger}, M., {Sauter}, T., {Jasperneite}, J.: The future of industrial
  communication: Automation networks in the era of the internet of things and
  industry 4.0.
\newblock IEEE Industrial Electronics Magazine \textbf{11}(1), 17--27 (2017)

\bibitem{xu2008clustering}
Xu, R., Wunsch, D.: Clustering, vol.~10.
\newblock John Wiley \& Sons (2008)

\bibitem{Kun-Hsing2018}
Yu, K.H., Beam, A.L., Kohane, I.S.: Artificial intelligence in healthcare.
\newblock Nature Biomedical Engineering \textbf{2}(10), 719--731 (2018).
\newblock \doi{10.1038/s41551-018-0305-z}.
\newblock \urlprefix\url{https://doi.org/10.1038/s41551-018-0305-z}

\end{thebibliography}

\clearpage
\begin{appendices}

\section{Primary studies}
\label{sec:appendix_A}

{\color{blue}
In this section all the 565 primary studies are presented as follows:\\[0.15cm]
% In this section all the 236 primary studies are presented as follows:\\[0.2cm]
\noindent
\footnotesize
\textbf{[S001]} Aarthy D., Vandanaa S., Varshini M. and Tijitha K., \textit{Automatic identification of traffic violations and theft avoidance}, 2016 2nd International Conference on Science Technology Engineering and Management, ICONSTEM 2016, no volume, pp. 72-76 (2016).\\[0.1cm]
\textbf{[S002]} Abdi Khojasteh H., Abbas Alipour A., Ansari E. and Razzaghi P., \textit{An Intelligent Safety System for Human-Centered Semi-autonomous Vehicles}, Lecture Notes on Data Engineering and Communications Technologies, \textbf{45}, pp. 322-336 (2020).\\[0.1cm]
\textbf{[S003]} Abdi L., Takrouni W. and Meddeb A., \textit{In-vehicle cooperative driver information systems}, 2017 13th International Wireless Communications and Mobile Computing Conference, IWCMC 2017, no volume, pp. 396-401 (2017).\\[0.1cm]
\textbf{[S004]} Abdullah D., Takahashi H. and Lakhani U., \textit{Improving the Understanding between Control Tower Operator and Pilot Using Semantic Techniques - A New Approach}, Proceedings - 2017 IEEE 13th International Symposium on Autonomous Decentralized Systems, ISADS 2017, no volume, pp. 207-211 (2017).\\[0.1cm]
\textbf{[S005]} Adesina D., Adagunodo O., Dong X. and Qian L., \textit{Aircraft Location Prediction using Deep Learning}, Proceedings - IEEE Military Communications Conference MILCOM, \textbf{2019-November}, (no pages found) (2019).\\[0.1cm]
\textbf{[S006]} Adgar A., Jantunen E. and Arnaiz A., \textit{Towards true dynamic decision making in maintenance}, Failure Prevention for System Availability - Proceedings of the 62nd Meeting of the Society for Machinery Failure Prevention Technology, no volume, (no pages found) (2008).\\[0.1cm]
\textbf{[S007]} Adler C. and Strassberger M., \textit{Putting together the pieces-A comprehensive view on cooperative local danger warning}, 13th World Congress on Intelligent Transport Systems and Services, no volume, (no pages found) (2006).\\[0.1cm]
\textbf{[S008]} Agrawal S. and Maheshwari P., \textit{Controlling of Smart Movable Road Divider and Clearance Ambulance Path Using IOT Cloud}, 2021 International Conference on Computer Communication and Informatics, ICCCI 2021, no volume, (no pages found) (2021).\\[0.1cm]
\textbf{[S009]} Ahmad I., Md Noor R. and Reza Z'aba M., \textit{LTE efficiency when used in traffic information systems: A stable interest aware clustering}, International Journal of Communication Systems, \textbf{32(2)}, (no pages found) (2019).\\[0.1cm]
\textbf{[S010]} Ahmed, H. O., \textit{25.3 GOPS Autonomous Landing Guidance Assistant System Using Systolic Fuzzy Logic System for Urban Air Mobility (UAM) Vehicles Using FPGA}, 2020 Integrated Communications Navigation and Surveillance Conference (ICNS), IEEE, p. 5D2-1-5D2-11 (2020).\\[0.1cm]
\textbf{[S011]} Ahsan W., Khan M., Aadil F., Maqsood M., Ashraf S., Nam Y. and Rho S., \textit{Optimized node clustering in VANETs by using meta-heuristic algorithms}, Electronics (Switzerland), \textbf{9(3)}, (no pages found) (2020).\\[0.1cm]
\textbf{[S012]} Ai B., Molisch A., Rupp M. and Zhong Z., \textit{5G key technologies for smart railways}, Proceedings of the IEEE, \textbf{108(6)}, pp. 856-893 (2020).\\[0.1cm]
\textbf{[S013]} Aissa M., Bouhdid B., Ben Mnaouer A., Belghith A. and AlAhmadi S., \textit{SOFCluster: Safety-oriented, fuzzy logic-based clustering scheme for vehicular ad hoc networks}, Transactions on Emerging Telecommunications Technologies, no volume, (no pages found) (2020).\\[0.1cm]
\textbf{[S014]} Akhil, K., et al., \textit{Design of an IOT Based Autonomous Vehicle with the Aid of Computer Vision}, International Journal of Advanced Science and Technology, vol. 29, no 9s, p. 2720–25 (2020).\\[0.1cm]
\textbf{[S015]} Akram R., Markantonakis K., Mayes K., Habachi O., Sauveron D., Steyven A. and Chaumette S., \textit{Security, privacy and safety evaluation of dynamic and static fleets of drones}, AIAA/IEEE Digital Avionics Systems Conference - Proceedings, \textbf{2017-September}, (no pages found) (2017).\\[0.1cm]
\textbf{[S016]} Al Najada H. and Mahgoub I., \textit{Big vehicular traffic Data mining: Towards accident and congestion prevention}, 2016 International Wireless Communications and Mobile Computing Conference, IWCMC 2016, no volume, pp. 256-261 (2016).\\[0.1cm]
\textbf{[S017]} Alam M., Katsikas S., Beltramello O. and Hadjiefthymiades S., \textit{Augmented and virtual reality based monitoring and safety system: A prototype IoT platform}, Journal of Network and Computer Applications, \textbf{89}, pp. 109-119 (2017).\\[0.1cm]
\textbf{[S018]} Alam M., Le Moullec Y., Ahmad R., Magarini M. and Reggiani L., \textit{A Primer On Public Safety Communication in the Context of Terror Attacks: The NATO SPS “COUNTER-TERROR” Project}, NATO Science for Peace and Security Series B: Physics and Biophysics, no volume, pp. 19-34 (2020).\\[0.1cm]
\textbf{[S019]} Alam, Md. F., et al. \textit{Wireless Personnel Safety System (WPSS), a Baseline towards Advance System Architecture}, Proceedings of the 18th Panhellenic Conference on Informatics, Association for Computing Machinery, p. 1–6 (2014).\\[0.1cm]
\textbf{[S020]} Al-Emadi S. and Al-Senaid F., \textit{Drone Detection Approach Based on Radio-Frequency Using Convolutional Neural Network}, 2020 IEEE International Conference on Informatics, IoT, and Enabling Technologies, ICIoT 2020, no volume, pp. 29-34 (2020).\\[0.1cm]
\textbf{[S021]} Alessandri A., Bolla R. and Repetto M., \textit{Identification of parameters for estimation of freeway traffic using information from a mobile cellular network}, IFAC Proceedings Volumes (IFAC-PapersOnline), \textbf{11(PART 1)}, pp. 561-566 (2006).\\[0.1cm]
\textbf{[S022]} Alghamdi S., \textit{Novel path similarity aware clustering and safety message dissemination via mobile gateway selection in cellular 5G-based V2X and D2D communication for urban environment}, Ad Hoc Networks, Volume 103 (2020).\\[0.1cm]
\textbf{[S023]} Alhammadi, S., et al., \textit{An Intelligent Agent based Novel Framework for Building Management System using Artificial Intelligence}, International Journal of Computer Science and Network Security, vol. 20, no 1, p. 167–172 (2020).\\[0.1cm]
\textbf{[S024]} Alheeti K., Gruebler A. and McDonald-Maier K., \textit{Intelligent intrusion detection of grey hole and rushing attacks in self-driving vehicular networks}, Computers, \textbf{5(3)}, (no pages found) (2016).\\[0.1cm]
\textbf{[S025]} Ali A. and Shah S., \textit{VANET Clustering Using Whale optimization Algorithm}, RAEE 2019 - International Symposium on Recent Advances in Electrical Engineering, no volume, (no pages found) (2019).\\[0.1cm]
\textbf{[S026]} Ali E., Hasan M., Hassan R., Saeed R., Hassan M., Islam S., Nafi N. and Bevinakoppa S., \textit{Machine Learning Technologies for Secure Vehicular Communication in Internet of Vehicles: Recent Advances and Applications}, Security and Communication Networks, \textbf{2021}, (no pages found) (2021).\\[0.1cm]
\textbf{[S027]} Ali M., Hu Y., Luong D., Oguntala G., Li J. and Abdo K., \textit{Adversarial attacks on AI based intrusion detection system for heterogeneous wireless communications networks}, AIAA/IEEE Digital Avionics Systems Conference - Proceedings, \textbf{2020-October}, (no pages found) (2020).\\[0.1cm]
\textbf{[S028]} Ali M., Miry A. and Salman T., \textit{IoT Based Water Tank Level Control System Using PLC}, Proceedings of the 2020 International Conference on Computer Science and Software Engineering, CSASE 2020, no volume, pp. 7-12 (2020).\\[0.1cm]
\textbf{[S029]} Ali, M., et al., \textit{CNN Based Subject-Independent Driver Emotion Recognition System Involving Physiological Signals for ADAS}, Advanced Microsystems for Automotive Applications 2016, organizado por Tim Schulze et al., Springer International Publishing, p. 125–38 (2016).\\[0.1cm]
\textbf{[S030]} Alimbuyog R., Cruz J. and Sevilla R., \textit{Development of Motorcycle Data Logging System with Visual Basic Data Simulation for Accident Analysis}, HNICEM 2017 - 9th International Conference on Humanoid, Nanotechnology, Information Technology, Communication and Control, Environment and Management, \textbf{2018-January}, pp. 1-5 (2017).\\[0.1cm]
\textbf{[S031]} Aljeri N. and Boukerche A., \textit{A Performance Evaluation of Time-Series Mobility Prediction for Connected Vehicular Networks}, Q2SWinet 2020 - Proceedings of the 16th ACM Symposium on QoS and Security for Wireless and Mobile Networks, no volume, pp. 127-131 (2020).\\[0.1cm]
\textbf{[S032]} Aljeri N. and Boukerche A., \textit{An efficient handover trigger scheme for vehicular networks using recurrent neural networks}, Q2SWinet 2019 - Proceedings of the 15th ACM International Symposium on QoS and Security for Wireless and Mobile Networks, no volume, pp. 85-91 (2019).\\[0.1cm]
\textbf{[S033]} Aljojo, N., et al., \textit{Alzheimer assistant: a mobile application using Machine Learning}, Revista Român? de Informatic? ?i Automatic?, vol. 30, no 4, p. 7–26. DOI.org (Crossref), (2020).\\[0.1cm]
\textbf{[S034]} Allal A., Mansouri K., Youssfi M. and Qbadou M., \textit{Reliable and cost-effective communication at high seas, for a safe operation of autonomous ship}, Proceedings - 2018 International Conference on Wireless Networks and Mobile Communications, WINCOM 2018, no volume, (no pages found) (2019).\\[0.1cm]
\textbf{[S035]} Al-Mehdhara M. and Ruan N., \textit{MSOM: Efficient Mechanism for Defense against DDoS Attacks in VANET}, Wireless Communications and Mobile Computing, \textbf{2021}, (no pages found) (2021).\\[0.1cm]
\textbf{[S036]} Alnami H., Mahgoub I. and Al-Najada H., \textit{Highway Accident Severity Prediction for Optimal Resource Allocation of Emergency Vehicles and Personnel}, 2021 IEEE 11th Annual Computing and Communication Workshop and Conference, CCWC 2021, no volume, pp. 1231-1238 (2021).\\[0.1cm]
\textbf{[S037]} Alodat M. and Abdullah I., \textit{Elicit the best ways through identify congestion places}, Proceedings - 3rd International Conference on Advanced Computer Science Applications and Technologies, ACSAT 2014, no volume, pp. 102-107 (2014).\\[0.1cm]
\textbf{[S038]} Alonso R., Sittón-Candanedo I., García Ó., Prieto J. and Rodríguez-González S., \textit{An intelligent Edge-IoT platform for monitoring livestock and crops in a dairy farming scenario}, Ad Hoc Networks, \textbf{98}, (no pages found) (2020).\\[0.1cm]
\textbf{[S039]} Al-Refai G. and Rawashdeh O., \textit{Improved candidate generation for pedestrian detection using background modeling in connected vehicles}, International Journal of Advanced Computer Science and Applications, \textbf{11(3)}, pp. 649-660 (2020).\\[0.1cm]
\textbf{[S040]} Alshammari N., Sarker M., Kamruzzaman M., Alruwaili M., Alanazi S., Raihan M. and AlQahtani S., \textit{Technology-driven 5G enabled e-healthcare system during COVID-19 pandemic}, IET Communications, no volume, (no pages found) (2021).\\[0.1cm]
\textbf{[S041]} Alsuhli G., Khattab A. and Fahmy Y., \textit{An evolutionary approach for optimized VANET clustering}, Proceedings of the International Conference on Microelectronics, ICM, \textbf{2019-December}, pp. 70-73 (2019).\\[0.1cm]
\textbf{[S042]} Alvin Yau K., Tan F., Komisarczuk P. and Teal P., \textit{Exploring new and emerging applications of cognitive radio systems: Preliminary insights and framework}, 2011 IEEE Colloquium on Humanities, Science and Engineering, CHUSER 2011, no volume, pp. 153-157 (2011).\\[0.1cm]
\textbf{[S043]} Alzamil M., Albugmi R., Alotaibi S., Alanazi G., Alzubaidi L. and Bashar A., \textit{COMPASS: IPS-based navigation system for visually impaired students}, Proceedings - 2020 IEEE 9th International Conference on Communication Systems and Network Technologies, CSNT 2020, no volume, pp. 161-166 (2020).\\[0.1cm]
\textbf{[S044]} Amoozadeh M., Deng H., Chuah C., Zhang H. and Ghosal D., \textit{Platoon management with cooperative adaptive cruise control enabled by VANET}, Vehicular Communications, \textbf{2(2)}, pp. 110-123 (2015).\\[0.1cm]
\textbf{[S045]} Amudhavel J., Premkumar K., Sai Smrithi R., Banumathi S., Rajaguru D. and Vengattaraman T., \textit{Performance evaluation of dynamic clustering of vehicles in VANET: Challenges and solutions}, ACM International Conference Proceeding Series, \textbf{06-07-March-2015}, (no pages found) (2015).\\[0.1cm]
\textbf{[S046]} Anaum, S., \textit{Smart car for the physically challenged: A unique addition to the Internet of Things}, 2017 International Conference on I-SMAC (IoT in Social, Mobile, Analytics and Cloud) (I-SMAC), p. 502–06 (2017).\\[0.1cm]
\textbf{[S047]} Anbalagan S., Bashir A., Raja G., Dhanasekaran P., Vijayaraghavan G., Tariq U. and Guizani M., \textit{Machine Learning-based Efficient and Secure RSU Placement Mechanism for Software Defined-IoV}, IEEE Internet of Things Journal, no volume, (no pages found) (2021).\\[0.1cm]
\textbf{[S048]} Aoyama M. and Takeichi H., \textit{Adaptive self-organizing overlay network for car-to-car communications}, Proc. 9th ACIS Int. Conf. Software Engineering, Artificial Intelligence, Networking and Parallel/Distributed Computing, SNPD 2008 and 2nd Int. Workshop on Advanced Internet Technology and Applications, no volume, pp. 605-610 (2008).\\[0.1cm]
\textbf{[S049]} Armgarth A., Pantzare S., Arven P., Lassnig R., Jinno H., Gabrielsson E., Kifle Y., Cherian D., Arbring Sjöström T., Berthou G., Dowling J., Someya T., Wikner J., Gustafsson G., Simon D. and Berggren M., \textit{A digital nervous system aiming toward personalized IoT healthcare}, Scientific Reports, \textbf{11(1)}, (no pages found) (2021).\\[0.1cm]
\textbf{[S050]} Aryan A. and Singh S., \textit{Protecting location privacy in augmented reality using k-anonymization and pseudo-id}, 2010 International Conference on Computer and Communication Technology, ICCCT-2010, no volume, pp. 119-124 (2010).\\[0.1cm]
\textbf{[S051]} Arzhmand E. and Rashid H., \textit{Expansion of vehicular cloud services on crossroads using fuzzy logic and Genetic Algorithm}, Proceedings of the 2015 12th International Joint Conference on Computer Science and Software Engineering, JCSSE 2015, no volume, pp. 224-229 (2015).\\[0.1cm]
\textbf{[S052]} Ashraf, J., et al., \textit{Novel Deep Learning-Enabled LSTM Autoencoder Architecture for Discovering Anomalous Events From Intelligent Transportation Systems}, IEEE Transactions on Intelligent Transportation Systems, vol. 22, no 7, p. 4507–18 (2021).\\[0.1cm]
\textbf{[S053]} Ashtaiwi A., \textit{A., ML-Based Localizing and Driving Direction Estimation System for Vehicular Networks}, 2021 International Conference on Artificial Intelligence in Information and Communication (ICAIIC), p. 465–70 (2021).\\[0.1cm]
\textbf{[S054]} Atallah R., Assi C. and Khabbaz M., \textit{Deep reinforcement learning-based scheduling for roadside communication networks}, 2017 15th International Symposium on Modeling and Optimization in Mobile, Ad Hoc, and Wireless Networks, WiOpt 2017, no volume, (no pages found) (2017).\\[0.1cm]
\textbf{[S055]} Balduccini M., Nguyen D. and Regli W., \textit{Coordinating UAVs in dynamic environments by network-aware mission planning}, Proceedings - IEEE Military Communications Conference MILCOM, no volume, pp. 983-988 (2014).\\[0.1cm]
\textbf{[S056]} Balga M. and Ba?çiftçi F., \textit{REST API based image processing interactive person tracking application}, 2nd International Conference on Computer Science and Engineering, UBMK 2017, no volume, pp. 22-27 (2017).\\[0.1cm]
\textbf{[S057]} Bali R., Kumar N. and Rodrigues J., \textit{An intelligent clustering algorithm for VANETs}, 2014 International Conference on Connected Vehicles and Expo, ICCVE 2014 - Proceedings, no volume, pp. 974-979 (2014).\\[0.1cm]
\textbf{[S058]} Bangui H. and Buhnova B., \textit{Recent advances in machine-learning driven intrusion detection in transportation: Survey}, Procedia Computer Science, \textbf{184}, pp. 877-886 (2021).\\[0.1cm]
\textbf{[S059]} Bangui H., Ge M. and Buhnova B., \textit{Improving Big Data Clustering for Jamming Detection in Smart Mobility}, IFIP Advances in Information and Communication Technology, \textbf{580 IFIP}, pp. 78-91 (2020).\\[0.1cm]
\textbf{[S060]} Barmpounakis S., Tsiatsios G., Papadakis M., Mitsianis E., Koursioumpas N. and Alonistioti N., \textit{Collision avoidance in 5G using MEC and NFV: The vulnerable road user safety use case}, Computer Networks, \textbf{172}, (no pages found) (2020).\\[0.1cm]
\textbf{[S061]} Barrachina J., Garrido P., Fogue M., Martinez F., Cano J., Calafate C. and Manzoni P., \textit{CAOVA: A car accident ontology for VANETs}, IEEE Wireless Communications and Networking Conference, WCNC, no volume, pp. 1864-1869 (2012).\\[0.1cm]
\textbf{[S062]} Barrachina J., Garrido P., Fogue M., Martinez F., Cano J., Calafate C. and Manzoni P., \textit{Using evolution strategies to reduce emergency services arrival time in case of accident}, Proceedings - International Conference on Tools with Artificial Intelligence, ICTAI, no volume, pp. 833-840 (2013).\\[0.1cm]
\textbf{[S063]} Basha S. and Shankar T., \textit{Fuzzy logic based forwarder selection for efficient data dissemination in VANETs}, Wireless Networks, \textbf{27(3)}, pp. 2193-2216 (2021).\\[0.1cm]
\textbf{[S064]} Batiha T. and Krömer P., \textit{Design and analysis of efficient neural intrusion detection for wireless sensor networks}, Concurrency Computation, no volume, (no pages found) (2020).\\[0.1cm]
\textbf{[S065]} Bauza R., Gozalvez J. and Sanchez-Soriano J., \textit{Road traffic congestion detection through cooperative Vehicle-to-Vehicle communications}, Proceedings - Conference on Local Computer Networks, LCN, no volume, pp. 606-612 (2010).\\[0.1cm]
\textbf{[S066]} Belobrajdic, Blaze, et al., \textit{Planetary Extravehicular Activity (EVA) Risk Mitigation Strategies for Long-Duration Space Missions}, Npj Microgravity, vol. 7, no 1, p. 1–9 (2021).\\[0.1cm]
\textbf{[S067]} Benamer I., Yahiouche A. and Ghenai A., \textit{Deep Learning Environment Perception and Self-tracking for Autonomous and Connected Vehicles}, Lecture Notes in Computer Science (including subseries Lecture Notes in Artificial Intelligence and Lecture Notes in Bioinformatics), \textbf{12629 LNCS}, pp. 305-319 (2021).\\[0.1cm]
\textbf{[S068]} Bensabera A., Diaz C. and Lahrounic Y.,\textit{Design and Modeling an Adaptive Neuro-Fuzzy Inference System (ANFIS) for the Prediction of a Security Index in VANET}, Journal of Computational Science, vol. 47, p. 101234 (2020).\\[0.1cm]
\textbf{[S069]} Bentjen K., Graham S. and Nykl S., \textit{Modelling misbehaviour in automated vehicle intersections in a synthetic environment}, Proceedings of the 13th International Conference on Cyber Warfare and Security, ICCWS 2018, \textbf{2018-March}, pp. 584-593 (2018).\\[0.1cm]
\textbf{[S070]} Bisio I., Garibotto C., Lavagetto F., Sciarrone A. and Zappatore S., \textit{Improving WiFi Statistical Fingerprint-Based Detection Techniques Against UAV Stealth Attacks}, 2018 IEEE Global Communications Conference, GLOBECOM 2018 - Proceedings, no volume, (no pages found) (2018).\\[0.1cm]
\textbf{[S071]} Blum J., Eskandarian A. and Hoffman L., \textit{Mobility management in IVC networks}, IEEE Intelligent Vehicles Symposium, Proceedings, no volume, pp. 150-155 (2003).\\[0.1cm]
\textbf{[S072]} Böckle M., Klingegard M., Habibovic A. and Bout M., \textit{SAV2P - Exploring the impact of an interface for shared automated vehicles on pedestrians' experience}, AutomotiveUI 2017 - 9th International ACM Conference on Automotive User Interfaces and Interactive Vehicular Applications, Adjunct Proceedings, no volume, pp. 136-140 (2017).\\[0.1cm]
\textbf{[S073]} Bononi L., Di Felice M. and Pizzi S., \textit{DBA-MAC: Dynamic backbone-assisted medium access control protocol for efficient broadcast in vanets}, Journal of Interconnection Networks, \textbf{10(4)}, pp. 321-344 (2009).\\[0.1cm]
\textbf{[S074]} Bosnak M. and Skrjanc I., \textit{Efficient time-to-collision estimation for a braking supervision system with LIDAR}, 2017 3rd IEEE International Conference on Cybernetics, CYBCONF 2017 - Proceedings, no volume, (no pages found) (2017).\\[0.1cm]
\textbf{[S075]} Brindha Devi V., Sindhuja S., Shanthini S. and Hemalatha M., \textit{A virtual keyboard security system for automated teller machine}, International Journal of Engineering and Technology(UAE), \textbf{7(2)}, pp. 59-63 (2018).\\[0.1cm]
\textbf{[S076]} Bunyakitanon M., da Silva A., Vasilakos X., Nejabati R. and Simeonidou D., \textit{AUTO-3P: An autonomous VNF performance prediction \& placement framework based on machine learning}, Computer Networks, \textbf{181}, (no pages found) (2020).\\[0.1cm]
\textbf{[S077]} Bustamante C., Mateu E., Hernández J. and Arrúe Á., \textit{Intelligent human-machine interface for wireless signposts}, SAE Technical Papers, \textbf{13}, (no pages found) (2013).\\[0.1cm]
\textbf{[S078]} Caballeros Morales M., Hong C. and Bang Y., \textit{An Adaptable mobility-aware clustering algorithm in vehicular networks}, APNOMS 2011 - 13th Asia-Pacific Network Operations and Management Symposium: Managing Clouds, Smart Networks and Services, Final Program, no volume, (no pages found) (2011).\\[0.1cm]
\textbf{[S079]} Calvaresi D. and Calbimonte J., \textit{Real-time compliant stream processing agents for physical rehabilitation}, Sensors (Switzerland), \textbf{20(3)}, (no pages found) (2020).\\[0.1cm]
\textbf{[S080]} Calvi A., D'Amico F., Ferrante C. and Bianchini Ciampoli L., \textit{Evaluation of augmented reality cues to improve the safety of left-turn maneuvers in a connected environment: A driving simulator study}, Accident Analysis and Prevention, \textbf{148}, (no pages found) (2020).\\[0.1cm]
\textbf{[S081]} Canonico R., Marrone S., Nardone R. and Vittorini V., \textit{A Framework to Evaluate 5G Networks for Smart and Fail-Safe Communications in ERTMS/ETCS}, Lecture Notes in Computer Science (including subseries Lecture Notes in Artificial Intelligence and Lecture Notes in Bioinformatics), \textbf{10598 LNCS}, pp. 34-50 (2017).\\[0.1cm]
\textbf{[S082]} Cao J., You Y., Ning Y. and Zhou W., \textit{Change Detection Network of Nearshore Ships for Multi-Temporal Optical Remote Sensing Images}, International Geoscience and Remote Sensing Symposium (IGARSS), no volume, pp. 2531-2534 (2020).\\[0.1cm]
\textbf{[S083]} Cao W., Zhang J., Cai C., Chen Q., Zhao Y., Lou Y., Jiang W. and Gui G., \textit{CNN-based intelligent safety surveillance in green IoT applications}, China Communications, \textbf{18(1)}, pp. 108-119 (2021).\\[0.1cm]
\textbf{[S084]} Cao W., Zou Y., Luo M., Zhang P., Wang W. and Huang W., \textit{Deep Discriminant Learning-based Asphalt Road Cracks Detection via Wireless Camera Network}, 2019 Computing, Communications and IoT Applications, ComComAp 2019, no volume, pp. 53-58 (2019).\\[0.1cm]
\textbf{[S085]} Cardenas, L., et al., \textit{A Multimetric Predictive ANN-Based Routing Protocol for Vehicular Ad Hoc Networks}, IEEE Access, vol. 9, p. 86037–53 (2021).\\[0.1cm]
\textbf{[S086]} Cecilio J., Martins P., Costa J. and Furtado P., \textit{State machine model-based middleware for control and processing in industrial wireless sensor and actuator networks}, IEEE International Conference on Industrial Informatics (INDIN), no volume, pp. 1142-1147 (2012).\\[0.1cm]
\textbf{[S087]} Centelles D., Soriano A., Marin R. and Sanz P., \textit{Wireless HROV Control with Compressed Visual Feedback Using Acoustic and RF Links}, Journal of Intelligent and Robotic Systems: Theory and Applications, \textbf{99(3-4)}, pp. 713-728 (2020).\\[0.1cm]
\textbf{[S088]} Chai R., Ge X. and Chen Q., \textit{Adaptive K-Harmonic Means clustering algorithm for VANETs}, 14th International Symposium on Communications and Information Technologies, ISCIT 2014, no volume, pp. 233-237 (2015).\\[0.1cm]
\textbf{[S089]} Chai R., Yang B., Li L., Sun X. and Chen Q., \textit{Clustering-based data transmission algorithms for VANET}, 2013 International Conference on Wireless Communications and Signal Processing, WCSP 2013, no volume, (no pages found) (2013).\\[0.1cm]
\textbf{[S090]} Chan C. and Gupta S., \textit{Utilizing machine-to-machine communication for speeding alerts and enforcement augmentation}, 20th ITS World Congress Tokyo 2013, no volume, (no pages found) (2013).\\[0.1cm]
\textbf{[S091]} Chanine, M., and Walke. B., \textit{Road traffic informatic (RTI) systems using short-range mobile communications}, [1991 Proceedings] 41st IEEE Vehicular Technology Conference, IEEE, p. 788–92 (1991).\\[0.1cm]
\textbf{[S092]} Chauhan A. and Zaveri M., \textit{Segmentation based data dissemination in inter vehicular communication}, Proceedings - 4th International Conference on Computational Intelligence and Communication Networks, CICN 2012, no volume, pp. 179-185 (2012).\\[0.1cm]
\textbf{[S093]} Chbib F., Fahs W., Haydar J., Khoukhi L. and Khatoun R., \textit{IEEE 802.11p performance enhancement based on Markov chain and neural networks for safety applications}, Annales des Telecommunications/Annals of Telecommunications, no volume, (no pages found) (2021).\\[0.1cm]
\textbf{[S094]} Chellaswamy C., Ganesh Babu R., Saravanan M., Abirami M., Boosuphasri R. and Balaji M., \textit{Machine learning based condition recognition system for bikers}, 2020 7th International Conference on Smart Structures and Systems, ICSSS 2020, no volume, (no pages found) (2020).\\[0.1cm]
\textbf{[S095]} Chen C., Liu L., Qiu T., Ren Z., Hu J. and Ti F., \textit{Driver's intention identification and risk evaluation at intersections in the internet of vehicles}, IEEE Internet of Things Journal, \textbf{5(3)}, pp. 1575-1587 (2018).\\[0.1cm]
\textbf{[S096]} Chen M., Liu K., Ma J., Zeng X., Dong Z., Tong G. and Liu C., \textit{MoLoc: Unsupervised Fingerprint Roaming for Device-Free Indoor Localization in a Mobile Ship Environment}, IEEE Internet of Things Journal, \textbf{7(12)}, pp. 11851-11862 (2020).\\[0.1cm]
\textbf{[S097]} Chen T., Zhang Y., Tuo Y. and Wang W., \textit{CgNet:Predicting Urban Congregations from Spatio-Temporal Data Using Deep Neural Networks}, 2020 IEEE Global Communications Conference, GLOBECOM 2020 - Proceedings, no volume, (no pages found) (2020).\\[0.1cm]
\textbf{[S098]} Chen X., Leng S. and Wu F., \textit{Reinforcement Learning Based Safety Message Broadcasting in Vehicular Networks}, 2018 10th International Conference on Wireless Communications and Signal Processing, WCSP 2018, no volume, (no pages found) (2018).\\[0.1cm]
\textbf{[S099]} Chen X., Leng S., He J. and Zhou L., \textit{Deep Learning Based Intelligent Inter-Vehicle Distance Control for 6G-Enabled Cooperative Autonomous Driving}, IEEE Internet of Things Journal, no volume, (no pages found) (2020).\\[0.1cm]
\textbf{[S100]} Chen Y., Cai X., Gao M., Wang X., Zhu L. and Li C., \textit{Dynamic overlay-based scheme for video delivery over VANETs}, IEEE Vehicular Technology Conference, no volume, (no pages found) (2014).\\[0.1cm]
\textbf{[S101]} Chen Y., Lin X., Khan T. and Mozaffari M., \textit{Efficient Drone Mobility Support Using Reinforcement Learning}, IEEE Wireless Communications and Networking Conference, WCNC, \textbf{2020-May}, (no pages found) (2020).\\[0.1cm]
\textbf{[S102]} Chen, L. and Chen., H., \textit{Driver Behavior Monitoring and Warning With Dangerous Driving Detection Based on the Internet of Vehicles}, IEEE Transactions on Intelligent Transportation Systems, p. 1–10 (2020).\\[0.1cm]
\textbf{[S103]} Cheng, S., \textit{An Intelligently Remote Infant Monitoring System Based on RFID}, Intelligent Information and Database Systems, organizado por Jeng-Shyang Pan et al., Springer, p. 246–54 (2012).\\[0.1cm]
\textbf{[S104]} Cheng, X., et al. \textit{5G-Enabled Cooperative Intelligent Vehicular (5GenCIV) Framework: When Benz Meets Marconi}, IEEE Intelligent Systems, vol. 32, no 3, p. 53–59 (2017).\\[0.1cm]
\textbf{[S105]} Chlamtac I., Fumagalli A., Montgomery D., Brooks R., Cerutti I., Tacca M. and Valcarenghi L., \textit{CATO: CAD tool for design, simulation, and optimization of optical telecommunications networks}, Proceedings of SPIE - The International Society for Optical Engineering, \textbf{3843}, pp. 148-159 (1999).\\[0.1cm]
\textbf{[S106]} Choe C., Ahn J., Choi J., Park D., Kim M. and Ahn S., \textit{A Robust Channel Access Using Cooperative Reinforcement Learning for Congested Vehicular Networks}, IEEE Access, \textbf{8}, pp. 135540-135557 (2020).\\[0.1cm]
\textbf{[S107]} Chou W., Wang T. and Zhang Y., \textit{Augmented reality based preoperative planning for robot assisted tele-neurosurgery}, Conference Proceedings - IEEE International Conference on Systems, Man and Cybernetics, \textbf{3}, pp. 2901-2906 (2004).\\[0.1cm]
\textbf{[S108]} Chu T. and Kalabic U., \textit{Model-based deep reinforcement learning for CACC in mixed-autonomy vehicle platoon}, Proceedings of the IEEE Conference on Decision and Control, \textbf{2019-December}, pp. 4079-4084 (2019).\\[0.1cm]
\textbf{[S109]} Clement A., Lançon H., Cachot E. and Servant C., \textit{Long term monitoring of Millau viaduct}, IABSE Symposium, Nantes 2018: Tomorrow's Megastructures, no volume, pp. S24-103-S24-110 (2018).\\[0.1cm]
\textbf{[S110]} Conceicao H., Ferreira M. and Steenkiste P., \textit{Virtual traffic lights in partial deployment scenarios}, IEEE Intelligent Vehicles Symposium, Proceedings, no volume, pp. 988-993 (2013).\\[0.1cm]
\textbf{[S111]} Coronado E., Cebrian-Marquez G. and Riggio R., \textit{Enabling computation offloading for autonomous and assisted driving in 5G networks}, 2019 IEEE Global Communications Conference, GLOBECOM 2019 - Proceedings, no volume, (no pages found) (2019).\\[0.1cm]
\textbf{[S112]} Cosovanu L., Zadobrischi E., DImian M. and Plascencia E., \textit{Unified road infrastructure safety system using visible light communication}, 2020 28th Telecommunications Forum, TELFOR 2020 - Proceedings, no volume, (no pages found) (2020).\\[0.1cm]
\textbf{[S113]} Cretu A., Radu D., Nandor K., Avram C., Domuta C. and Astilean A., \textit{Neural Networks Fire Detection in a Public Safety Fog System}, 2020 22nd IEEE International Conference on Automation, Quality and Testing, Robotics - THETA, AQTR 2020 - Proceedings, no volume, (no pages found) (2020).\\[0.1cm]
\textbf{[S114]} Dai S., Chi Y., Qiao Z. and Ji X., \textit{A microgrid controller security monitoring model based on message flow}, 2020 IEEE 4th Conference on Energy Internet and Energy System Integration: Connecting the Grids Towards a Low-Carbon High-Efficiency Energy System, EI2 2020, no volume, pp. 3822-3826 (2020).\\[0.1cm]
\textbf{[S115]} Das P. and Sengupta S., \textit{Proposing the systems to provide protection of vehicles against theft and accident}, 2016 IEEE International Conference on Recent Trends in Electronics, Information and Communication Technology, RTEICT 2016 - Proceedings, no volume, pp. 1681-1685 (2017).\\[0.1cm]
\textbf{[S116]} Dawande, J. R., et al, \textit{Enhanced Distributed Multi-Hop Clustering Algorithm for VANETs Based on Neighborhood Follow (EDMCNF) Collaborated with Road Side Units}, 2015 International Conference on Computational Intelligence and Communication Networks (CICN), IEEE, p. 106–13 (2015).\\[0.1cm]
\textbf{[S117]} Dayekh S., Affes S., Kandil N. and Nerguizian C., \textit{Cooperative localization in mines using fingerprinting and neural networks}, IEEE Wireless Communications and Networking Conference, WCNC, no volume, (no pages found) (2010).\\[0.1cm]
\textbf{[S118]} Dayekh S., Affes S., Kandil N. and Nerguizian C., \textit{Cost-effective localization in underground mines using new SIMO/MIMO-like fingerprints and artificial neural networks}, 2014 IEEE International Conference on Communications Workshops, ICC 2014, no volume, pp. 730-735 (2014).\\[0.1cm]
\textbf{[S119]} Dayekh S., Affes S., Kandil N. and Nerguizian C., \textit{Radio-localization in underground narrow-vein mines using neural networks with in-built tracking and time diversity}, 2011 IEEE Wireless Communications and Networking Conference, WCNC 2011, no volume, pp. 1788-1793 (2011).\\[0.1cm]
\textbf{[S120]} Dayekh S., Affes S., Kandil N. and Nerguizian C., \textit{Smart spatio-temporal fingerprinting for cooperative ANN-based wireless localization in underground narrow-vein mines}, ACM International Conference Proceeding Series, no volume, (no pages found) (2011).\\[0.1cm]
\textbf{[S121]} De Felice M., Calcagni I., Pesci F., Cuomo F. and Baiocchi A., \textit{Self-Healing Infotainment and Safety Application for VANET dissemination}, 2015 IEEE International Conference on Communication Workshop, ICCW 2015, no volume, pp. 2495-2500 (2015).\\[0.1cm]
\textbf{[S122]} de Souza P., Rubin F., Hohemberger R., Ferreto T., Lorenzon A., Luizelli M. and Rossi F., \textit{Detecting abnormal sensors via machine learning: An IoT farming WSN-based architecture case study}, Measurement: Journal of the International Measurement Confederation, \textbf{164}, (no pages found) (2020).\\[0.1cm]
\textbf{[S123]} Dechouniotis D., Athanasopoulos N., Leivadeas A., Mitton N., Jungers R. and Papavassiliou S., \textit{Edge computing resource allocation for dynamic networks: The DRUID-NET vision and perspective}, Sensors (Switzerland), \textbf{20(8)}, (no pages found) (2020).\\[0.1cm]
\textbf{[S124]} Desai K., Mane P., Dsilva M., Zare A., Shingala P. and Ambawade D., \textit{A novel machine learning based wearable belt for fall detection}, 2020 IEEE International Conference on Computing, Power and Communication Technologies, GUCON 2020, no volume, pp. 502-505 (2020).\\[0.1cm]
\textbf{[S125]} Desai N. and Punnekkat S., \textit{Enhancing Fault Detection in Time Sensitive Networks using Machine Learning}, 2020 International Conference on COMmunication Systems and NETworkS, COMSNETS 2020, no volume, pp. 714-719 (2020).\\[0.1cm]
\textbf{[S126]} Dey P., Kumar C., Mitra M., Mishra R., Chaulya S., Prasad G., Mandal S. and Banerjee G., \textit{Deep convolutional neural network based secure wireless voice communication for underground mines}, Journal of Ambient Intelligence and Humanized Computing, no volume, (no pages found) (2021).\\[0.1cm]
\textbf{[S127]} Di Felice M., Bedogni L. and Bononi L., \textit{Group communication on highways: An evaluation study of geocast protocols and applications}, Ad Hoc Networks, \textbf{11(3)}, pp. 818-832 (2013).\\[0.1cm]
\textbf{[S128]} Ding H., Wu H., Dong L. and Li Z., \textit{Vehicle intersection collision monitoring algorithm based on VANETs and uncertain trajectories}, Proceedings of 2018 16th International Conference on Intelligent Transport System Telecommunications, ITST 2018, no volume, (no pages found) (2018).\\[0.1cm]
\textbf{[S129]} Ding, D. , \textit{Deformation Detection Model of High-Rise Building Foundation Pit Support Structure Based on Neural Network and Wireless Communication}, Security and Communication Networks, vol. 2021, (2021).\\[0.1cm]
\textbf{[S130]} Dini P. and Saponara S., \textit{Analysis, design, and comparison of machine-learning techniques for networking intrusion detection}, Designs, \textbf{5(1)}, pp. 1-22 (2021).\\[0.1cm]
\textbf{[S131]} Dogru N. and Subasi A., \textit{Traffic accident detection using random forest classifier}, 2018 15th Learning and Technology Conference, L and T 2018, no volume, pp. 40-45 (2018).\\[0.1cm]
\textbf{[S132]} Dong L., Sun D., Han G., Li X., Hu Q. and Shu L., \textit{Velocity-Free Localization of Autonomous Driverless Vehicles in Underground Intelligent Mines}, IEEE Transactions on Vehicular Technology, \textbf{69(9)}, pp. 9292-9303 (2020).\\[0.1cm]
\textbf{[S133]} Dong W., Chen Y., Huang X. and Yang Q., \textit{Hybrid Mac Protocol Based on Security in Clustering Topology}, Advances in Intelligent Systems and Computing, \textbf{1283}, pp. 439-444 (2021).\\[0.1cm]
\textbf{[S134]} Doshi K., Yilmaz Y. and Uludag S., \textit{Timely Detection and Mitigation of Stealthy DDoS Attacks via IoT Networks}, IEEE Transactions on Dependable and Secure Computing, no volume, (no pages found) (2021).\\[0.1cm]
\textbf{[S135]} Dror E., Avin C. and Lotker Z., \textit{Fast randomized algorithm for hierarchical clustering in Vehicular Ad-Hoc Networks}, 2011 the 10th IFIP Annual Mediterranean Ad Hoc Networking Workshop, Med-Hoc-Net'2011, no volume, pp. 1-8 (2011).\\[0.1cm]
\textbf{[S136]} Du X., Zhang H., Van Nguyen H. and Han Z., \textit{Stacked LSTM deep learning model for traffic prediction in vehicle-to-vehicle communication}, IEEE Vehicular Technology Conference, \textbf{2017-September}, pp. 1-5 (2018).\\[0.1cm]
\textbf{[S137]} Duo R., Nie X., Yang N., Yue C. and Wang Y., \textit{Anomaly Detection and Attack Classification for Train Real-Time Ethernet}, IEEE Access, \textbf{9}, pp. 22528-22541 (2021).\\[0.1cm]
\textbf{[S138]} Eigner R. and Lutz G., \textit{Collision avoidance in VANETs an application for ontological context models}, 6th Annual IEEE International Conference on Pervasive Computing and Communications, PerCom 2008, no volume, pp. 412-416 (2008).\\[0.1cm]
\textbf{[S139]} Eiter T., Parreira J. and Schneider P., \textit{Detecting mobility patterns using spatial query answering over streams}, CEUR Workshop Proceedings, \textbf{1936}, pp. 17-32 (2017).\\[0.1cm]
\textbf{[S140]} Eiter T., Parreira J. and Schneider P., \textit{Spatial ontology-mediated query answering over mobility streams}, Lecture Notes in Computer Science (including subseries Lecture Notes in Artificial Intelligence and Lecture Notes in Bioinformatics), \textbf{10249 LNCS}, pp. 219-237 (2017).\\[0.1cm]
\textbf{[S141]} Eren H., Pakka H., Alghamdi A. and Yue Y., \textit{Instrumentation for safe vehicular flow in intelligent traffic control systems using wireless networks}, Conference Record - IEEE Instrumentation and Measurement Technology Conference, no volume, pp. 1301-1305 (2013).\\[0.1cm]
\textbf{[S142]} Erturk M., Jamal H. and Matolak D., \textit{Potential Future Aviation Communication Technologies}, AIAA/IEEE Digital Avionics Systems Conference - Proceedings, \textbf{2019-September}, (no pages found) (2019).\\[0.1cm]
\textbf{[S143]} Ezema L. and Ani C., \textit{Artificial Neural Network Approach to Mobile Location Estimation in GSM Network}, International Journal of Electronics and Telecommunications, \textbf{63(1)}, pp. 39-44 (2017).\\[0.1cm]
\textbf{[S144]} Eziama E., Tepe K., Balador A., Nwizege K. and Jaimes L., \textit{Malicious Node Detection in Vehicular Ad-Hoc Network Using Machine Learning and Deep Learning}, 2018 IEEE Globecom Workshops, GC Wkshps 2018 - Proceedings, no volume, (no pages found) (2019).\\[0.1cm]
\textbf{[S145]} Fabian P., Rachedi A. and Guéguen C., \textit{Selection of relays based on the classification of mobility-type and localized network metrics in the Internet of Vehicles}, Transactions on Emerging Telecommunications Technologies, \textbf{32(4)}, (no pages found) (2021).\\[0.1cm]
\textbf{[S146]} Falahatraftar F., Pierre S. and Chamberland S., \textit{A multiple linear regression model for predicting congestion in heterogeneous vehicular networks}, International Conference on Wireless and Mobile Computing, Networking and Communications, \textbf{2020-October}, (no pages found) (2020).\\[0.1cm]
\textbf{[S147]} Fan X., Su Y., Dong T., Jie Y., Zhang Y., Ren F., Niu J., Zhang J. and Wang J., \textit{Enhancing the Credibility of the Optical Performance Monitor with Adversarial Training}, IEEE Access, \textbf{8}, pp. 75682-75690 (2020).\\[0.1cm]
\textbf{[S148]} Fang L., Ge C., Zu G., Wang X., Ding W., Xiao C. and Zhao L., \textit{A mobile edge computing architecture for safety in mining industry}, Proceedings - 2019 IEEE SmartWorld, Ubiquitous Intelligence and Computing, Advanced and Trusted Computing, Scalable Computing and Communications, Internet of People and Smart City Innovation, SmartWorld/UIC/ATC/SCALCOM/IOP/SCI 2019, no volume, pp. 1494-1498 (2019).\\[0.1cm]
\textbf{[S149]} Faron C., Ghidini C., Eiter T., Ichise R., Parreira J., Schneider P. and Zhao L., \textit{Deploying spatial-stream query answering in C-ITS scenarios}, Semantic Web, \textbf{12(1)}, pp. 41-77 (2020).\\[0.1cm]
\textbf{[S150]} Farroha B., Farroha D., Cook J. and Dutta A., \textit{Exploring the security and operational aspects of the 5th generation wireless communication system}, Proceedings of SPIE - The International Society for Optical Engineering, \textbf{11015}, (no pages found) (2019).\\[0.1cm]
\textbf{[S151]} Fatemidokht H. and Kuchaki Rafsanjani M., \textit{F-Ant: an effective routing protocol for ant colony optimization based on fuzzy logic in vehicular ad hoc networks}, Neural Computing and Applications, \textbf{29(11)}, pp. 1127-1137 (2018).\\[0.1cm]
\textbf{[S152]} Fatemidokht H. and Kuchaki Rafsanjani M., \textit{QMM-VANET: An efficient clustering algorithm based on QoS and monitoring of malicious vehicles in vehicular ad hoc networks}, Journal of Systems and Software, \textbf{165}, (no pages found) (2020).\\[0.1cm]
\textbf{[S153]} Faust O., Lei N., Chew E., Ciaccio E. and Acharya U., \textit{A smart service platform for cost efficient cardiac health monitoring}, International Journal of Environmental Research and Public Health, \textbf{17(17)}, pp. 1-18 (2020).\\[0.1cm]
\textbf{[S154]} Feki S., Belghith A. and Zarai F., \textit{A reinforcement learning-based radio resource management Algorithm for D2D-based V2V communication}, 2019 15th International Wireless Communications and Mobile Computing Conference, IWCMC 2019, no volume, pp. 1367-1372 (2019).\\[0.1cm]
\textbf{[S155]} Fernandez S. and Ito T., \textit{Semantic integration of sensor data with SSN ontology in a multi-agent architecture for intelligent transportation systems}, IEICE Transactions on Information and Systems, \textbf{E100D(12)}, pp. 2915-2922 (2017).\\[0.1cm]
\textbf{[S156]} Fink G. and McKenzie P., \textit{Helping IT and OT Defenders Collaborate}, Proceedings - 2018 IEEE International Conference on Industrial Internet, ICII 2018, no volume, pp. 188-194 (2018).\\[0.1cm]
\textbf{[S157]} Flushing E., Gambardella L. and Di Caro G., \textit{On Using Mobile Robotic Relays for Adaptive Communication in Search and Rescue Missions}, SSRR 2016 - International Symposium on Safety, Security and Rescue Robotics, no volume, pp. 370-377 (2016).\\[0.1cm]
\textbf{[S158]} Francois F., Abdelrahman O. and Gelenbe E., \textit{Impact of signaling storms on energy consumption and latency of LTE user equipment}, Proceedings - 2015 IEEE 17th International Conference on High Performance Computing and Communications, 2015 IEEE 7th International Symposium on Cyberspace Safety and Security and 2015 IEEE 12th International Conference on Embedded Software and Systems, HPCC-CSS-ICESS 2015, no volume, pp. 1248-1255 (2015).\\[0.1cm]
\textbf{[S159]} Freysinger W., Truppe M., Gunkel A. and Thumfart W., \textit{Stereotactic telepresence in ear, nose, and throat surgery Stereotaktische telepräsenz in der hals-, nasen- und ohrenchirurgie}, HNO, \textbf{50(5)}, pp. 424-432 (2002).\\[0.1cm]
\textbf{[S160]} Fritz, H., \textit{Longitudinal and lateral control of heavy duty trucks for automated vehicle following in mixed traffic: experimental results from the CHAUFFEUR project}, Proceedings of the 1999 IEEE International Conference on Control Applications (Cat. No.99CH36328), vol. 2, p. 1348–52 vol. 2 (1999).\\[0.1cm]
\textbf{[S161]} Fu L., Zhang W., Tan X. and Zhu H., \textit{An Algorithm for Detection of Traffic Attribute Exceptions Based on Cluster Algorithm in Industrial Internet of Things}, IEEE Access, \textbf{9}, pp. 53370-53378 (2021).\\[0.1cm]
\textbf{[S162]} Fu Y., Li C., Yu F., Luan T. and Zhang Y., \textit{An Autonomous Lane-Changing System with Knowledge Accumulation and Transfer Assisted by Vehicular Blockchain}, IEEE Internet of Things Journal, \textbf{7(11)}, pp. 11123-11136 (2020).\\[0.1cm]
\textbf{[S163]} Gaikwad T., Rabinowitz A., Motallebiaraghi F., Bradley T., Asher Z., Fong A. and Meyer R., \textit{Vehicle Velocity Prediction Using Artificial Neural Network and Effect of Real World Signals on Prediction Window}, SAE Technical Papers, \textbf{2020-April(April)}, (no pages found) (2020).\\[0.1cm]
\textbf{[S164]} Gamwarige S. and Kulasekere C., \textit{An energy efficient distributed clustering algorithm for ad hoc deployed wireless sensor networks in building monitoring applications}, Electronic Journal of Structural Engineering, \textbf{9}, pp. 11-27 (2009).\\[0.1cm]
\textbf{[S165]} Gao B., Bu B., Zhang W. and Li X., \textit{An Intrusion Detection Method Based on Machine Learning and State Observer for Train-Ground Communication Systems}, IEEE Transactions on Intelligent Transportation Systems, no volume, (no pages found) (2021).\\[0.1cm]
\textbf{[S166]} Gao Z., Eisen M. and Ribeiro A., \textit{Optimal WDM power allocation via deep learning for radio on free space optics systems}, 2019 IEEE Global Communications Conference, GLOBECOM 2019 - Proceedings, no volume, (no pages found) (2019).\\[0.1cm]
\textbf{[S167]} García-Sánchez D., Iglesias F., Diez J., Piñero I., Fernández-Navamuel A., Sánchez D. and Jiménez-Fernandez J., \textit{Gradient-Boosting Applied for Proactive Maintenance System in a Railway Bridge}, None, \textbf{127}, pp. 236-244 (2021).\\[0.1cm]
\textbf{[S168]} Garip M., Lin J., Reiher P. and Gerla M., \textit{SHIELDNET: An Adaptive Detection Mechanism against Vehicular Botnets in VANETs}, IEEE Vehicular Networking Conference, VNC, \textbf{2019-December}, (no pages found) (2019).\\[0.1cm]
\textbf{[S169]} Gasmi R. and Aliouat M., \textit{A Weight Based Clustering Algorithm for Internet of Vehicles}, Automatic Control and Computer Sciences, \textbf{54(6)}, pp. 493-500 (2020).\\[0.1cm]
\textbf{[S170]} Gawad S., El Mougy A. and El-Meligy M., \textit{Dynamic mapping of Road conditions using smartphone sensors and machine learning techniques}, IEEE Vehicular Technology Conference, \textbf{0}, (no pages found) (2016).\\[0.1cm]
\textbf{[S171]} Ge F., Chen Q., Wang Y., Bostian C., Rondeau T. and Le B., \textit{Cognitive radio: From spectrum sharing to adaptive learning and reconfiguration}, IEEE Aerospace Conference Proceedings, no volume, (no pages found) (2008).\\[0.1cm]
\textbf{[S172]} Ge X., Gao Q. and Quan X., \textit{A novel clustering algorithm based on mobility for VANET}, International Conference on Communication Technology Proceedings, ICCT, \textbf{2019-October}, pp. 473-477 (2019).\\[0.1cm]
\textbf{[S173]} Geacar C., Ion D. and Stoica A., \textit{Language modeling in air traffic control}, UPB Scientific Bulletin, Series D: Mechanical Engineering, \textbf{74(4)}, pp. 27-36 (2012).\\[0.1cm]
\textbf{[S174]} Ghaleb F., Al-Rimy B., Almalawi A., Ali A., Zainal A., Rassam M., Shaid S. and Maarof M., \textit{Deep Kalman Neuro Fuzzy-Based Adaptive Broadcasting Scheme for Vehicular Ad Hoc Network: A Context-Aware Approach}, IEEE Access, \textbf{8}, pp. 217744-217761 (2020).\\[0.1cm]
\textbf{[S175]} Ghaleb F., Zainal A., Rassam M. and Mohammed F., \textit{An effective misbehavior detection model using artificial neural network for vehicular ad hoc network applications}, 2017 IEEE Conference on Applications, Information and Network Security, AINS 2017, \textbf{2018-January}, pp. 13-18 (2017).\\[0.1cm]
\textbf{[S176]} Ghandour A., Fawaz K., Artail H., Di Felice M. and Bononi L., \textit{Improving vehicular safety message delivery through the implementation of a cognitive vehicular network}, Ad Hoc Networks, \textbf{11(8)}, pp. 2408-2422 (2013).\\[0.1cm]
\textbf{[S177]} Ghosh A., Vardhan V., Mapp G., Gemikonakli O. and Loo J., \textit{Providing ubiquitous communication using road-side units in VANET systems: Unveiling the challenges}, 2013 13th International Conference on ITS Telecommunications, ITST 2013, no volume, pp. 74-79 (2013).\\[0.1cm]
\textbf{[S178]} Goli, . A., et al., \textit{Vehicle Trajectory Prediction with Gaussian Process Regression in Connected Vehicle Environment$\star$}, 2018 IEEE Intelligent Vehicles Symposium (IV), p. 550–55 (2018).\\[0.1cm]
\textbf{[S179]} Gopinath, A. J. and Nithya., A., \textit{Fuzzy Logic based Cooperative Rebroadcasting (CoRe) Algorithm for Multi-hop Vehicular Network}, 2020 IEEE International Conference on Electronics, Computing and Communication Technologies (CONECCT), IEEE, p. 1–6 (2020).\\[0.1cm]
\textbf{[S180]} Gorkin R., Adams K., Berryman M., Aubin S., Li W., Davis A. and Barthelemy J., \textit{Sharkeye: Real-time autonomous personal shark alerting via aerial surveillance}, Drones, \textbf{4(2)}, pp. 1-17 (2020).\\[0.1cm]
\textbf{[S181]} Grabowski M., Rowen A. and Rancy J., \textit{Evaluation of wearable immersive augmented reality technology in safety-critical systems}, Safety Science, \textbf{103}, pp. 23-32 (2018).\\[0.1cm]
\textbf{[S182]} Guo J., Li X., Liu Z., Ma J., Yang C., Zhang J. and Wu D., \textit{TROVE: A Context-Awareness Trust Model for VANETs Using Reinforcement Learning}, IEEE Internet of Things Journal, \textbf{7(7)}, pp. 6647-6662 (2020).\\[0.1cm]
\textbf{[S183]} Guo L. and Jia Y., \textit{Anticipative and Predictive Control of Automated Vehicles in Communication-Constrained Connected Mixed Traffic}, IEEE Transactions on Intelligent Transportation Systems, no volume, (no pages found) (2021).\\[0.1cm]
\textbf{[S184]} Guo, X., et al., \textit{Value-Decomposition Networks based Distributed Interference Control in Multi -platoon Groupcast} 2020 IEEE 18th International Conference on Industrial Informatics (INDIN), IEEE, p. 348–53 (2020).\\[0.1cm]
\textbf{[S185]} Gupta S., Khan J. and Ngo D., \textit{A D2D multicast network architecture for vehicular communications}, IEEE Vehicular Technology Conference, \textbf{2019-April}, (no pages found) (2019).\\[0.1cm]
\textbf{[S186]} Gyawali S., Qian Y. and Hu R., \textit{A Privacy-Preserving Misbehavior Detection System in Vehicular Communication Networks}, IEEE Transactions on Vehicular Technology, no volume, (no pages found) (2021).\\[0.1cm]
\textbf{[S187]} Habibi G. and How J., \textit{Human Trajectory Prediction Using Similarity-Based Multi-Model Fusion}, IEEE Robotics and Automation Letters, \textbf{6(2)}, pp. 715-722 (2021).\\[0.1cm]
\textbf{[S188]} Hadded M., Muhlethaler P., Laouiti A. and Azzouz Saidane L., \textit{A novel angle-based clustering algorithm for vehicular ad hoc networks}, Advances in Intelligent Systems and Computing, \textbf{548}, pp. 27-38 (2017).\\[0.1cm]
\textbf{[S189]} Hagenauer F., Sommer C., Higuchi T., Altintas O. and Dressler F., \textit{Parked cars as virtual network infrastructure: Enabling stable V2I access for long-lasting data flows}, CarSys 2017 - Proceedings of the 2nd ACM International Workshop on Smart, Autonomous, and Connected Vehicular Systems and Services, co-located with MobiCom 2017, no volume, pp. 57-64 (2017).\\[0.1cm]
\textbf{[S190]} Hajj H., El-Hajj W., El Dana M., Dakroub M. and Fawaz F., \textit{An extensible software framework for building vehicle to vehicle applications}, IWCMC 2010 - Proceedings of the 6th International Wireless Communications and Mobile Computing Conference, no volume, pp. 26-31 (2010).\\[0.1cm]
\textbf{[S191]} Han C., Yang T., Wei S., Feng H., Wang J. and Zhang G., \textit{Steering Machine Learning Mechanism Based on Big Data Integrated Cooperative Collision Avoidance for MASS}, Lecture Notes in Electrical Engineering, \textbf{571 LNEE}, pp. 2542-2549 (2020).\\[0.1cm]
\textbf{[S192]} Hasan S., Lee K., Moon D., Kwon S., Jinwoo S. and Lee S., \textit{Augmented reality and digital twin system for interaction with construction machinery}, Journal of Asian Architecture and Building Engineering, no volume, (no pages found) (2021).\\[0.1cm]
\textbf{[S193]} Hashim A., Shariff A. and Fadilah S., \textit{The modified safe clustering algorithm for vehicular ad hoc networks}, IEEE Student Conference on Research and Development: Inspiring Technology for Humanity, SCOReD 2017 - Proceedings, \textbf{2018-January}, pp. 263-268 (2018).\\[0.1cm]
\textbf{[S194]} Hassan A., Ahmed M. and Rahman M., \textit{Adaptive beaconing system based on fuzzy logic approach for vehicular network}, IEEE International Symposium on Personal, Indoor and Mobile Radio Communications, PIMRC, no volume, pp. 2581-2585 (2013).\\[0.1cm]
\textbf{[S195]} He X., Cai W., He H., Dong C., Sun B., Gong L., Tian M. and Wang X., \textit{Top-level design of electricity consumption information collection system based on edge computing}, Proceedings - 2021 13th International Conference on Measuring Technology and Mechatronics Automation, ICMTMA 2021, no volume, pp. 522-526 (2021).\\[0.1cm]
\textbf{[S196]} Hei X., Du X., Lin S. and Lee I., \textit{PIPAC: Patient infusion pattern based access control scheme for wireless insulin pump system}, Proceedings - IEEE INFOCOM, no volume, pp. 3030-3038 (2013).\\[0.1cm]
\textbf{[S197]} Heimdahl, M. P. E., \textit{Verifying communication constraints in RSML specifications}, Proceedings 1997 High-Assurance Engineering Workshop, p. 56–61 (1997).\\[0.1cm]
\textbf{[S198]} Heinrich M., Renkel D., Arul T. and Katzenbeisser S., \textit{Predicting Railway Signalling Commands Using Neural Networks for Anomaly Detection}, Lecture Notes in Computer Science (including subseries Lecture Notes in Artificial Intelligence and Lecture Notes in Bioinformatics), \textbf{12234 LNCS}, pp. 164-178 (2020).\\[0.1cm]
\textbf{[S199]} Hesselbarth A., Medina D., Ziebold R., Sandler M., Hoppe M. and Uhlemann M., \textit{Enabling Assistance Functions for the Safe Navigation of Inland Waterways}, IEEE Intelligent Transportation Systems Magazine, \textbf{12(3)}, pp. 123-135 (2020).\\[0.1cm]
\textbf{[S200]} Hossain N., Ovi J., Tasnim S., Islam N. and Zishan S., \textit{Design and development of Wearable multisensory smart device for human safety}, 2021 IEEE International IOT, Electronics and Mechatronics Conference, IEMTRONICS 2021 - Proceedings, no volume, (no pages found) (2021).\\[0.1cm]
\textbf{[S201]} Hota G., Sharma S., Rathore A., Joshi S. and Shah H., \textit{An Integrated Visual Signalling, Localisation Health Monitoring System for Soldier Assistance}, Proceedings of 2019 3rd IEEE International Conference on Electrical, Computer and Communication Technologies, ICECCT 2019, no volume, (no pages found) (2019).\\[0.1cm]
\textbf{[S202]} Hsieh C., Tsai H., Yang S. and Lin S., \textit{Predict scooter's stopping event using smartphone as the sensing device}, Proceedings - 2014 IEEE International Conference on Internet of Things, iThings 2014, 2014 IEEE International Conference on Green Computing and Communications, GreenCom 2014 and 2014 IEEE International Conference on Cyber-Physical-Social Computing, CPS 2014, no volume, pp. 17-23 (2014).\\[0.1cm]
\textbf{[S203]} Hsu C., Yang S. and Wu W., \textit{Constructing intelligent home-security system design with combining phone-net and bluetooth mechanism}, Proceedings of the 2009 International Conference on Machine Learning and Cybernetics, \textbf{6}, pp. 3316-3323 (2009).\\[0.1cm]
\textbf{[S204]} Hsu Y. and Matsuoka M., \textit{A Deep Reinforcement Learning Approach for Anomaly Network Intrusion Detection System}, Proceedings - 2020 IEEE 9th International Conference on Cloud Networking, CloudNet 2020, no volume, (no pages found) (2020).\\[0.1cm]
\textbf{[S205]} Hu Q., Cao J., Gao G., Xu L. and Song M., \textit{Study of an evaluation model for AIS receiver sensitivity measurements}, IEEE Transactions on Instrumentation and Measurement, \textbf{69(4)}, pp. 1118-1126 (2020).\\[0.1cm]
\textbf{[S206]} Hua G., Zhu L., Wu J., Shen C., Zhou L. and Lin Q., \textit{Blockchain-based federated learning for intelligent control in heavy haul railway}, IEEE Access, \textbf{8}, pp. 176830-176839 (2020).\\[0.1cm]
\textbf{[S207]} Huang C., Chuang Y., Yang D., Chen I., Chen Y. and Hu K., \textit{A mobility-aware link enhancement mechanism for vehicular Ad Hoc networks}, Eurasip Journal on Wireless Communications and Networking, \textbf{2008}, (no pages found) (2008).\\[0.1cm]
\textbf{[S208]} Huang C., Ciou Y., Shen P., Lin Y. and Mai J., \textit{The application of IoT for intelligence navigation}, ISPACS 2012 - IEEE International Symposium on Intelligent Signal Processing and Communications Systems, no volume, pp. 222-226 (2012).\\[0.1cm]
\textbf{[S209]} Huang J., Zhong Z. and Huo H., \textit{A dynamic energy-saving strategy for green cellular railway communication network}, Eurasip Journal on Wireless Communications and Networking, \textbf{2015(1)}, (no pages found) (2015).\\[0.1cm]
\textbf{[S210]} Huang Y., Wang W., Wang H., Jiang T. and Zhang Q., \textit{Authenticating On-Body IoT Devices: An Adversarial Learning Approach}, IEEE Transactions on Wireless Communications, \textbf{19(8)}, pp. 5234-5245 (2020).\\[0.1cm]
\textbf{[S211]} Huang, M., et al., \textit{VR Technology and the Theoretical Model and Method of Road Network Traffic State Information Accident Prediction and Perception under the Environment of Internet of Vehicles}, IEEE Access, p. 1–1 (2020).\\[0.1cm]
\textbf{[S212]} Huch S., Ongel A., Betz J. and Lienkamp M., \textit{Multi-task end-to-end self-driving architecture for cav platoons}, Sensors (Switzerland), \textbf{21(4)}, pp. 1-20 (2021).\\[0.1cm]
\textbf{[S213]} "Hüffmeier J., Sanchez-Heres L., Rylander R., Nord S., Alissa S., Bagge A., Bergljung P., Henkel P., Rydlinger A. and Hansen P., \textit{""pREParE SHIPS"" for Automated Ship Passages by Modern Decision Support Tools by Exchanging Future Positions}, IOP Conference Series: Materials Science and Engineering, \textbf{929(1)}, (no pages found) (2020)."\\[0.1cm]
\textbf{[S214]} Hussain T., Muhammad K., Ser J., Baik S. and De Albuquerque V., \textit{Intelligent Embedded Vision for Summarization of Multiview Videos in IIoT}, IEEE Transactions on Industrial Informatics, \textbf{16(4)}, pp. 2592-2602 (2020).\\[0.1cm]
\textbf{[S215]} Inedjaren Y., Zeddini B., Maachaoui M. and Barbot J., \textit{Securing intelligent communications on the vehicular AdHoc networks using fuzzy logic based trust OLSR}, Proceedings of IEEE/ACS International Conference on Computer Systems and Applications, AICCSA, \textbf{2019-November}, (no pages found) (2019).\\[0.1cm]
\textbf{[S216]} Inoue K., Yamamoto M., Mae Y., Takubo T. and Arai T., \textit{Design of search balls with wide field of view for searching inside of rubble}, Proceedings of the 2005 IEEE International Workshop on Safety, Security and Rescue Robotics, \textbf{2005}, pp. 170-175 (2005).\\[0.1cm]
\textbf{[S217]} Iordanova, B. N., \textit{Global neural network for automated air space-time allocation and control}, Proceedings of the 2004 IEEE International Conference on Computational Intelligence for Homeland Security and Personal Safety, 2004. CIHSPS 2004., p. 72–79 (2004).\\[0.1cm]
\textbf{[S218]} Islam, M., et al., \textit{Vision-Based Personal Safety Messages (PSMs) Generation for Connected Vehicles}, IEEE Transactions on Vehicular Technology, vol. 69, no 9, p. 9402–16 (2020).\\[0.1cm]
\textbf{[S219]} Jabbar, R., et al., \textit{Urban Traffic Monitoring and Modeling System: An IoT Solution for Enhancing Road Safety}, 2019 International Conference on Internet of Things, Embedded Systems and Communications (IINTEC), IEEE, p. 13–18 (2019).\\[0.1cm]
\textbf{[S220]} Jakovlev S., Daranda A., Voznak M., Lektauers A., Eglynas T. and Jusis M., \textit{Analysis of the Possibility to Detect Fake Vessels in the Automatic Identification System}, 2020 61st International Scientific Conference on Information Technology and Management Science of Riga Technical University, ITMS 2020 - Proceedings, no volume, (no pages found) (2020).\\[0.1cm]
\textbf{[S221]} Jalil Piran M., Ali A. and Suh D., \textit{Fuzzy-based sensor fusion for cognitive radio-based vehicular Ad hoc and sensor networks}, Mathematical Problems in Engineering, \textbf{2015}, (no pages found) (2015).\\[0.1cm]
\textbf{[S222]} Jana B., Mitra S. and Poray J., \textit{An analysis of security threats and countermeasures in VANET}, 2016 International Conference on Computer, Electrical and Communication Engineering, ICCECE 2016, no volume, (no pages found) (2017).\\[0.1cm]
\textbf{[S223]} Jasmine G., Marry D., Lakshmi S., Rishiwanth R., Sreehariprasath K. and Surendhar J., \textit{Camera based text and Product Lable Reading for Blind People}, 2021 7th International Conference on Advanced Computing and Communication Systems, ICACCS 2021, no volume, pp. 1122-1126 (2021).\\[0.1cm]
\textbf{[S224]} Ji S., Kim J. and You C., \textit{CADMA: collision-avoidance directional medium access for vehicular ad hoc networks}, Wireless Networks, \textbf{22(4)}, pp. 1181-1197 (2016).\\[0.1cm]
\textbf{[S225]} Jiang, Z., \textit{Emergency Communication Resource Prediction and Scheduling System Based on Deep Learning}, 2021 IEEE 5th Advanced Information Technology, Electronic and Automation Control Conference (IAEAC), vol. 5, 2021, p. 944–49 (2021).\\[0.1cm]
\textbf{[S226]} Jiménez F., Naranjo J., Anaya J., García F., Ponz A. and Armingol J., \textit{Advanced Driver Assistance System for Road Environments to Improve Safety and Efficiency}, Transportation Research Procedia, \textbf{14}, pp. 2245-2254 (2016).\\[0.1cm]
\textbf{[S227]} Jin Z., Li N., Liu C., Li M., An S. and Huang W., \textit{A Spatial-Temporal Features Based Fingerprinting Method for Machine tools in DNC Networks}, Proceedings - 2020 IEEE 22nd International Conference on High Performance Computing and Communications, IEEE 18th International Conference on Smart City and IEEE 6th International Conference on Data Science and Systems, HPCC-SmartCity-DSS 2020, no volume, pp. 360-368 (2020).\\[0.1cm]
\textbf{[S228]} Jinxiang Z. and Chuanwei S., \textit{Research on elevator general uninterrupted safety system based on Zigbee and Internet}, 2017 IEEE 2nd International Conference on Signal and Image Processing, ICSIP 2017, \textbf{2017-January}, pp. 484-488 (2017).\\[0.1cm]
\textbf{[S229]} John O. and Reimann M., \textit{Increasing quality of maritime communication through intelligent speech recognition and radio direction finding}, 2020 European Navigation Conference, ENC 2020, no volume, (no pages found) (2020).\\[0.1cm]
\textbf{[S230]} Jornod, G., et al., \textit{Packet Inter-Reception Time Conditional Density Estimation Based on Surrounding Traffic Distribution}, IEEE Open Journal of Intelligent Transportation Systems, vol. 1, p. 51–62 (2020).\\[0.1cm]
\textbf{[S231]} Juyal P. and Sharma S., \textit{Locating people in Real-World for Assisting Crowd Behaviour Analysis Using SSD and Deep SORT Algorithm}, 2021 International Conference on Wireless Communications, Signal Processing and Networking, WiSPNET 2021, no volume, pp. 350-353 (2021).\\[0.1cm]
\textbf{[S232]} Kaadan A. and Refai H., \textit{iICAS: Intelligent intersection collision avoidance system}, IEEE Conference on Intelligent Transportation Systems, Proceedings, ITSC, no volume, pp. 1184-1190 (2012).\\[0.1cm]
\textbf{[S233]} Kamal R., Chemmanam A., Jose B., Mathews S. and Varghese E., \textit{Construction safety surveillance using machine learning}, 2020 International Symposium on Networks, Computers and Communications, ISNCC 2020, no volume, (no pages found) (2020).\\[0.1cm]
\textbf{[S234]} Kamejima, K., \textit{Image-Based Satellite-Roadway-Vehicle Integration for Informatic Vicinity Generation}, ROMAN 2006 - The 15th IEEE International Symposium on Robot and Human Interactive Communication, p. 334–39 (2006).\\[0.1cm]
\textbf{[S235]} Kamel, J., et al., \textit{Misbehavior Detection in C-ITS: A comparative approach of local detection mechanisms}, 2019 IEEE Vehicular Networking Conference (VNC), p. 1–8 (2019).\\[0.1cm]
\textbf{[S236]} Kammath A., Nair K., Sudar S. and Gujarathi A., \textit{Design of an intelligent electric vehicle for blind}, 7th International Conference on Intelligent Systems and Control, ISCO 2013, no volume, pp. 244-249 (2013).\\[0.1cm]
\textbf{[S237]} Kanagaraj S., Arjun G. and Shahina A., \textit{Cheeka: A mobile application for personal safety}, Proceedings of the 9th IEEE International Conference on Collaborative Computing: Networking, Applications and Worksharing, COLLABORATECOM 2013, no volume, pp. 289-294 (2013).\\[0.1cm]
\textbf{[S238]} Kang L., Zhao W., Qi B. and Banerjee S., \textit{Augmenting self-driving with remote control: Challenges and directions}, HotMobile 2018 - Proceedings of the 19th International Workshop on Mobile Computing Systems and Applications, \textbf{2018-February}, pp. 19-24 (2018).\\[0.1cm]
\textbf{[S239]} Kapoor R., Birok R. and Manoj D., \textit{Soundless Horn and Remote Patroller}, 2014 IEEE International Conference on Vehicular Electronics and Safety, ICVES 2014, no volume, pp. 112-116 (2014).\\[0.1cm]
\textbf{[S240]} Kar, A., et al., \textit{SPEED CONTROLLING \& TRAFFIC MANAGEMENT SYSTEM (SCTMS)}, 2019 IEEE 10th Annual Ubiquitous Computing, Electronics \& Mobile Communication Conference (UEMCON), IEEE, p. 1058–63 (2019).\\[0.1cm]
\textbf{[S241]} Karam R., Salomon M. and Couturier R., \textit{A Comparative Study of Deep Learning Architectures for Detection of Anomalous ADS-B Messages}, 7th International Conference on Control, Decision and Information Technologies, CoDIT 2020, no volume, pp. 241-246 (2020).\\[0.1cm]
\textbf{[S242]} Kasturi G., Jain A. and Singh J., \textit{Detection and classification of radio frequency Jamming attacks using machine learning}, Journal of Wireless Mobile Networks, Ubiquitous Computing, and Dependable Applications, \textbf{11(4)}, pp. 49-62 (2020).\\[0.1cm]
\textbf{[S243]} Kasturi G., Jain A. and Singh J., \textit{Machine Learning-Based RF Jamming Classification Techniques in Wireless Ad Hoc Networks}, Lecture Notes on Data Engineering and Communications Technologies, \textbf{51}, pp. 99-111 (2020).\\[0.1cm]
\textbf{[S244]} Katsikas L., Chatzikokolakis K. and Alonistioti N., \textit{Implementing clustering for vehicular ad-hoc networks in NS-3}, ACM International Conference Proceeding Series, no volume, pp. 25-31 (2015).\\[0.1cm]
\textbf{[S245]} Kebria P., Khosravi A., Nahavandi S., Watters D., Guest G. and Shi P., \textit{Robust adaptive control of internet-based bilateral teleoperation systems with time-varying delay and model uncertainties}, Proceedings of the IEEE International Conference on Industrial Technology, \textbf{2019-February}, pp. 187-192 (2019).\\[0.1cm]
\textbf{[S246]} Kerczewski B., Wilson J. and Bishop B., \textit{Parameter impact on sharing studies between UAS CNPC satellite transmitters and terrestrial systems}, ICNS 2015 - Innovation in Operations, Implementation Benefits and Integration of the CNS Infrastructure, Conference Proceedings, no volume, (no pages found) (2015).\\[0.1cm]
\textbf{[S247]} Khan M., Reggiani L., Alam M., Moullec Y., Sharma N., Yaacoub E. and Magarini M., \textit{Q-learning based joint energy-spectral efficiency optimization in multi-hop device-to-device communication}, Sensors (Switzerland), \textbf{20(22)}, pp. 1-23 (2020).\\[0.1cm]
\textbf{[S248]} Khan, S., et al., \textit{Smart Object Detection and Home Appliances Control System in Smart Cities}, Computers, Materials \& Continua, vol. 67, no 1, p. 895–915 (2021).\\[0.1cm]
\textbf{[S249]} Khatri S., Vachhani H., Shah S., Bhatia J., Chaturvedi M., Tanwar S. and Kumar N., \textit{Machine learning models and techniques for VANET based traffic management: Implementation issues and challenges}, Peer-to-Peer Networking and Applications, \textbf{14(3)}, pp. 1778-1805 (2021).\\[0.1cm]
\textbf{[S250]} Khemapech, I., \textit{Bridge structural monitoring and warning system application in Thailand — Experiences learned}, 2017 TRON Symposium (TRONSHOW), p. 1–8 (2017).\\[0.1cm]
\textbf{[S251]} Kihei B., Copeland J. and Chang Y., \textit{Automotive Doppler sensing: The Doppler profile with machine learning in vehicle-to-vehicle networks for road safety}, IEEE Workshop on Signal Processing Advances in Wireless Communications, SPAWC, \textbf{2017-July}, pp. 1-5 (2017).\\[0.1cm]
\textbf{[S252]} Kim D. and Lee I., \textit{Deep learning-based power control scheme for perfect fairness in device-to-device communication systems}, Electronics (Switzerland), \textbf{9(10)}, pp. 1-17 (2020).\\[0.1cm]
\textbf{[S253]} Kim H., Becerra R., Bolufé S., Azurdia-Meza C., Montejo-Sánchez S. and Zabala-Blanco D., \textit{Neuroevolution-based adaptive antenna array beamforming scheme to improve the v2v communication performance at intersections}, Sensors, \textbf{21(9)}, (no pages found) (2021).\\[0.1cm]
\textbf{[S254]} Kim T., Ko K., Hwang I., Hong D., Choi S. and Wang H., \textit{RSRP-Based Doppler Shift Estimator Using Machine Learning in High-Speed Train Systems}, IEEE Transactions on Vehicular Technology, \textbf{70(1)}, pp. 371-380 (2021).\\[0.1cm]
\textbf{[S255]} Kim, C., et al. \textit{A Medical Device Safety Supervision over Wireless}, Reliable and Autonomous Computational Science, organizado por Sung Y. Shin et al., Springer, p. 21–40 (2010).\\[0.1cm]
\textbf{[S256]} Kist M., Faganello L., Bondan L., Marotta M., Granville L., Rochol J. and Both C., \textit{Adaptive threshold architecture for spectrum sensing in public safety radio channels}, 2015 IEEE Wireless Communications and Networking Conference, WCNC 2015, no volume, pp. 287-292 (2015).\\[0.1cm]
\textbf{[S257]} Klus, R., et al., \textit{Neural Network Fingerprinting and GNSS Data Fusion for Improved Localization in 5G}, 2021 International Conference on Localization and GNSS (ICL-GNSS), IEEE, p. 1–6 (2021).\\[0.1cm]
\textbf{[S258]} Koulakezian A. and Leon-Garcia A., \textit{CVI: Connected vehicle Infrastructure for ITS}, IEEE International Symposium on Personal, Indoor and Mobile Radio Communications, PIMRC, no volume, pp. 750-755 (2011).\\[0.1cm]
\textbf{[S259]} Kourtis M., Blanco B., Perez-Romero J., Makris D., McGrath M., Xilouris G., Munaretto D., Solozabal R., Sanchoyerto A., Giannoulakis I., Kafetzakis E., Riccobene V., Jimeno E., Kourtis A., Ferrus R., Liberal F., Koumaras H., Kostopoulos A. and Chochliouros I., \textit{A Cloud-Enabled Small Cell Architecture in 5G Networks for Broadcast/Multicast Services}, IEEE Transactions on Broadcasting, \textbf{65(2)}, pp. 414-424 (2019).\\[0.1cm]
\textbf{[S260]} Krishnaveni P. and Sutha J., \textit{Novel deep learning framework for broadcasting abnormal events obtained from surveillance applications}, Journal of Ambient Intelligence and Humanized Computing, no volume, (no pages found) (2020).\\[0.1cm]
\textbf{[S261]} Kui L., Xin J., Wenke K., Caihua L. and Yang Y., \textit{The customized 5G secondary authentication scheme combined with security protection strategy for electrical automation system}, None, no volume, pp. 757-761 (2020).\\[0.1cm]
\textbf{[S262]} Kumar N., Chilamkurti N. and Park J., \textit{ALCA: Agent learning-based clustering algorithm in vehicular ad hoc networks}, Personal and Ubiquitous Computing, \textbf{17(8)}, pp. 1683-1692 (2013).\\[0.1cm]
\textbf{[S263]} Kushwah N. and Sonker A., \textit{Malicious Node Detection on Vehicular Ad-Hoc Network Using Dempster Shafer Theory for Denial of Services Attack}, Proceedings - 2016 8th International Conference on Computational Intelligence and Communication Networks, CICN 2016, no volume, pp. 432-436 (2017).\\[0.1cm]
\textbf{[S264]} Kusumoto A., De Vasconcelos L., Leite N., Lopes C. and Pirk R., \textit{Tracking track targets in external store separation using computer vision}, Proceedings of the International Telemetering Conference, \textbf{50}, pp. 381-388 (2014).\\[0.1cm]
\textbf{[S265]} Kwon E., Shin W., Park H., Byon S., Jung E., Lee Y. and Lee K., \textit{A Moving Pattern Classification Based on Multimodal Data for Public Safety Services}, International Conference on ICT Convergence, \textbf{2020-October}, pp. 1589-1591 (2020).\\[0.1cm]
\textbf{[S266]} Kwon J. and Im C., \textit{Subject-Independent Functional Near-Infrared Spectroscopy-Based Brain–Computer Interfaces Based on Convolutional Neural Networks}, Frontiers in Human Neuroscience, \textbf{15}, (no pages found) (2021).\\[0.1cm]
\textbf{[S267]} Lages A., Delicato F., Vianna G. and Pirmez L., \textit{A service-oriented fuzzy reputation system to increase the security of a broadband wireless metropolitan network}, Proceedings - 2006 IEEE International Conference on Networks, ICON 2006 - Networking-Challenges and Frontiers, \textbf{2}, pp. 306-311 (2006).\\[0.1cm]
\textbf{[S268]} Lalitha R. and Suma G., \textit{An Adaptive Approach for RFID Based Data Dissemination in VANETs through ABC Algorithm Using Android Mobiles}, Proceedings - 2014 4th International Conference on Artificial Intelligence with Applications in Engineering and Technology, ICAIET 2014, no volume, pp. 298-303 (2015).\\[0.1cm]
\textbf{[S269]} Lan, W., \textit{Target Tracking and Risk Avoidance System for Intelligent Driving System Based on 5G Signal Anomaly Detection}, International Journal of Communication Systems, Wiley Online Library, (2020).\\[0.1cm]
\textbf{[S270]} Lavanya K., Ramya Shree D., Nischitha B., Asha T. and Gururaj C., \textit{Vision Interfaced War Field Robot with Wireless Video Transmission}, Proceedings of the 2nd International Conference on Trends in Electronics and Informatics, ICOEI 2018, no volume, pp. 833-838 (2018).\\[0.1cm]
\textbf{[S271]} Lee B., Su S. and Rudas I., \textit{Content-independent image processing based fall detection}, Proceedings 2011 International Conference on System Science and Engineering, ICSSE 2011, no volume, pp. 654-659 (2011).\\[0.1cm]
\textbf{[S272]} Lee H. and Park D., \textit{AI TTS Smartphone App for Communication of Speech Impaired People}, Studies in Computational Intelligence, \textbf{929}, pp. 219-229 (2021).\\[0.1cm]
\textbf{[S273]} Lee K., Lee Y. and Xie M., \textit{Designing a human vehicle interface for an intelligent community vehicle}, IEEE Conference on Intelligent Transportation Systems, Proceedings, ITSC, no volume, pp. 740-745 (1999).\\[0.1cm]
\textbf{[S274]} Lee Y. and Lee C., \textit{Real-Time Smart Home Surveillance System of Based on Raspberry Pi}, 2nd IEEE Eurasia Conference on IOT, Communication and Engineering 2020, ECICE 2020, no volume, pp. 72-74 (2020).\\[0.1cm]
\textbf{[S275]} Li C., Guo W., Sun S., Al-Rubaye S. and Tsourdos A., \textit{Trustworthy Deep Learning in 6G-Enabled Mass Autonomy: From Concept to Quality-of-Trust Key Performance Indicators}, IEEE Vehicular Technology Magazine, \textbf{15(4)}, pp. 112-121 (2020).\\[0.1cm]
\textbf{[S276]} Li G., Ota K., Dong M., Wu J. and Li J., \textit{DeSVig: Decentralized Swift Vigilance against Adversarial Attacks in Industrial Artificial Intelligence Systems}, IEEE Transactions on Industrial Informatics, \textbf{16(5)}, pp. 3267-3277 (2020).\\[0.1cm]
\textbf{[S277]} Li L., Ota K. and Dong M., \textit{Human in the Loop: Distributed Deep Model for Mobile Crowdsensing}, IEEE Internet of Things Journal, \textbf{5(6)}, pp. 4957-4964 (2018).\\[0.1cm]
\textbf{[S278]} Li M. and Wang J., \textit{5G and Artificial Intelligence with Multi-Sensor Data Fusion Technology Support Intelligent Operation Monitoring of Power Grid in the Future}, Proceedings - 2020 Chinese Automation Congress, CAC 2020, no volume, pp. 177-180 (2020).\\[0.1cm]
\textbf{[S279]} Li M., Gao J., Zhang N., Zhao L. and Shen X., \textit{Collaborative Computing in Vehicular Networks: A Deep Reinforcement Learning Approach}, IEEE International Conference on Communications, \textbf{2020-June}, (no pages found) (2020).\\[0.1cm]
\textbf{[S280]} Li T., Lei S., Wang W. and Wang Q., \textit{Multi-person collaborative hoisting training system based on mixed reality}, Proceedings - 2020 International Conference on Computer Engineering and Intelligent Control, ICCEIC 2020, no volume, pp. 258-265 (2020).\\[0.1cm]
\textbf{[S281]} Li T., Zhao M. and Wong K., \textit{Machine learning based code dissemination by selection of reliability mobile vehicles in 5G networks}, Computer Communications, \textbf{152}, pp. 109-118 (2020).\\[0.1cm]
\textbf{[S282]} Li, K., et al., \textit{Cooperative and cognitive wireless networks for communication-based train control (CBTC) systems}, 2015 IEEE International Conference on Communications (ICC), p. 3958–62 (2015).\\[0.1cm]
\textbf{[S283]} Li, Y., \textit{Anti-Fatigue and Collision Avoidance Systems for Intelligent Vehicles with Ultrasonic and Li-Fi Sensors}, IEEE 3rd International Conference on Information Communication and Signal Processing (ICICSP), p. 203–09 (2020).\\[0.1cm]
\textbf{[S284]} Liang H., Zhang X., Hong X., Zhang Z., Li M., Hu G. and Hou F., \textit{Reinforcement Learning Enabled Dynamic Resource Allocation in the Internet of Vehicles}, IEEE Transactions on Industrial Informatics, \textbf{17(7)}, pp. 4957-4967 (2021).\\[0.1cm]
\textbf{[S285]} Liao C., Shou G., Liu Y., Hu Y. and Guo Z., \textit{Intelligent traffic accident detection system based on mobile edge computing}, 2017 3rd IEEE International Conference on Computer and Communications, ICCC 2017, \textbf{2018-January}, pp. 2110-2115 (2018).\\[0.1cm]
\textbf{[S286]} Liao H., Wang W., Wang S. and Xu X., \textit{Multi-Scale Ship Tracking Based on Maritime Monitoring Platform}, Proceedings of the 15th IEEE Conference on Industrial Electronics and Applications, ICIEA 2020, no volume, pp. 945-949 (2020).\\[0.1cm]
\textbf{[S287]} Lim K., Islam T., Kim H. and Joung J., \textit{A Sybil Attack Detection Scheme based on ADAS Sensors for Vehicular Networks}, 2020 IEEE 17th Annual Consumer Communications and Networking Conference, CCNC 2020, no volume, (no pages found) (2020).\\[0.1cm]
\textbf{[S288]} Limouchi E. and Mahgoub I., \textit{Smart fuzzy logic-based density and distribution adaptive scheme for efficient data dissemination in vehicular ad hoc networks}, Electronics (Switzerland), \textbf{9(8)}, pp. 1-22 (2020).\\[0.1cm]
\textbf{[S289]} Lin C., Lin Y., Chen L. and Wang Y., \textit{Front vehicle blind spot translucentization based on augmented reality}, IEEE Vehicular Technology Conference, no volume, (no pages found) (2013).\\[0.1cm]
\textbf{[S290]} Lin Z., Yu L., Zi W. and Yingtian D., \textit{Cross-layer safety-critical broadcast service architecture integrating VANETs with 3g networks in Iot environments}, China Communications, \textbf{8(8)}, pp. 13-24 (2011).\\[0.1cm]
\textbf{[S291]} Liu F., Guo Y., Cai Z., Xiao N. and Zhao Z., \textit{Edge-enabled disaster rescue: A case study of searching for missing people}, ACM Transactions on Intelligent Systems and Technology, \textbf{10(6)}, (no pages found) (2019).\\[0.1cm]
\textbf{[S292]} Liu J., Qian L., Zhang Y., Han J. and Sun J., \textit{Towards Safety-Risk Prediction of CBTC Systems with Deep Learning and Formal Methods}, IEEE Access, \textbf{8}, pp. 16618-16626 (2020).\\[0.1cm]
\textbf{[S293]} Liu J., Wu J. and Liu M., \textit{UAV monitoring and forecasting model in intelligent traffic oriented applications}, Computer Communications, \textbf{153}, pp. 499-506 (2020).\\[0.1cm]
\textbf{[S294]} Liu J., Zhang Y., Han J., He J., Sun J. and Zhou T., \textit{Intelligent Hazard-Risk Prediction Model for Train Control Systems}, IEEE Transactions on Intelligent Transportation Systems, \textbf{21(11)}, pp. 4693-4704 (2020).\\[0.1cm]
\textbf{[S295]} Liu J., Zhao L., Zheng K. and Zhou Q., \textit{A Distributed Driving Decision Scheme Based on Reinforcement Learning for Autonomous Driving Vehicles}, IEEE Vehicular Technology Conference, \textbf{2020-May}, (no pages found) (2020).\\[0.1cm]
\textbf{[S296]} Liu L., Zhang X., Qiao M. and Shi W., \textit{SafeShareRide: Edge-based attack detection in ridesharing services}, Proceedings - 2018 3rd ACM/IEEE Symposium on Edge Computing, SEC 2018, no volume, pp. 17-29 (2018).\\[0.1cm]
\textbf{[S297]} Liu R., He Y., Zhao Y., Jiang X. and Ren S., \textit{Tunnel construction ventilation frequency-control based on radial basis function neural network}, Automation in Construction, \textbf{118}, (no pages found) (2020).\\[0.1cm]
\textbf{[S298]} Liu X., Liu Y., Chen Y. and Hanzo L., \textit{Enhancing the Fuel-Economy of V2I-Assisted Autonomous Driving: A Reinforcement Learning Approach}, IEEE Transactions on Vehicular Technology, \textbf{69(8)}, pp. 8329-8342 (2020).\\[0.1cm]
\textbf{[S299]} Liu Y. and Yuan L., \textit{Research on Train Control System Based on Train to Train Communication}, 2018 International Conference on Intelligent Rail Transportation, ICIRT 2018, no volume, (no pages found) (2019).\\[0.1cm]
\textbf{[S300]} Lo C., Cheng D., Chen C. and Yan J., \textit{Design and implementation of situation-aware medical tourism service search system}, 2008 International Conference on Wireless Communications, Networking and Mobile Computing, WiCOM 2008, no volume, (no pages found) (2008).\\[0.1cm]
\textbf{[S301]} Long C., Du X., Wang D. and Liu W., \textit{Research on Integrated Security Management and Control Technology of Big Data Information Platform in the Intelligent Community Based on 5G}, Proceedings of 2020 IEEE International Conference on Advances in Electrical Engineering and Computer Applications, AEECA 2020, no volume, pp. 1016-1020 (2020).\\[0.1cm]
\textbf{[S302]} Lopez-De-Teruel P., Gil Perez M., Garcia Clemente F., Ruiz Garcia A. and Martinez Perez G., \textit{5G-CAGE: A context and situational awareness system for city public safety with video processing at a virtualized ecosystem}, Proceedings - 2019 International Conference on Computer Vision Workshop, ICCVW 2019, no volume, pp. 2749-2757 (2019).\\[0.1cm]
\textbf{[S303]} Lu Y., Shi Y., Jia G. and Yang J., \textit{A new method for semantic consistency verification of aviation radiotelephony communication based on LSTM-RNN}, International Conference on Digital Signal Processing, DSP, \textbf{0}, pp. 422-426 (2016).\\[0.1cm]
\textbf{[S304]} Lü, Z., et al., \textit{An Architecture of System of Systems (SoS) for Commercial Flight Security in 5G OGCE}, 2018 5th International Conference on Systems and Informatics (ICSAI), p. 288–93 (2018).\\[0.1cm]
\textbf{[S305]} Luo L., Sheng L., Yu H. and Sun G., \textit{Intersection-Based V2X Routing via Reinforcement Learning in Vehicular Ad Hoc Networks}, IEEE Transactions on Intelligent Transportation Systems, no volume, (no pages found) (2021).\\[0.1cm]
\textbf{[S306]} Luong D., Hu Y., Li J., Ali M., Abdo K. and Rihacek C., \textit{Deep learning approach for the multilink selection problem in avionic networks}, AIAA/IEEE Digital Avionics Systems Conference - Proceedings, \textbf{2020-October}, (no pages found) (2020).\\[0.1cm]
\textbf{[S307]} Lv Z., Lloret J. and Song H., \textit{Guest Editorial Software Defined Internet of Vehicles}, IEEE Transactions on Intelligent Transportation Systems, \textbf{22(6)}, pp. 3504-3510 (2021).\\[0.1cm]
\textbf{[S308]} Lyu F., Cheng N., Zhu H., Zhou H., Xu W., Li M. and Shen X., \textit{Intelligent Context-Aware Communication Paradigm Design for IoVs Based on Data Analytics}, IEEE Network, \textbf{32(6)}, pp. 74-82 (2018).\\[0.1cm]
\textbf{[S309]} Lyu F., Zhu H., Cheng N., Zhou H., Xu W., Li M. and Shen X., \textit{Characterizing Urban Vehicle-to-Vehicle Communications for Reliable Safety Applications}, IEEE Transactions on Intelligent Transportation Systems, \textbf{21(6)}, pp. 2586-2602 (2020).\\[0.1cm]
\textbf{[S310]} Ma X., Huo E., Yu H. and Li H., \textit{Mining truck platooning patterns through massive trajectory data}, Knowledge-Based Systems, \textbf{221}, (no pages found) (2021).\\[0.1cm]
\textbf{[S311]} Ma Y., Liu X., Bao J., Liu C. and Dai Z., \textit{Design and implementation of vehicle safety robot based on computer vision}, Proceedings of the IEEE International Conference on Software Engineering and Service Sciences, ICSESS, \textbf{2019-October}, pp. 340-344 (2019).\\[0.1cm]
\textbf{[S312]} Ma, Y., et al., \textit{Real-Time Highway Traffic Condition Assessment Framework Using Vehicle–Infrastructure Integration (VII) With Artificial Intelligence (AI)}, IEEE Transactions on Intelligent Transportation Systems, vol. 10, no 4, p. 615–27 (2009).\\[0.1cm]
\textbf{[S313]} Maaroufi S. and Pierre S., \textit{BCOOL: A Novel Blockchain Congestion Control Architecture Using Dynamic Service Function Chaining and Machine Learning for Next Generation Vehicular Networks}, IEEE Access, \textbf{9}, pp. 53096-53122 (2021).\\[0.1cm]
\textbf{[S314]} Mahjoub H., Tahmasbi-Sarvestani A., Kazemi H. and Fallah Y., \textit{A Learning-Based Framework for Two-Dimensional Vehicle Maneuver Prediction over V2V Networks}, Proceedings - 2017 IEEE 15th International Conference on Dependable, Autonomic and Secure Computing, 2017 IEEE 15th International Conference on Pervasive Intelligence and Computing, 2017 IEEE 3rd International Conference on Big Data Intelligence and Computing and 2017 IEEE Cyber Science and Technology Congress, DASC-PICom-DataCom-CyberSciTec 2017, \textbf{2018-January}, pp. 156-163 (2018).\\[0.1cm]
\textbf{[S315]} Mahmood A., Siddiqui S., Zhang W. and Sheng Q., \textit{A hybrid trust management model for secure and resource efficient vehicular ad hoc networks}, Proceedings - 2019 20th International Conference on Parallel and Distributed Computing, Applications and Technologies, PDCAT 2019, no volume, pp. 154-159 (2019).\\[0.1cm]
\textbf{[S316]} Malik K., Ahmad M., Khalid S., Ahmad H., Al-Turjman F. and Jabbar S., \textit{Image and command hybrid model for vehicle control using Internet of Vehicles}, Transactions on Emerging Telecommunications Technologies, \textbf{31(5)}, (no pages found) (2020).\\[0.1cm]
\textbf{[S317]} Mammeri A., Boukerche A. and Almulla M., \textit{Design of traffic sign detection, recognition, and transmission systems for smart vehicles}, IEEE Wireless Communications, \textbf{20(6)}, pp. 36-43 (2013).\\[0.1cm]
\textbf{[S318]} Mano L., Faiçal B., Nakamura L., Gomes P., Libralon G., Meneguete R., Filho G., Giancristofaro G., Pessin G., Krishnamachari B. and Ueyama J., \textit{Exploiting IoT technologies for enhancing Health Smart Homes through patient identification and emotion recognition}, Computer Communications, \textbf{89-90}, pp. 178-190 (2016).\\[0.1cm]
\textbf{[S319]} Mao L. and Chen S., \textit{The growth of mobile payment and effect on consumption via cash and bankcard}, Proceedings - 2015 8th International Conference on BioMedical Engineering and Informatics, BMEI 2015, no volume, pp. 872-877 (2016).\\[0.1cm]
\textbf{[S320]} Maragkos C., Vosniakos G. and Matsas E., \textit{Virtual reality assisted robot programming for human collaboration}, Procedia Manufacturing, \textbf{38}, pp. 1697-1704 (2019).\\[0.1cm]
\textbf{[S321]} Marginean A., Petrovai A., Negru M. and Nedevschi S., \textit{Cooperative application for lane change maneuver on Smart Mobile devices}, Proceedings - 2015 IEEE 11th International Conference on Intelligent Computer Communication and Processing, ICCP 2015, no volume, pp. 279-286 (2015).\\[0.1cm]
\textbf{[S322]} Mariappan, M., et al., \textit{Medical Tele-Diagnosis Robot (MTR) - Internet Based Communication \& Navigation System}, Applied Mechanics and Materials, vol. 490–491, p. 1177–89 (2014).\\[0.1cm]
\textbf{[S323]} Martin G., Koizia L., Kooner A., Cafferkey J., Ross C., Purkayastha S., Sivananthan A., Tanna A., Pratt P. and Kinross J., \textit{Use of the HoloLens2 mixed reality headset for protecting health care workers during the COVID-19 pandemic: Prospective, observational evaluation}, Journal of Medical Internet Research, \textbf{22(8)}, (no pages found) (2020).\\[0.1cm]
\textbf{[S324]} Martinelli D., Cerbaro J., Fabro J., De Oliveira A. and Simoes Teixeira M., \textit{Human-robot interface for remote control via IoT communication using deep learning techniques for motion recognition}, 2020 Latin American Robotics Symposium, 2020 Brazilian Symposium on Robotics and 2020 Workshop on Robotics in Education, LARS-SBR-WRE 2020, no volume, (no pages found) (2020).\\[0.1cm]
\textbf{[S325]} Martirosyan A., Boukerche A. and Nelem Pazzi R., \textit{Energy-aware and quality of service-based routing in wireless sensor networks and vehicular ad hoc networks}, Annales des Telecommunications/Annals of Telecommunications, \textbf{63(11-12)}, pp. 669-681 (2008).\\[0.1cm]
\textbf{[S326]} Marvasti E., Raftari A., Marvasti A. and Fallah Y., \textit{Bandwidth-adaptive feature sharing for cooperative LIDAR object detection}, 2020 IEEE 3rd Connected and Automated Vehicles Symposium, CAVS 2020 - Proceedings, no volume, (no pages found) (2020).\\[0.1cm]
\textbf{[S327]} Mathews E. and Poigné A., \textit{An echo state network based pedestrian counting system using wireless sensor networks}, Proceedings of the 6th Workshop on Intelligent Solutions in Embedded Systems, WISES'08, no volume, (no pages found) (2008).\\[0.1cm]
\textbf{[S328]} McCall J., Mallick S. and Trivedi M., \textit{Real-time driver affect analysis and tele-viewing system}, IEEE Intelligent Vehicles Symposium, Proceedings, no volume, pp. 372-377 (2003).\\[0.1cm]
\textbf{[S329]} Mekala M., Viswanathan P., Srinivasu N. and Varma G., \textit{Accurate Decision-making System for Mining Environment using Li-Fi 5G Technology over IoT Framework}, Proceedings of the 4th International Conference on Contemporary Computing and Informatics, IC3I 2019, no volume, pp. 74-79 (2019).\\[0.1cm]
\textbf{[S330]} Meneguette R. and Nakamura L., \textit{A flow control policy based on the class of applications of the vehicular networks}, MobiWac 2017 - Proceedings of the 15th ACM International Symposium on Mobility Management and Wireless Access, Co-located with MSWiM 2017, no volume, pp. 137-144 (2017).\\[0.1cm]
\textbf{[S331]} Meng Y. and Li J., \textit{Research on intelligent configuration method of mine IoT communication resources based on data flow behavior}, IEEE Access, \textbf{8}, pp. 172065-172075 (2020).\\[0.1cm]
\textbf{[S332]} Meng Y., Shangguan W., Cai B. and Zhang J., \textit{Fault prediction method of the on-board equipment of train control system based on grey-ENN}, Proceedings of 2019 11th CAA Symposium on Fault Detection, Supervision, and Safety for Technical Processes, SAFEPROCESS 2019, no volume, pp. 944-949 (2019).\\[0.1cm]
\textbf{[S333]} Menon, Z., et al., \textit{Multi Interactive Chatbot Communication Framework for Health Care}, International Journal of Computer Science and Network Security, vol. 20, no 4, p. 121–124 (2020).\\[0.1cm]
\textbf{[S334]} Merenda M., Fedele R., Pratico F., Carotenuto R., Della Corte F. and Iero D., \textit{Augmented Information Discovery using NFC Technology within a Platform for Disaster Monitoring}, 2020 5th International Conference on Smart and Sustainable Technologies, SpliTech 2020, no volume, (no pages found) (2020).\\[0.1cm]
\textbf{[S335]} Messaoud S., Bradai A. and Moulay E., \textit{Online GMM Clustering and Mini-Batch Gradient Descent Based Optimization for Industrial IoT 4.0}, IEEE Transactions on Industrial Informatics, \textbf{16(2)}, pp. 1427-1435 (2020).\\[0.1cm]
\textbf{[S336]} Mishra V., Shivankar N., Gadpayle S., Shinde S., Khan M. and Zunke S., \textit{Women's Safety System by Voice Recognition}, 2020 IEEE International Students' Conference on Electrical, Electronics and Computer Science, SCEECS 2020, no volume, (no pages found) (2020).\\[0.1cm]
\textbf{[S337]} Moltaji M., Heidar H. and Bizaki H., \textit{AIS link budget design and probability of detection analysis in multi user system}, 2020 28th Iranian Conference on Electrical Engineering, ICEE 2020, no volume, (no pages found) (2020).\\[0.1cm]
\textbf{[S338]} Moon H., Park P., Lee J. and Jung J., \textit{V2XRSF: V2X integrated runtime simulation framework based on virtual reality for cooperative vehicular communication applications evaluation}, Information (Japan), \textbf{16(3 B)}, pp. 2411-2420 (2013).\\[0.1cm]
\textbf{[S339]} Moubayed A. and Shami A., \textit{Softwarization, Virtualization, \& Machine Learning For Intelligent \& Effective V2X Communications}, IEEE Intelligent Transportation Systems Magazine, no volume, (no pages found) (2020).\\[0.1cm]
\textbf{[S340]} Mu'Azu A., Tang L., Hasbullah H. and Lawal I., \textit{A cluster-based stable routing algorithm for vehicular ad hoc network}, 2015 International Symposium on Mathematical Sciences and Computing Research, iSMSC 2015 - Proceedings, no volume, pp. 286-291 (2016).\\[0.1cm]
\textbf{[S341]} Mühlbacher-Karrer S., Mosa A., Faller L., Ali M., Hamid R., Zangl H. and Kyamakya K., \textit{A Driver State Detection System - Combining a Capacitive Hand Detection Sensor With Physiological Sensors}, IEEE Transactions on Instrumentation and Measurement, \textbf{66(4)}, pp. 624-636 (2017).\\[0.1cm]
\textbf{[S342]} Mukherjee A., Jain D., Goswami P., Xin Q., Yang L. and Rodrigues J., \textit{Back Propagation Neural Network Based Cluster Head Identification in MIMO Sensor Networks for Intelligent Transportation Systems}, IEEE Access, \textbf{8}, pp. 28524-28532 (2020).\\[0.1cm]
\textbf{[S343]} Mukherjee S. and Vu T., \textit{On Distributed Model-Free Reinforcement Learning Control with Stability Guarantee}, IEEE Control Systems Letters, \textbf{5(5)}, pp. 1615-1620 (2021).\\[0.1cm]
\textbf{[S344]} Murphey Y., Liu C., Tayyab M. and Narayan D., \textit{Accurate pedestrian path prediction using neural networks}, 2017 IEEE Symposium Series on Computational Intelligence, SSCI 2017 - Proceedings, \textbf{2018-January}, pp. 1-7 (2018).\\[0.1cm]
\textbf{[S345]} Mutoh N. and Shibata T., \textit{An indoor positioning system using a wearable wireless sensor and the support vector machine}, Asia-Pacific Microwave Conference Proceedings, APMC, \textbf{2019-December}, pp. 616-618 (2019).\\[0.1cm]
\textbf{[S346]} Nafi N. and Khan J., \textit{A VANET based intelligent road traffic signalling system}, Australasian Telecommunication Networks and Applications Conference, ATNAC 2012, no volume, (no pages found) (2012).\\[0.1cm]
\textbf{[S347]} Nagarani N., Venkatakrishnan P. and Balaji N., \textit{Unmanned Aerial vehicle's runway landing system with efficient target detection by using morphological fusion for military surveillance system}, Computer Communications, \textbf{151}, pp. 463-472 (2020).\\[0.1cm]
\textbf{[S348]} Nahar A. and Das D., \textit{Adaptive Reinforcement Routing in Software Defined Vehicular Networks}, 2020 International Wireless Communications and Mobile Computing, IWCMC 2020, no volume, pp. 2118-2123 (2020).\\[0.1cm]
\textbf{[S349]} Najada H. and Mahgoub I., \textit{Autonomous vehicles safe-optimal trajectory selection based on big data analysis and predefined user preferences}, 2016 IEEE 7th Annual Ubiquitous Computing, Electronics and Mobile Communication Conference, UEMCON 2016, no volume, (no pages found) (2016).\\[0.1cm]
\textbf{[S350]} Najdi R., Shaban T., Mourad M. and Karaki S., \textit{Hydrogen production and filling of fuel cell cars}, 2016 3rd International Conference on Advances in Computational Tools for Engineering Applications, ACTEA 2016, no volume, pp. 43-48 (2016).\\[0.1cm]
\textbf{[S351]} Nasim I., Ibrahim A. and Kim S., \textit{Learning-Based Beamforming for Multi-User Vehicular Communications: A Combinatorial Multi-Armed Bandit Approach}, IEEE Access, no volume, (no pages found) (2020).\\[0.1cm]
\textbf{[S352]} Naskath J., Paramasivan B. and Aldabbas H., \textit{A study on modeling vehicles mobility with MLC for enhancing vehicle-to-vehicle connectivity in VANET}, Journal of Ambient Intelligence and Humanized Computing, \textbf{12(8)}, pp. 8255-8264 (2021).\\[0.1cm]
\textbf{[S353]} Nassef O., Sequeira L., Salam E. and Mahmoodi T., \textit{Building a Lane Merge Coordination for Connected Vehicles Using Deep Reinforcement Learning}, IEEE Internet of Things Journal, \textbf{8(4)}, pp. 2540-2557 (2021).\\[0.1cm]
\textbf{[S354]} Nekovee M., Sharma S., Uniyal N., Nag A., Nejabati R. and Simeonidou D., \textit{Towards AI-enabled Microservice Architecture for Network Function Virtualization}, 2020 8th International Conference on Communications and Networking, ComNet2020 - Proceedings, no volume, (no pages found) (2020).\\[0.1cm]
\textbf{[S355]} Neudecker T., An N. and Hartenstein H., \textit{Verification and evaluation of fail-safe Virtual Traffic Light applications}, IEEE Vehicular Networking Conference, VNC, no volume, pp. 158-165 (2013).\\[0.1cm]
\textbf{[S356]} Nguyen H., Ta T., Nguyen N., Bui V., Pham H. and Nguyen D., \textit{YOLO Based Real-Time Human Detection for Smart Video Surveillance at the Edge}, ICCE 2020 - 2020 IEEE 8th International Conference on Communications and Electronics, no volume, pp. 439-444 (2021).\\[0.1cm]
\textbf{[S357]} Nguyen M., Garcia-Palacios E., Do-Duy T., Nguyen L., Mai S. and Duong T., \textit{Spectrum-Sharing UAV-Assisted Mission-Critical Communication: Learning-Aided Real-Time Optimisation}, IEEE Access, \textbf{9}, pp. 11622-11632 (2021).\\[0.1cm]
\textbf{[S358]} Nkoro A. and Vershinin Y., \textit{Current and future trends in applications of Intelligent Transport Systems on cars and infrastructure}, 2014 17th IEEE International Conference on Intelligent Transportation Systems, ITSC 2014, no volume, pp. 514-519 (2014).\\[0.1cm]
\textbf{[S359]} Nygate J., Hochgraf C., Indelicato M., Johnson W., Bazdresch M. and Reyes R., \textit{Applying Machine Learning in Managing Deployable Systems}, 2018 IEEE International Symposium on Technologies for Homeland Security, HST 2018, no volume, (no pages found) (2018).\\[0.1cm]
\textbf{[S360]} Obaid N., Hamad I., Madkhane A., Hamad Y. and El-Hassan F., \textit{Design and testing of a practical smart walking cane for the visually impaired}, Proceedings of IEEE/ACS International Conference on Computer Systems and Applications, AICCSA, \textbf{2019-November}, (no pages found) (2019).\\[0.1cm]
\textbf{[S361]} Oboe, R., \textit{Force-reflecting teleoperation over the Internet: the JBIT project}, Proceedings of the IEEE, vol. 91, no 3, p. 449–62 (2003).\\[0.1cm]
\textbf{[S362]} Ojeda F., Bisulco A., Kepple D., Isler V. and Lee D., \textit{On-Device Event Filtering with Binary Neural Networks for Pedestrian Detection Using Neuromorphic Vision Sensors}, Proceedings - International Conference on Image Processing, ICIP, \textbf{2020-October}, pp. 3084-3088 (2020).\\[0.1cm]
\textbf{[S363]} Olowononi F., Rawat D. and Liu C., \textit{Federated learning with differential privacy for resilient vehicular cyber physical systems}, 2021 IEEE 18th Annual Consumer Communications and Networking Conference, CCNC 2021, no volume, (no pages found) (2021).\\[0.1cm]
\textbf{[S364]} O'Mahony G., Harris P. and Murphy C., \textit{Detecting Interference in Wireless Sensor Network Received Samples: A Machine Learning Approach}, IEEE World Forum on Internet of Things, WF-IoT 2020 - Symposium Proceedings, no volume, (no pages found) (2020).\\[0.1cm]
\textbf{[S365]} Othman N. and Aydin I., \textit{A new IoT combined body detection of people by using computer vision for security application}, Proceedings - 9th International Conference on Computational Intelligence and Communication Networks, CICN 2017, \textbf{2018-January}, pp. 108-112 (2018).\\[0.1cm]
\textbf{[S366]} Palconit M., Formentera A., Aying R., Dianon K., Tadle J. and Dadios E., \textit{Speech Activation for Internet of Things Security System in Public Utility Vehicles and Taxicabs}, 2019 IEEE 11th International Conference on Humanoid, Nanotechnology, Information Technology, Communication and Control, Environment, and Management, HNICEM 2019, no volume, (no pages found) (2019).\\[0.1cm]
\textbf{[S367]} Park S. and Yoo Y., \textit{Real-time scheduling using reinforcement learning technique for the connected vehicles}, IEEE Vehicular Technology Conference, \textbf{2018-June}, pp. 1-5 (2018).\\[0.1cm]
\textbf{[S368]} Pasin M., Seghrouchni A., Belbachir A., Peres S. and Brandao A., \textit{Computational Intelligence and Adaptation in VANETs: Current Research and New Perspectives}, Proceedings of the International Joint Conference on Neural Networks, \textbf{2018-July}, (no pages found) (2018).\\[0.1cm]
\textbf{[S369]} Patrizi N., Fragkos G., Tsiropoulou E. and Papavassiliou S., \textit{Contract-Theoretic Resource Control in Wireless Powered Communication Public Safety Systems}, 2020 IEEE Global Communications Conference, GLOBECOM 2020 - Proceedings, \textbf{2020-January}, (no pages found) (2020).\\[0.1cm]
\textbf{[S370]} Paul A. and Kanaga E., \textit{Query- Focused Content Retrieval from Surveillance videos: Techniques and Challenges}, 2021 7th International Conference on Advanced Computing and Communication Systems, ICACCS 2021, no volume, pp. 1777-1784 (2021).\\[0.1cm]
\textbf{[S371]} Peixoto M., Cruz E., Maia A., Santos M., Lobato W. and Villas L., \textit{Exploiting Fog Computing with an Adapted DBSCAN for Traffic Congestion Detection System}, IEEE Vehicular Technology Conference, \textbf{2020-November}, (no pages found) (2020).\\[0.1cm]
\textbf{[S372]} Peng C., Qi Q., Hu J. and Xie X., \textit{Design of Smart Home Safety System for the Aged Based on ARM}, 2021 IEEE 2nd International Conference on Big Data, Artificial Intelligence and Internet of Things Engineering, ICBAIE 2021, no volume, pp. 727-730 (2021).\\[0.1cm]
\textbf{[S373]} Peng R., Li W., Yang T. and Huafeng K., \textit{An internet of vehicles intrusion detection system based on a convolutional neural network}, Proceedings - 2019 IEEE Intl Conf on Parallel and Distributed Processing with Applications, Big Data and Cloud Computing, Sustainable Computing and Communications, Social Computing and Networking, ISPA/BDCloud/SustainCom/SocialCom 2019, no volume, pp. 1595-1599 (2019).\\[0.1cm]
\textbf{[S374]} Peng, F., et al., \textit{Traffic flow statistics algorithm based on YOLOv3}, 2021 International Conference on Communications, Information System and Computer Engineering (CISCE), IEEE, p. 627–30 (2021).\\[0.1cm]
\textbf{[S375]} Perng J., Lin J., Hsu Y. and Ma L., \textit{Multi-sensor fusion in safety monitoring systems at intersections}, Conference Proceedings - IEEE International Conference on Systems, Man and Cybernetics, \textbf{2014-January(January)}, pp. 2131-2137 (2014).\\[0.1cm]
\textbf{[S376]} Petric P., Hudej R., Al-Hammadi N. and Segedin B., \textit{Virtual modelling of novel applicator prototypes for cervical cancer brachytherapy}, Radiology and Oncology, \textbf{50(4)}, pp. 433-441 (2016).\\[0.1cm]
\textbf{[S377]} Phule S. and Sawant S., \textit{Abnormal activities detection for security purpose unattainded bag and crowding detection by using image processing}, Proceedings of the 2017 International Conference on Intelligent Computing and Control Systems, ICICCS 2017, \textbf{2018-January}, pp. 1069-1073 (2017).\\[0.1cm]
\textbf{[S378]} Potluri S., Henry N. and Diedrich C., \textit{Evaluation of hybrid deep learning techniques for ensuring security in networked control systems}, IEEE International Conference on Emerging Technologies and Factory Automation, ETFA, no volume, pp. 1-8 (2017).\\[0.1cm]
\textbf{[S379]} Pruthvi S., Shama M., Harithas H. and Chiploonkar S., \textit{A Perceptual Field of Vision, Using Image Processing}, 2018 9th International Conference on Computing, Communication and Networking Technologies, ICCCNT 2018, no volume, (no pages found) (2018).\\[0.1cm]
\textbf{[S380]} Pu, C., \textit{A Novel Blockchain-Based Trust Management Scheme for Vehicular Networks}, 2021 Wireless Telecommunications Symposium (WTS), p. 1–6 (2021).\\[0.1cm]
\textbf{[S381]} Pughat A., Pritam A., Sharma D. and Gupta S., \textit{Adaptive Traffic Light Controller for Vehicular Ad-Hoc Networks}, 2019 International Conference on Signal Processing and Communication, ICSC 2019, no volume, pp. 338-344 (2019).\\[0.1cm]
\textbf{[S382]} Qafzezi E., Bylykbashi K., Ampririt P., Ikeda M., Matsuo K. and Barolli L., \textit{A fuzzy-based approach for resource management in SDN-VANETs: Effect of trustworthiness on assessment of available edge computing resources}, Journal of High Speed Networks, \textbf{27(1)}, pp. 33-44 (2021).\\[0.1cm]
\textbf{[S383]} Qi W., Song Q., Wang X., Guo L. and Ning Z., \textit{SDN-Enabled Social-Aware Clustering in 5G-VANET Systems}, IEEE Access, \textbf{6}, pp. 28213-28224 (2018).\\[0.1cm]
\textbf{[S384]} Qi, Q., et al., \textit{Design of Wireless Smart Home Safety System Based on Visual Identity}, 2021 International Conference on Communications, Information System and Computer Engineering (CISCE), IEEE, p. 415–18 (2021).\\[0.1cm]
\textbf{[S385]} Qian M., Gao H. and Liu W., \textit{Android Based Vehicle Anti-Theft Alarm and Tracking System in Hand-Held Communication Terminal}, 2018 IEEE International Conference on Consumer Electronics-Taiwan, ICCE-TW 2018, no volume, (no pages found) (2018).\\[0.1cm]
\textbf{[S386]} Qian Q., Pandiyan A. and Boyle D., \textit{Optimal Recharge Scheduler for Drone-to-Sensor Wireless Power Transfer}, IEEE Access, \textbf{9}, pp. 59301-59312 (2021).\\[0.1cm]
\textbf{[S387]} Qian Y., Qi J., Kuai X., Han G., Sun H. and Hong S., \textit{Specific Emitter Identification Based on Multi-Level Sparse Representation in Automatic Identification System}, IEEE Transactions on Information Forensics and Security, \textbf{16}, pp. 2872-2884 (2021).\\[0.1cm]
\textbf{[S388]} Qian Y., Zhang Y., Fortino G., Miao Y., Hu L. and Hwang K., \textit{Security-Enhanced Content Caching for the 5G-Based Cognitive Internet of Vehicles}, IEEE Network, \textbf{35(2)}, pp. 40-45 (2021).\\[0.1cm]
\textbf{[S389]} Quevedo C., Quevedo A., Campos G., Gomes R., Celestino J. and Serhrouchni A., \textit{An Intelligent Mechanism for Sybil Attacks Detection in VANETs}, IEEE International Conference on Communications, \textbf{2020-June}, (no pages found) (2020).\\[0.1cm]
\textbf{[S390]} Rael K., Fragkos G., Plusquellic J. and Tsiropoulou E., \textit{UAV-enabled Human Internet of Things}, Proceedings - 16th Annual International Conference on Distributed Computing in Sensor Systems, DCOSS 2020, no volume, pp. 312-319 (2020).\\[0.1cm]
\textbf{[S391]} Rahimian P., O'Neal E., Yon J., Franzen L., Jiang Y., Plumert J. and Kearney J., \textit{Using a virtual environment to study the impact of sending traffic alerts to texting pedestrians}, Proceedings - IEEE Virtual Reality, \textbf{2016-July}, pp. 141-149 (2016).\\[0.1cm]
\textbf{[S392]} Raj B. and Chandrasekaran S., \textit{Optimal adaptive data dissemination protocol for vanet road safety using optimal congestion control algorithm}, Recent Advances in Computer Science and Communications, \textbf{13(6)}, pp. 1089-1105 (2020).\\[0.1cm]
\textbf{[S393]} Raja G., Dhanasekaran P., Anbalagan S., Ganapathisubramaniyan A. and Bashir A., \textit{SDN-enabled traffic alert system for IoV in smart cities}, IEEE INFOCOM 2020 - IEEE Conference on Computer Communications Workshops, INFOCOM WKSHPS 2020, no volume, pp. 1093-1098 (2020).\\[0.1cm]
\textbf{[S394]} Raja G., Ganapathisubramaniyan A., Anbalagan S., Baskaran S., Raja K. and Bashir A., \textit{Intelligent Reward-Based Data Offloading in Next-Generation Vehicular Networks}, IEEE Internet of Things Journal, \textbf{7(5)}, pp. 3747-3758 (2020).\\[0.1cm]
\textbf{[S395]} Rajalakshmi S., Angel Deborah S., Soundarya G., Varshitha V. and Shyam Sundhar K., \textit{Safety device for children using iot and deep learning techniques}, Advances in Intelligent Systems and Computing, \textbf{1163}, pp. 375-390 (2021).\\[0.1cm]
\textbf{[S396]} Ramya V. and Akilan T., \textit{Autonomous and voice enabled embedded wheel chair}, ARPN Journal of Engineering and Applied Sciences, \textbf{10(12)}, pp. 5455-5460 (2015).\\[0.1cm]
\textbf{[S397]} Ranka S., Rangarajan A., Elefteriadou L., Srinivasan S., Poasadas E., Hoffman D., Ponnulari R., Dilmore J. and Byron T., \textit{A Vision of Smart Traffic Infrastructure for Traditional, Connected, and Autonomous Vehicles}, Proceedings - 2020 International Conference on Connected and Autonomous Driving, MetroCAD 2020, no volume, pp. 1-8 (2020).\\[0.1cm]
\textbf{[S398]} Rao M., Xu Y. and Peng Z., \textit{INTEMOR real-time intelligent monitoring and incident prevention system - Efficient application of artificial intelligence and computer technology in safety}, Process in Safety Science and Technology Part A, \textbf{3}, pp. 213-220 (2002).\\[0.1cm]
\textbf{[S399]} Rasheed I., Hu F. and Zhang L., \textit{Deep reinforcement learning approach for autonomous vehicle systems for maintaining security and safety using LSTM-GAN}, Vehicular Communications, \textbf{26}, (no pages found) (2020).\\[0.1cm]
\textbf{[S400]} Reebadiya D., Rathod T., Gupta R., Tanwar S. and Kumar N., \textit{Blockchain-based Secure and Intelligent Sensing Scheme for Autonomous Vehicles Activity Tracking Beyond 5G Networks}, Peer-to-Peer Networking and Applications, no volume, (no pages found) (2021).\\[0.1cm]
\textbf{[S401]} Regin R. and Menakadevi T., \textit{Dynamic Clustering Mechanism to Avoid Congestion Control in Vehicular Ad Hoc Networks Based on Node Density}, Wireless Personal Communications, \textbf{107(4)}, pp. 1911-1931 (2019).\\[0.1cm]
\textbf{[S402]} Rehan S. and Singh R., \textit{Industrial and Home Automation, Control, Safety and Security System using Bolt IoT Platform}, Proceedings - International Conference on Smart Electronics and Communication, ICOSEC 2020, no volume, pp. 787-793 (2020).\\[0.1cm]
\textbf{[S403]} Renduli? I., Bibuli? A. and Miškovi? N., \textit{Estimating diver orientation from video using body markers}, 2015 38th International Convention on Information and Communication Technology, Electronics and Microelectronics, MIPRO 2015 - Proceedings, no volume, pp. 1054-1059 (2015).\\[0.1cm]
\textbf{[S404]} Rezgui J., Gagne E., Blain G., Harvey M., St-Pierre O. and Cherkaoui S., \textit{Open Source Platform for Extended Perception Using Communications and Machine Learning on a Small-Scale Vehicular Testbed}, 2020 Global Information Infrastructure and Networking Symposium, GIIS 2020, no volume, (no pages found) (2020).\\[0.1cm]
\textbf{[S405]} Rezgui J., Gagne E., Blain G., St-Pierre O. and Harvey M., \textit{Platooning of Autonomous Vehicles with Artificial Intelligence V2I Communications and Navigation Algorithm}, 2020 Global Information Infrastructure and Networking Symposium, GIIS 2020, no volume, (no pages found) (2020).\\[0.1cm]
\textbf{[S406]} Rihan, M., et al., \textit{Deep-VFog: When Artificial Intelligence Meets Fog Computing in V2X}, IEEE Systems Journal, p. 1–14 (2020).\\[0.1cm]
\textbf{[S407]} Rodríguez-Rodríguez I., Rodríguez J., Elizondo-Moreno A. and Heras-González P., \textit{An autonomous alarm system for personal safety assurance of intimate partner violence survivors based on passive continuous monitoring through biosensor}, Symmetry, \textbf{12(3)}, (no pages found) (2020).\\[0.1cm]
\textbf{[S408]} Roman C., Sapienza M., Ball P., Ou S., Cuzzolin F. and Torr P., \textit{Heterogeneous wireless system testbed for remote image processing in automated vehicles}, 2016 10th International Symposium on Communication Systems, Networks and Digital Signal Processing, CSNDSP 2016, no volume, (no pages found) (2016).\\[0.1cm]
\textbf{[S409]} Rong C., Jin L. and Yu J., \textit{Research on safety driving assistant system of expressway on-ramp merging area based on wireless network communication}, Proceedings of the 2009 International Conference on Machine Learning and Cybernetics, \textbf{6}, pp. 3139-3144 (2009).\\[0.1cm]
\textbf{[S410]} Rosende S., Sánchez-Soriano J., Muñoz C. and Andrés J., \textit{Remote management architecture of uav fleets for maintenance, surveillance, and security tasks in solar power plants}, Energies, \textbf{13(21)}, (no pages found) (2020).\\[0.1cm]
\textbf{[S411]} Rossi G., Fan Z., Chin W. and Leung K., \textit{Stable clustering for Ad-Hoc vehicle networking}, IEEE Wireless Communications and Networking Conference, WCNC, no volume, (no pages found) (2017).\\[0.1cm]
\textbf{[S412]} Rotman N., Schapira M. and Tamar A., \textit{Online Safety Assurance for Learning-Augmented Systems}, HotNets 2020 - Proceedings of the 19th ACM Workshop on Hot Topics in Networks, no volume, pp. 88-95 (2020).\\[0.1cm]
\textbf{[S413]} Roy T., Tariq A. and Dey S., \textit{A Socio-Technical Approach for Resilient Connected Transportation Systems in Smart Cities}, IEEE Transactions on Intelligent Transportation Systems, no volume, (no pages found) (2021).\\[0.1cm]
\textbf{[S414]} Saeed U., Lee Y., Jan S. and Koo I., \textit{CAFD: Context-aware fault diagnostic scheme towards sensor faults utilizing machine learning}, Sensors (Switzerland), \textbf{21(2)}, pp. 1-15 (2021).\\[0.1cm]
\textbf{[S415]} Sahid D. and Alaydrus M., \textit{Multi Sensor Fire Detection in Low Voltage Electrical Panel Using Modular Fuzzy Logic}, 2020 2nd International Conference on Broadband Communications, Wireless Sensors and Powering, BCWSP 2020, no volume, pp. 31-35 (2020).\\[0.1cm]
\textbf{[S416]} Saifan R., Dweik W. and Abdel-Majeed M., \textit{A machine learning based deaf assistance digital system}, Computer Applications in Engineering Education, \textbf{26(4)}, pp. 1008-1019 (2018).\\[0.1cm]
\textbf{[S417]} Salazar E., Azurdia-Meza C., Zabala-Blanco D., Bolufé S. and Soto I., \textit{Semi-supervised extreme learning machine channel estimator and equalizer for vehicle to vehicle communications}, Electronics (Switzerland), \textbf{10(8)}, (no pages found) (2021).\\[0.1cm]
\textbf{[S418]} Sambana B. and Ramesh Y., \textit{An artificial intelligence approach to intelligent vehicle control and monitoring system}, Proceedings - 2020 IEEE International Symposium on Sustainable Energy, Signal Processing and Cyber Security, iSSSC 2020, no volume, (no pages found) (2020).\\[0.1cm]
\textbf{[S419]} Sandino J., Vanegas F., Maire F., Caccetta P., Sanderson C. and Gonzalez F., \textit{UAV framework for autonomous onboard navigation and people/object detection in cluttered indoor environments}, Remote Sensing, \textbf{12(20)}, pp. 1-31 (2020).\\[0.1cm]
\textbf{[S420]} Sankaranarayanan M., Mala C. and Mathew S., \textit{Improved Security Schemes for Efficient Traffic Management in Vehicular Ad-Hoc Network}, Communications in Computer and Information Science, \textbf{1121}, pp. 129-144 (2020).\\[0.1cm]
\textbf{[S421]} Sankhe K., Jaisinghani D. and Chowdhury K., \textit{ReLy: Machine Learning for Ultra-Reliable, Low-Latency Messaging in Industrial Robots}, IEEE Communications Magazine, \textbf{59(4)}, pp. 75-81 (2021).\\[0.1cm]
\textbf{[S422]} Santara A., Rudra S., Buridi S., Kaushik M., Naik A., Kaul B. and Ravindran B., \textit{MADRaS : Multi agent driving simulator}, Journal of Artificial Intelligence Research, \textbf{70}, pp. 1517-1555 (2021).\\[0.1cm]
\textbf{[S423]} Sasaki N., Iijima N. and Uchiyama D., \textit{Development of ranging method for inter-vehicle distance using visible light communication and image processing}, ICCAS 2015 - 2015 15th International Conference on Control, Automation and Systems, Proceedings, no volume, pp. 666-670 (2015).\\[0.1cm]
\textbf{[S424]} Sassi M. and Fourati L., \textit{Investigation on Deep Learning Methods for Privacy and Security Challenges of Cognitive IoV}, 2020 International Wireless Communications and Mobile Computing, IWCMC 2020, no volume, pp. 714-720 (2020).\\[0.1cm]
\textbf{[S425]} Sawabe A., Iwai T., Satoda K. and Nakao A., \textit{Edge Concierge: Democratizing Cost-Effective and Flexible Network Operations using Network Layer AI at Private Network Edges}, Proceedings of IEEE/IFIP Network Operations and Management Symposium 2020: Management in the Age of Softwarization and Artificial Intelligence, NOMS 2020, no volume, (no pages found) (2020).\\[0.1cm]
\textbf{[S426]} Scazzoli D., Magarini M., Reggiani L., Moullec Y. and Mahtab Alam M., \textit{A deep learning approach for LoS/NLoS identification via PRACH in UAV-assisted public safety networks}, IEEE International Symposium on Personal, Indoor and Mobile Radio Communications, PIMRC, \textbf{2020-August}, (no pages found) (2020).\\[0.1cm]
\textbf{[S427]} Sellami L. and Alaya B., \textit{SAMNET: Self-adaptative multi-kernel clustering algorithm for urban VANETs}, Vehicular Communications, \textbf{29}, (no pages found) (2021).\\[0.1cm]
\textbf{[S428]} Semanjski S., Semanjski I., De Wilde W. and Gautama S., \textit{GNSS spoofing detection by supervised machine learning with validation on real-world meaconing and spoofing data—part II}, Sensors (Switzerland), \textbf{20(7)}, (no pages found) (2020).\\[0.1cm]
\textbf{[S429]} Semanjski, S., et al., \textit{Use of Supervised Machine Learning for GNSS Signal Spoofing Detection with Validation on Real-World Meaconing and Spoofing Data—Part I},  Sensors (Basel, Switzerland), vol. 20, no 4, p. 1171 (2020).\\[0.1cm]
\textbf{[S430]} Sene A., Kamsu-Foguem B. and Rumeau P., \textit{Decision support system for in-flight emergency events}, Cognition, Technology and Work, \textbf{20(2)}, pp. 245-266 (2018).\\[0.1cm]
\textbf{[S431]} Shan L., Miura R., Kagawa T., Ono F., Li H. and Kojima F., \textit{Machine Learning-Based Field Data Analysis and Modeling for Drone Communications}, IEEE Access, \textbf{7}, pp. 79127-79135 (2019).\\[0.1cm]
\textbf{[S432]} ShangGuan W., Shi B., Cai B., Wang J. and Zang Y., \textit{Multiple V2V communication mode competition method in cooperative vehicle infrastructure system}, IEEE Conference on Intelligent Transportation Systems, Proceedings, ITSC, no volume, pp. 1200-1205 (2016).\\[0.1cm]
\textbf{[S433]} Sharma A., Awasthi Y. and Kumar S., \textit{The Role of Blockchain, AI and IoT for Smart Road Traffic Management System}, Proceedings - 2020 IEEE India Council International Subsections Conference, INDISCON 2020, no volume, pp. 289-296 (2020).\\[0.1cm]
\textbf{[S434]} Sharma P. and Liu H., \textit{A Machine-Learning-Based Data-Centric Misbehavior Detection Model for Internet of Vehicles}, IEEE Internet of Things Journal, \textbf{8(6)}, pp. 4991-4999 (2021).\\[0.1cm]
\textbf{[S435]} Sharma P., Liu H., Honggang W. and Shelley Z., \textit{Securing wireless communications of connected vehicles with artificial intelligence}, 2017 IEEE International Symposium on Technologies for Homeland Security, HST 2017, no volume, (no pages found) (2017).\\[0.1cm]
\textbf{[S436]} Sharma P., Siddanagaiah U. and Kul G., \textit{Towards an AI-based after-collision forensic analysis protocol for autonomous vehicles}, Proceedings - 2020 IEEE Symposium on Security and Privacy Workshops, SPW 2020, no volume, pp. 240-243 (2020).\\[0.1cm]
\textbf{[S437]} Sharma S. and Singh B., \textit{Cooperative Reinforcement Learning Based Adaptive Resource Allocation in V2V Communication}, 2019 6th International Conference on Signal Processing and Integrated Networks, SPIN 2019, no volume, pp. 489-494 (2019).\\[0.1cm]
\textbf{[S438]} Sharma V., Kim J., Kwon S., You I. and Chen H., \textit{Fuzzy-based protocol for secure remote diagnosis of IoT devices in 5G networks}, Lecture Notes of the Institute for Computer Sciences, Social-Informatics and Telecommunications Engineering, LNICST, \textbf{246}, pp. 54-63 (2018).\\[0.1cm]
\textbf{[S439]} Shen L., Zhang Q., Cao G. and Xu H., \textit{Fall detection system based on deep learning and image processing in cloud environment}, Advances in Intelligent Systems and Computing, \textbf{772}, pp. 590-598 (2019).\\[0.1cm]
\textbf{[S440]} Shen X., Mark J. and Ye J., \textit{Mobile location estimation in CDMA cellular networks by using fuzzy logic}, Wireless Personal Communications, \textbf{22(1)}, pp. 57-70 (2002).\\[0.1cm]
\textbf{[S441]} Shi D., Lu J., Wang J., Li L., Liu K. and Pan M., \textit{No One Left Behind: Avoid Hot Car Deaths via WiFi Detection}, IEEE International Conference on Communications, \textbf{2020-June}, (no pages found) (2020).\\[0.1cm]
\textbf{[S442]} Shi Y., Peng X., Shen H. and Bai G., \textit{Cluster-based cooperative data service for VANETs}, Lecture Notes of the Institute for Computer Sciences, Social-Informatics and Telecommunications Engineering, LNICST, \textbf{230}, pp. 119-129 (2018).\\[0.1cm]
\textbf{[S443]} Shrestha K., Shrestha P. and Yfantis E., \textit{Framework development for construction safety visualization}, Proceedings, Annual Conference - Canadian Society for Civil Engineering, \textbf{2(January)}, pp. 1050-1059 (2013).\\[0.1cm]
\textbf{[S444]} Shri S. and Jothilakshmi S., \textit{Image processing based fire detection and alerting system for crowd monitoring}, Journal of Advanced Research in Dynamical and Control Systems, \textbf{11(2 Special Issue)}, pp. 280-286 (2019).\\[0.1cm]
\textbf{[S445]} Siddiqui S., Mahmood A., Zhang W. and Sheng Q., \textit{Machine learning based trust model for misbehaviour detection in internet-of-vehicles}, Communications in Computer and Information Science, \textbf{1142 CCIS}, pp. 512-520 (2019).\\[0.1cm]
\textbf{[S446]} Siewert S., Sampigethaya K., Buchholz J. and Rizor S., \textit{Fail-Safe, Fail-Secure Experiments for Small UAS and UAM Traffic in Urban Airspace}, AIAA/IEEE Digital Avionics Systems Conference - Proceedings, \textbf{2019-September}, (no pages found) (2019).\\[0.1cm]
\textbf{[S447]} Sigamani R. and Ganapathi P., \textit{GOF-SLFN- An Intelligent Attack Detection System against Denial of Service (DoS) attacks based on Glow Worm Swarm optimized Single Layer Feed Forward Networks for vehicular Cyber Physical Systems (VCPS)}, IOP Conference Series: Materials Science and Engineering, vol. 925, p. 012001 (2020). \\[0.1cm]
\textbf{[S448]} Silva A., Bunyakitanon M., Vassallo R., Nejabati R. and Simeonidou D., \textit{An Open 5G NFV Platform for Smart City Applications Using Network Softwarization}, 2019 IEEE Wireless Communications and Networking Conference Workshop, WCNCW 2019, no volume, (no pages found) (2019).\\[0.1cm]
\textbf{[S449]} Silva M., Souza E., Alsina P., Francisco H., Medeiros A., Nogueira M., De Alburquerque G. and Dantas J., \textit{Communication network architecture specification for multi-UAV system applied to scanning rocket impact area first results}, Proceedings - 2017 LARS 14th Latin American Robotics Symposium and 2017 5th SBR Brazilian Symposium on Robotics, LARS-SBR 2017 - Part of the Robotics Conference 2017, \textbf{2017-December}, pp. 1-6 (2017).\\[0.1cm]
\textbf{[S450]} Sinaeepourfard A., Sengupta S., Krogstie J. and Delgado R., \textit{Cybersecurity in Large-Scale Smart Cities: Novel Proposals for Anomaly Detection from Edge to Cloud}, 2019 International Conference on Internet of Things, Embedded Systems and Communications, IINTEC 2019 - Proceedings, no volume, pp. 130-135 (2019).\\[0.1cm]
\textbf{[S451]} Singh G., Kumar P., Mishra R., Sharma S. and Singh K., \textit{Security System for Railway Crossings using Machine Learning}, Proceedings - IEEE 2020 2nd International Conference on Advances in Computing, Communication Control and Networking, ICACCCN 2020, no volume, pp. 135-139 (2020).\\[0.1cm]
\textbf{[S452]} Singh K. and Sood S., \textit{Optical fog-assisted cyber-physical system for intelligent surveillance in the education system}, Computer Applications in Engineering Education, \textbf{28(3)}, pp. 692-704 (2020).\\[0.1cm]
\textbf{[S453]} Singkang L., Ping K., Kunsei H., Senthilkumar K., Pirapaharan K., Haidar A. and Hoole P., \textit{Model based-testing of spatial and time domain artificial intelligence smart antenna for ultra-high frequency electric discharge detection in digital power substations}, Progress In Electromagnetics Research M, \textbf{99}, pp. 91-101 (2021).\\[0.1cm]
\textbf{[S454]} Sodhro A., Sodhro G., Guizani M., Pirbhulal S. and Boukerche A., \textit{AI-Enabled Reliable Channel Modeling Architecture for Fog Computing Vehicular Networks}, IEEE Wireless Communications, \textbf{27(2)}, pp. 14-21 (2020).\\[0.1cm]
\textbf{[S455]} Sohn J., Chillakuru Y., Lee S., Lee A., Kelil T., Hess C., Seo Y., Vu T. and Joe B., \textit{An Open-Source, Vender Agnostic Hardware and Software Pipeline for Integration of Artificial Intelligence in Radiology Workflow}, Journal of Digital Imaging, \textbf{33(4)}, pp. 1041-1046 (2020).\\[0.1cm]
\textbf{[S456]} Soleymani S., Anisi M., Abdullah A., Ngadi M., Goudarzi S., Khan M. and Kama M., \textit{An authentication and plausibility model for big data analytic under LOS and NLOS conditions in 5G-VANET}, Science China Information Sciences, \textbf{63(12)}, (no pages found) (2020).\\[0.1cm]
\textbf{[S457]} Song L., Sun G., Yu H., Du X. and Guizani M., \textit{FBIA: A Fog-Based Identity Authentication Scheme for Privacy Preservation in Internet of Vehicles}, IEEE Transactions on Vehicular Technology, \textbf{69(5)}, pp. 5403-5415 (2020).\\[0.1cm]
\textbf{[S458]} Sonker A. and Gupta R., \textit{A new procedure for misbehavior detection in vehicular ad-hoc networks using machine learning}, International Journal of Electrical and Computer Engineering, \textbf{11(3)}, pp. 2535-2547 (2021).\\[0.1cm]
\textbf{[S459]} Sonker, A. and Gupta, R. K., \textit{A New Combination of Machine Learning Algorithms using Stacking Approach for Misbehavior Detection in VANETs}, International Journal of Computer Science and Network Security, vol. 20, no 10, p. 94–100 (2020).\\[0.1cm]
\textbf{[S460]} Soomro, A. H. and Jilani, M. T., \textit{Application of IoT and Artificial Neural Networks (ANN) for Monitoring of Underground Coal Mines}, 2020 International Conference on Information Science and Communication Technology (ICISCT), IEEE, p. 1–8 (2020).\\[0.1cm]
\textbf{[S461]} Sophia Jasmine G., Magdalin Mary D., Naveen S., Murugan V., Mohamed Ibrahim A. and Praveen S., \textit{Load control using projected VR system of wallmounted buttons}, 2021 7th International Conference on Advanced Computing and Communication Systems, ICACCS 2021, no volume, pp. 1155-1159 (2021).\\[0.1cm]
\textbf{[S462]} Srivastava D., Shaikh S. and Shah P., \textit{Automatic traffic surveillance system Utilizing object detection and image processing}, 2021 International Conference on Computer Communication and Informatics, ICCCI 2021, no volume, (no pages found) (2021).\\[0.1cm]
\textbf{[S463]} Su H., Zhang X. and Chen H., \textit{Cluster-based DSRC architecture for QoS provisioning over vehicle ad hoc networks}, GLOBECOM - IEEE Global Telecommunications Conference, no volume, (no pages found) (2006).\\[0.1cm]
\textbf{[S464]} Su K., Lee K., Huang P. and Chen I., \textit{Developing a usable mobile flight case learning system in air traffic control miscommunications}, Lecture Notes in Computer Science (including subseries Lecture Notes in Artificial Intelligence and Lecture Notes in Bioinformatics), \textbf{5613 LNCS(PART 4)}, pp. 770-777 (2009).\\[0.1cm]
\textbf{[S465]} Šubik S., Kaulbars D., Bök P. and Wietfeld C., \textit{Dynamic link classification based on neuronal networks for QoS enabled access to limited resources}, Proceedings - International Conference on Computer Communications and Networks, ICCCN, no volume, (no pages found) (2013).\\[0.1cm]
\textbf{[S466]} Sumalatha, M. S. and Nandalal, V., \textit{An intelligent cross layer security based fuzzy trust calculation mechanism (CLS-FTCM) for securing wireless sensor network (WSN)}, Journal of Ambient Intelligence and Humanized Computing, vol. 12, no 5, p. 4559–73 (2021).\\[0.1cm]
\textbf{[S467]} Sun H., Yu Y., Sha K. and Lou B., \textit{MVideo: Edge Computing Based Mobile Video Processing Systems}, IEEE Access, \textbf{8}, pp. 11615-11623 (2020).\\[0.1cm]
\textbf{[S468]} Sun P. and Boukerche A., \textit{A Novel Internet-of-Vehicles Assisted Collaborative Low-visible Pedestrian Detection Approach}, 2020 IEEE Global Communications Conference, GLOBECOM 2020 - Proceedings, no volume, (no pages found) (2020).\\[0.1cm]
\textbf{[S469]} Sun Y., Xu D., Huang Z., Zhang H. and Liang X., \textit{LIDAUS: Localization of IoT Device via Anchor UAV SLAM}, 2020 IEEE 39th International Performance Computing and Communications Conference, IPCCC 2020, no volume, (no pages found) (2020).\\[0.1cm]
\textbf{[S470]} Sun, P. and Boukerche, A., \textit{AI ?assisted Data Dissemination Methods for Supporting Intelligent Transportation Systems}, Internet Technology Letters, vol. 4, no 1, (2021).\\[0.1cm]
\textbf{[S471]} Swarnamugi M. and Chinnaiyan R., \textit{Context—aware smart reliable service model for intelligent transportation system based on ontology}, Lecture Notes in Electrical Engineering, \textbf{597}, pp. 23-30 (2020).\\[0.1cm]
\textbf{[S472]} Tao L., Hong T., Guo Y., Chen H. and Zhang J., \textit{Drone identification based on CenterNet-TensorRT}, IEEE International Symposium on Broadband Multimedia Systems and Broadcasting, BMSB, \textbf{2020-October}, (no pages found) (2020).\\[0.1cm]
\textbf{[S473]} Tayeh G., Azar J., Makhoul A., Guyeux C. and Demerjian J., \textit{A Wearable LoRa-Based Emergency System for Remote Safety Monitoring}, 2020 International Wireless Communications and Mobile Computing, IWCMC 2020, no volume, pp. 120-125 (2020).\\[0.1cm]
\textbf{[S474]} Tayyaba S., Khattak H., Almogren A., Shah M., Ud Din I., Alkhalifa I. and Guizani M., \textit{5G vehicular network resource management for improving radio access through machine learning}, IEEE Access, \textbf{8}, pp. 6792-6800 (2020).\\[0.1cm]
\textbf{[S475]} Thakur Y. and Sakravdia D., \textit{An improved reliability on wireless sensor network for energy irregularity models using fuzzy deep learning protocol: Fdlp}, International Journal of Scientific and Technology Research, \textbf{9(1)}, pp. 4384-4389 (2020).\\[0.1cm]
\textbf{[S476]} Tian B., Wang G., Xu Z., Zhang Y. and Zhao X., \textit{Communication delay compensation for string stability of CACC system using LSTM prediction}, Vehicular Communications, \textbf{29}, (no pages found) (2021).\\[0.1cm]
\textbf{[S477]} Toh C., Cano J., Fernandez-Laguia C., Manzoni P. and Calafate C., \textit{Wireless digital traffic signs of the future}, IET Networks, \textbf{8(1)}, pp. 74-78 (2019).\\[0.1cm]
\textbf{[S478]} Torres-Figueroa L., Schepker H. and Jiru J., \textit{QoS Evaluation and Prediction for C-V2X Communication in Commercially-Deployed LTE and Mobile Edge Networks}, IEEE Vehicular Technology Conference, \textbf{2020-May}, (no pages found) (2020).\\[0.1cm]
\textbf{[S479]} Toulni H. and Nsiri B., \textit{A Hybrid Routing Protocol for VANET Using Ontology}, Procedia Computer Science, \textbf{73}, pp. 94-101 (2015).\\[0.1cm]
\textbf{[S480]} Toulni H. and Nsiri B., \textit{Cluster-based routing protocol using traffic information}, International Journal of High Performance Computing and Networking, \textbf{11(2)}, pp. 108-116 (2018).\\[0.1cm]
\textbf{[S481]} Trinh H., Giupponi L. and Dini P., \textit{Urban Anomaly Detection by processing Mobile Traffic Traces with LSTM Neural Networks}, Annual IEEE Communications Society Conference on Sensor, Mesh and Ad Hoc Communications and Networks workshops, \textbf{2019-June}, (no pages found) (2019).\\[0.1cm]
\textbf{[S482]} Tsai, M., \textit{Streamlining Information Representation during Construction Accidents}, KSCE Journal of Civil Engineering, vol. 18, no 7, p. 1945–54 (2014). \\[0.1cm]
\textbf{[S483]} Tseng Y., Hsu P., Chen J. and Tseng Y., \textit{Computer Vision-Assisted Instant Alerts in 5G}, Proceedings - International Conference on Computer Communications and Networks, ICCCN, \textbf{2020-August}, (no pages found) (2020).\\[0.1cm]
\textbf{[S484]} Tseng Y., Hung T., Pan C. and Wu R., \textit{Motion control system of unmanned railcars based on image recognition}, Applied System Innovation, \textbf{2(1)}, pp. 1-16 (2019).\\[0.1cm]
\textbf{[S485]} Tung L., Mena J., Gerla M. and Sommer C., \textit{A cluster based architecture for intersection collision avoidance using heterogeneous networks}, 12th Annual Mediterranean Ad Hoc Networking Workshop, MED-HOC-NET 2013, no volume, pp. 82-88 (2013).\\[0.1cm]
\textbf{[S486]} Turcanu I., Sommer C., Baiocchi A. and Dressler F., \textit{Pick the right guy: CQI-based LTE forwarder selection in VANETs}, IEEE Vehicular Networking Conference, VNC, \textbf{0}, (no pages found) (2016).\\[0.1cm]
\textbf{[S487]} Ucar S., Ergen S. and Ozkasap O., \textit{Multihop-Cluster-Based IEEE 802.11p and LTE Hybrid Architecture for VANET Safety Message Dissemination}, IEEE Transactions on Vehicular Technology, \textbf{65(4)}, pp. 2621-2636 (2016).\\[0.1cm]
\textbf{[S488]} Ulema, M., \textit{Fundamentals of Public Safety Networks and Critical Communications Systems: Technologies, Deployment, and Management}, 1o ed, Wiley (2018).\\[0.1cm]
\textbf{[S489]} Ullah, M. A., et. al., \textit{Social Networks of Things for Smart Homes Using Fuzzy Logic }, International Journal of Computer Science and Network Security, vol. 18, no 2, p. 168–173 (2020).\\[0.1cm]
\textbf{[S490]} Uprety A., Rawat D. and Li J., \textit{Privacy preserving misbehavior detection in IoV using federated machine learning}, 2021 IEEE 18th Annual Consumer Communications and Networking Conference, CCNC 2021, no volume, (no pages found) (2021).\\[0.1cm]
\textbf{[S491]} Valle F., Cespedes S. and Hafid A., \textit{Automated Decision System to Exploit Network Diversity for Connected Vehicles}, IEEE Transactions on Vehicular Technology, \textbf{70(1)}, pp. 858-871 (2021).\\[0.1cm]
\textbf{[S492]} Van Willigen W., Neef R., Van Lieburg A. and Schut M., \textit{WILLEM: A Wireless InteLLigent Evacuation Method}, Proceedings - 2009 3rd International Conference on Sensor Technologies and Applications, SENSORCOMM 2009, no volume, pp. 382-387 (2009).\\[0.1cm]
\textbf{[S493]} Veneeswari J. and Balasubramanian C., \textit{An urban road traffic with the dedicated fuzzy control system in vanet}, ARPN Journal of Engineering and Applied Sciences, \textbf{11(2)}, pp. 1171-1176 (2016).\\[0.1cm]
\textbf{[S494]} Verde S., Marcon M., Milani S. and Tubaro S., \textit{Advanced assistive maintenance based on augmented reality and 5g networking}, Sensors (Switzerland), \textbf{20(24)}, pp. 1-16 (2020).\\[0.1cm]
\textbf{[S495]} Vidal V., Honório L., Santos M., Silva M., Cerqueira A. and Oliveira E., \textit{UAV vision aided positioning system for location and landing}, 2017 18th International Carpathian Control Conference, ICCC 2017, no volume, pp. 228-233 (2017).\\[0.1cm]
\textbf{[S496]} Vilaplana M. and Goodchild C., \textit{Application of distributed artificial intelligence in autonomous aircraft operations}, AIAA/IEEE Digital Avionics Systems Conference - Proceedings, \textbf{2}, pp. 7B31-7B314 (2001).\\[0.1cm]
\textbf{[S497]} Vogiatzaki M., Zerefos S. and Tania M., \textit{Enhancing city sustainability through smart technologies: A framework for automatic pre-emptive action to promote safety and security using lighting and ICT-based surveillance}, Sustainability (Switzerland), \textbf{12(15)}, (no pages found) (2020).\\[0.1cm]
\textbf{[S498]} Wahab O., Otrok H. and Mourad A., \textit{VANET QoS-OLSR: QoS-based clustering protocol for Vehicular Ad hoc Networks}, Computer Communications, \textbf{36(13)}, pp. 1422-1435 (2013).\\[0.1cm]
\textbf{[S499]} Wan S., Gu Z. and Ni Q., \textit{Cognitive computing and wireless communications on the edge for healthcare service robots}, Computer Communications, \textbf{149}, pp. 99-106 (2020).\\[0.1cm]
\textbf{[S500]} Wan, L., et al., \textit{Deep Learning Based Autonomous Vehicle Super Resolution DOA Estimation for Safety Driving}, IEEE Transactions on Intelligent Transportation Systems, vol. 22, no 7, p. 4301–15 (2021).\\[0.1cm]
\textbf{[S501]} Wang J., Chen L., Wang J., Gao Z., Dong J. and Yan X., \textit{A wireless localization algorithm with bpnn-mea-qpso for connected vehicle}, IEEE Intelligent Transportation Systems Magazine, \textbf{11(2)}, pp. 96-109 (2019).\\[0.1cm]
\textbf{[S502]} Wang J., Lv T., Huang P. and Mathiopoulos P., \textit{Mobility-aware partial computation offloading in vehicular networks: A deep reinforcement learning based scheme}, China Communications, \textbf{17(10)}, pp. 31-49 (2020).\\[0.1cm]
\textbf{[S503]} Wang J., Yue X., Liu Y., Song H., Yuan J., Yang T. and Seker R., \textit{Integrating ground surveillance with aerial surveillance for enhanced amateur drone detection}, Proceedings of SPIE - The International Society for Optical Engineering, \textbf{10652}, (no pages found) (2018).\\[0.1cm]
\textbf{[S504]} Wang L., Yang C., Yu Z., Liu Y., Wang Z. and Guo B., \textit{CrackSense: A CrowdSourcing based urban road crack detection system}, Proceedings - 2019 IEEE SmartWorld, Ubiquitous Intelligence and Computing, Advanced and Trusted Computing, Scalable Computing and Communications, Internet of People and Smart City Innovation, SmartWorld 2019, no volume, pp. 944-951 (2019).\\[0.1cm]
\textbf{[S505]} Wang S., Yang S. and Zhao C., \textit{SurveilEdge: Real-time Video Query based on Collaborative Cloud-Edge Deep Learning}, Proceedings - IEEE INFOCOM, \textbf{2020-July}, pp. 2519-2528 (2020).\\[0.1cm]
\textbf{[S506]} Wang X., Zhou Z., Zhao Y., Zhang X., Xing K., Xiao F., Yang Z. and Liu Y., \textit{Improving Urban Crowd Flow Prediction on Flexible Region Partition}, IEEE Transactions on Mobile Computing, \textbf{19(12)}, pp. 2804-2817 (2020).\\[0.1cm]
\textbf{[S507]} Wang Y., Pan B., Li Y., Huang F. and Guo Z., \textit{Real-time dynamic logic loop mechanism of vehicular network management based on instantaneous load sensing}, Proceedings - 2019 12th International Symposium on Computational Intelligence and Design, ISCID 2019, no volume, pp. 27-30 (2019).\\[0.1cm]
\textbf{[S508]} Watta P., Zhang X. and Murphey Y., \textit{Vehicle Position and Context Detection using V2V Communication}, IEEE Transactions on Intelligent Vehicles, no volume, (no pages found) (2020).\\[0.1cm]
\textbf{[S509]} Wei X., Jiang S., Li Y., Li C., Jia L. and Li Y., \textit{Defect detection of pantograph slide based on deep learning and image processing technology}, IEEE Transactions on Intelligent Transportation Systems, \textbf{21(3)}, pp. 947-958 (2020).\\[0.1cm]
\textbf{[S510]} Weng M., Chen C. and Kao H., \textit{Remote surveillance system for driver drowsiness in real-time using low-cost embedded platform}, Proceedings of the 2008 IEEE International Conference on Vehicular Electronics and Safety, ICVES 2008, no volume, pp. 288-292 (2008).\\[0.1cm]
\textbf{[S511]} Wu C., Wu Y., Zhu C., Zhang C. and Song Q., \textit{Research on mobile intelligent mine platform based on risk control}, IOP Conference Series: Earth and Environmental Science, \textbf{601(1)}, (no pages found) (2020).\\[0.1cm]
\textbf{[S512]} Wu Z., Qiu K. and Gao H., \textit{Driving policies of V2X autonomous vehicles based on reinforcement learning methods}, IET Intelligent Transport Systems, \textbf{14(5)}, pp. 331-337 (2020).\\[0.1cm]
\textbf{[S513]} Xiao H., Zhang W., Li W., Chronopoulos A. and Zhang Z., \textit{Joint Clustering and Blockchain for Real-time Information Security Transmission at the Crossroads in C-V2X Networks}, IEEE Internet of Things Journal, no volume, (no pages found) (2021).\\[0.1cm]
\textbf{[S514]} Xiao J. and Feng H., \textit{A low-cost extendable framework for embedded smart car security system}, Proceedings of the 2009 IEEE International Conference on Networking, Sensing and Control, ICNSC 2009, no volume, pp. 829-833 (2009).\\[0.1cm]
\textbf{[S515]} Xie B., Li K., Qin X., Yang H. and Wang J., \textit{Approaching index based collision avoidance for V2V cooperative systems}, 2014 17th IEEE International Conference on Intelligent Transportation Systems, ITSC 2014, no volume, pp. 127-132 (2014).\\[0.1cm]
\textbf{[S516]} Xing W., Wang N., Wang C., Liu F. and Ji Y., \textit{Resource allocation schemes for D2D communication used in VANETs}, IEEE Vehicular Technology Conference, no volume, (no pages found) (2014).\\[0.1cm]
\textbf{[S517]} Xiong G., Zhang D. and Xu J., \textit{An Effective Data Filtering Scheme Based on Context in VANETs}, International Conference on Communication Technology Proceedings, ICCT, \textbf{2020-October}, pp. 752-756 (2020).\\[0.1cm]
\textbf{[S518]} Xu Y. and Wang W., \textit{A Security Situational Awareness Method for Cloud Platform}, 2019 IEEE 5th International Conference on Computer and Communications, ICCC 2019, no volume, pp. 1927-1934 (2019).\\[0.1cm]
\textbf{[S519]} Xu Y., Yan X., Wu Y., Hu Y., Liang W. and Zhang J., \textit{Hierarchical Bidirectional RNN for Safety-enhanced B5G Heterogeneous Networks}, IEEE Transactions on Network Science and Engineering, no volume, (no pages found) (2021).\\[0.1cm]
\textbf{[S520]} Xu Y., Zhou H., Chen J., Qian B., Zhuang W. and Shen S., \textit{V2X Empowered Non-Signalized Intersection Management in the AI Era: Opportunities and Solutions}, IEEE Communications Standards Magazine, \textbf{4(4)}, pp. 18-25 (2020).\\[0.1cm]
\textbf{[S521]} Xu Y., Zhou H., Ma T., Zhao J., Qian B. and Shen S., \textit{Leveraging Multi-Agent Learning for Automated Vehicles Scheduling at Non-signalized Intersections}, IEEE Internet of Things Journal, no volume, (no pages found) (2021).\\[0.1cm]
\textbf{[S522]} Xu, Q., et al., \textit{Wireless AI in Smart Car: How Smart a Car Can Be?}, IEEE Access, vol. 8, p. 55091–112 (2020).\\[0.1cm]
\textbf{[S523]} Xu, Y., et al., \textit{Deep Deterministic Policy Gradient (DDPG)-Based Resource Allocation Scheme for NOMA Vehicular Communications}, IEEE Access, vol. 8, p. 18797–807 (2020).\\[0.1cm]
\textbf{[S524]} Yanbo W., Bizu B. and Praveena V., \textit{Deep learning based smart monitoring of indoor stadium video surveillance}, Journal of Multiple-Valued Logic and Soft Computing, \textbf{36(1)}, pp. 151-167 (2021).\\[0.1cm]
\textbf{[S525]} Yang A., Zhang C., Chen Y., Zhuansun Y. and Liu H., \textit{Security and Privacy of Smart Home Systems Based on the Internet of Things and Stereo Matching Algorithms}, IEEE Internet of Things Journal, \textbf{7(4)}, pp. 2521-2530 (2020).\\[0.1cm]
\textbf{[S526]} Yang H., Sun Z., Jiang G., Zhao F., Lu X. and Mei X., \textit{Cloud-Manufacturing-Based Condition Monitoring Platform with 5G and Standard Information Model}, IEEE Internet of Things Journal, \textbf{8(8)}, pp. 6940-6948 (2021).\\[0.1cm]
\textbf{[S527]} Yang Q., Jang S. and Yoo S., \textit{Q-Learning-Based Fuzzy Logic for Multi-objective Routing Algorithm in Flying Ad Hoc Networks}, Wireless Personal Communications, \textbf{113(1)}, pp. 115-138 (2020).\\[0.1cm]
\textbf{[S528]} Yang S., Bailey E., Yang Z., Ostrometzky J., Zussman G., Seskar I. and Kostic Z., \textit{COSMOS Smart Intersection: Edge Compute and Communications for Bird's Eye Object Tracking}, 2020 IEEE International Conference on Pervasive Computing and Communications Workshops, PerCom Workshops 2020, no volume, (no pages found) (2020).\\[0.1cm]
\textbf{[S529]} Yang T., Han C., Qin M. and Huang C., \textit{Learning-Aided Intelligent Cooperative Collision Avoidance Mechanism in Dynamic Vessel Networks}, IEEE Transactions on Cognitive Communications and Networking, \textbf{6(1)}, pp. 74-82 (2020).\\[0.1cm]
\textbf{[S530]} Yang X., Zhou N., Liu Y., Quan W., Lu X. and Zhang W., \textit{Online Pantograph-Catenary Contact Point Detection in Complicated Background Based on Multiple Strategies}, IEEE Access, no volume, (no pages found) (2020).\\[0.1cm]
\textbf{[S531]} Yang Y., He D., Kumar N. and Zeadally S., \textit{Compact Hardware Implementation of a SHA-3 Core for Wireless Body Sensor Networks}, IEEE Access, \textbf{6}, pp. 40128-40136 (2018).\\[0.1cm]
\textbf{[S532]} Yang Z. and Zhu M., \textit{Integration method for wireless communication modes in Internet of Vehicles in the big data environment}, Proceedings - 2017 International Conference on Smart Grid and Electrical Automation, ICSGEA 2017, \textbf{2017-January}, pp. 602-607 (2017).\\[0.1cm]
\textbf{[S533]} Yang, Q., et al., \textit{Machine-Learning-Enabled Cooperative Perception for Connected Autonomous Vehicles: Challenges and Opportunities}, IEEE Network, vol. 35, no 3, p. 96–101 (2021).\\[0.1cm]
\textbf{[S534]} Yankson, B. \textit{Autonomous Vehicle Security Through Privacy Integrated Context Ontology(PICO)}, 2020 IEEE International Conference on Systems, Man, and Cybernetics (SMC), IEEE, p. 4372–78 (2020).\\[0.1cm]
\textbf{[S535]} Yasuhara Y., Tanioka T., Kai Y., Tsujigami Y., Uematsu K., Dino M., De Castro Locsin R. and Schoenhofer S., \textit{Potential legal issues when caring healthcare robot with communication in caring functions are used for older adult care}, Enfermeria clinica, \textbf{30}, pp. 54-59 (2020).\\[0.1cm]
\textbf{[S536]} "Yaswanth B., Darshan R., Pavan H., Srinivasa D. and Venkatesh Murthy B., \textit{Smart Safety and Security Solution for Women using kNN Algorithm and IoT}, MPCIT 2020 - Proceedings: IEEE 3rd International Conference on ""Multimedia Processing, Communication and Information Technology"", no volume, pp. 87-92 (2020)."\\[0.1cm]
\textbf{[S537]} Ye J., Zou J., Gao J., Zhang G., Kong M., Pei Z. and Cui K., \textit{A New Frequency Hopping Signal Detection of Civil UAV Based on Improved K-Means Clustering Algorithm}, IEEE Access, \textbf{9}, pp. 53190-53204 (2021).\\[0.1cm]
\textbf{[S538]} Yu K., Lin L., Alazab M., Tan L. and Gu B., \textit{Deep Learning-Based Traffic Safety Solution for a Mixture of Autonomous and Manual Vehicles in a 5G-Enabled Intelligent Transportation System}, IEEE Transactions on Intelligent Transportation Systems, no volume, (no pages found) (2020).\\[0.1cm]
\textbf{[S539]} Yu S., Yuanbo Y., He X., Lu M., Wang P., An X. and Fang X., \textit{On-Board Fast and Intelligent Perception of Ships with the 'Jilin-1' Spectrum 01/02 Satellites}, IEEE Access, \textbf{8}, pp. 48005-48014 (2020).\\[0.1cm]
\textbf{[S540]} Yuan Q., Li J., Zhou H., Lin T., Luo G. and Shen X., \textit{A Joint Service Migration and Mobility Optimization Approach for Vehicular Edge Computing}, IEEE Transactions on Vehicular Technology, \textbf{69(8)}, pp. 9041-9052 (2020).\\[0.1cm]
\textbf{[S541]} Yuan Q., Li J., Zhou H., Luo G., Lin T., Yang F. and Shen X., \textit{Cross-Domain Resource Orchestration for the Edge-Computing-Enabled Smart Road}, IEEE Network, \textbf{34(5)}, pp. 60-67 (2020).\\[0.1cm]
\textbf{[S542]} Yue X., Liu Y., Wang J., Song H. and Cao H., \textit{Software Defined Radio and Wireless Acoustic Networking for Amateur Drone Surveillance}, IEEE Communications Magazine, \textbf{56(4)}, pp. 90-97 (2018).\\[0.1cm]
\textbf{[S543]} Zadobrischi E. and Dimian M., \textit{Inter-urban analysis of pedestrian and drivers through a vehicular network based on hybrid communications embedded in a portable car system and advanced image processing technologies}, Remote Sensing, \textbf{13(7)}, (no pages found) (2021).\\[0.1cm]
\textbf{[S544]} Zhang H., Bochem A., Sun X. and Hogrefe D., \textit{A Security Aware Fuzzy Enhanced Ant Colony Optimization Routing in Mobile Ad hoc Networks}, International Conference on Wireless and Mobile Computing, Networking and Communications, \textbf{2018-October}, (no pages found) (2018).\\[0.1cm]
\textbf{[S545]} Zhang H., Zeng Y., Wang Y., Guo Q., Guo X. and Liu F., \textit{Design of a new intelligent city bus system}, Proceedings of IEEE Asia-Pacific Conference on Image Processing, Electronics and Computers, IPEC 2021, no volume, pp. 509-514 (2021).\\[0.1cm]
\textbf{[S546]} Zhang J., Ye Z. and Li K., \textit{Multi-sensor information fusion detection system for fire robot through back propagation neural network}, PLoS ONE, \textbf{15(7 July)}, (no pages found) (2020).\\[0.1cm]
\textbf{[S547]} Zhang P., Hang Y., Ye X., Guan P., Jiang J., Tan J. and Hu W., \textit{A United CNN-LSTM Algorithm Combining RR Wave Signals to Detect Arrhythmia in the 5G-Enabled Medical Internet of Things}, IEEE Internet of Things Journal, no volume, (no pages found) (2021).\\[0.1cm]
\textbf{[S548]} Zhang Q., Miao J., Zhang Z., Yu F., Fu F. and Wu T., \textit{Energy-Efficient Video Streaming in UAV-Enabled Wireless Networks: A Safe-DQN Approach}, 2020 IEEE Global Communications Conference, GLOBECOM 2020 - Proceedings, \textbf{2020-January}, (no pages found) (2020).\\[0.1cm]
\textbf{[S549]} Zhang T. and Zhu Q., \textit{Distributed Privacy-Preserving Collaborative Intrusion Detection Systems for VANETs}, IEEE Transactions on Signal and Information Processing over Networks, \textbf{4(1)}, pp. 148-161 (2018).\\[0.1cm]
\textbf{[S550]} Zhang X., Peng M., Yan S. and Sun Y., \textit{Deep-Reinforcement-Learning-Based Mode Selection and Resource Allocation for Cellular V2X Communications}, IEEE Internet of Things Journal, \textbf{7(7)}, pp. 6380-6391 (2020).\\[0.1cm]
\textbf{[S551]} Zhang Z., Xia C., Chen S., Yang T. and Chen Z., \textit{Reachability Analysis of Networked Finite State Machine with Communication Losses: A Switched Perspective}, IEEE Journal on Selected Areas in Communications, \textbf{38(5)}, pp. 845-853 (2020).\\[0.1cm]
\textbf{[S552]} Zhang, D., et al., \textit{Software-Defined Vehicular Networks With Trust Management: A Deep Reinforcement Learning Approach}, IEEE Transactions on Intelligent Transportation Systems, p. 1–15 (2020).\\[0.1cm]
\textbf{[S553]} Zhang, M., et al., \textit{Fuzzy Logic-based Resource Allocation Algorithm for V2X Communications in 5G Cellular Networks}, IEEE Journal on Selected Areas in Communications, p. 1–1 (2021).\\[0.1cm]
\textbf{[S554]} Zhang, T., \textit{Toward Automated Vehicle Teleoperation: Vision, Opportunities, and Challenges}, IEEE Internet of Things Journal, vol. 7, no 12, p. 11347–54 (2020).\\[0.1cm]
\textbf{[S555]} Zhang, Y., et al., \textit{Deep Reinforcement Learning-aided Transmission Design for Multi-user V2V Networks}, 2021 IEEE Wireless Communications and Networking Conference (WCNC), IEEE, p. 1–6 (2021).\\[0.1cm]
\textbf{[S556]} Zhao H., Mao T., Duan J., Wang Y. and Zhu H., \textit{FMCNN: A factorization machine combined neural network for driving safety prediction in vehicular communication}, IEEE Access, \textbf{7}, pp. 11698-11706 (2019).\\[0.1cm]
\textbf{[S557]} Zhao H., Zhang J., Li X., Wang Q. and Zhu H., \textit{Deep Learning-based Prediction of Traffic Accident Risk in Vehicular Networks}, 2020 IEEE Globecom Workshops, GC Wkshps 2020 - Proceedings, no volume, (no pages found) (2020).\\[0.1cm]
\textbf{[S558]} Zhao J., Huang S. and Wei Y., \textit{Design and implementation of the Intelligent Patrol Management System based on RFID}, 2011 International Conference on Electrical and Control Engineering, ICECE 2011 - Proceedings, no volume, pp. 3879-3881 (2011).\\[0.1cm]
\textbf{[S559]} Zheyuan C., Rahman M., Tao H., Liu Y., Pengxuan D. and Yaseen Z., \textit{Need for developing a security robot-based risk management for emerging practices in the workplace using the Advanced Human-Robot Collaboration Model}, Work, \textbf{68(3)}, pp. 825-834 (2021).\\[0.1cm]
\textbf{[S560]} Zhong X., Peng X., Yan S., Shen M. and Zhai Y., \textit{Assessment of the feasibility of detecting concrete cracks in images acquired by unmanned aerial vehicles}, Automation in Construction, \textbf{89}, pp. 49-57 (2018).\\[0.1cm]
\textbf{[S561]} Zhou C., Lin Z., Du C., Wang Z. and Li F., \textit{Research on Key Technologies of Tunnel Robot Based on Cloud Edge Collaboration}, Proceedings of 2020 IEEE 2nd International Conference on Civil Aviation Safety and Information Technology, ICCASIT 2020, no volume, pp. 661-666 (2020).\\[0.1cm]
\textbf{[S562]} Zhou H., Wang X., Umehira M., Chen X., Wu C. and Ji Y., \textit{Wireless Access Control in Edge-Aided Disaster Response: A Deep Reinforcement Learning-Based Approach}, IEEE Access, \textbf{9}, pp. 46600-46611 (2021).\\[0.1cm]
\textbf{[S563]} Zhou S., Huang H., Chen W., Zhou P., Zheng Z. and Guo S., \textit{PiRATE: A Blockchain-Based Secure Framework of Distributed Machine Learning in 5G Networks}, IEEE Network, \textbf{34(6)}, pp. 84-91 (2020).\\[0.1cm]
\textbf{[S564]} Zhu L., He Y., Yu F., Ning B., Tang T. and Zhao N., \textit{Communication-based train control system performance optimization using deep reinforcement learning}, IEEE Transactions on Vehicular Technology, \textbf{66(12)}, pp. 10705-10717 (2017).\\[0.1cm]
\textbf{[S565]} Zhu Z., Hou G., Chu Z., Li X., Sun G., Hao W. and Sehdev P., \textit{Research and Analysis of Urllc Technology Based on Artificial Intelligence}, IEEE Communications Standards Magazine, no volume, (no pages found) (2021).\\[0.1cm]
}

\renewcommand\tabularxcolumn[1]{m{#1}}% for vertical centering text in X column

\clearpage
\section{Classification of primary studies}
\label{sec:appendix_B}

\begin{table*}[!b]

\begin{subtable}[h]{\textwidth}
\centering

% \begin{tabularx}{\columnwidth}{>{\vskip-0.2cm\hsize=2.2cm\color{blue}}X|>{\vskip-0.2cm\color{blue}}X}
\begin{tabularx}{\columnwidth}{>{\hsize=2.2cm\color{blue}}X|>{\color{blue}}X}
  \toprule
  \textbf{2G} & S001, S046, S056, S075, S094, S124, S143, S228, S236, S332, S360, S366, S377, S379, S396, S444, S514 \\ \midrule
  \textbf{3G} & S208, S209, S300, S359, S375, S432, S510 \\ \midrule
  \textbf{4G} & S009, S154, S158, S185, S437, S465, S481, S485, S486, S487, S511, S516, S532 \\ \midrule
  \textbf{5G} & S012, S022, S040, S060, S081, S099, S104, S111, S150, S186, S254, S257, S259, S261, S269, S275, S276, S277, S278, S281, S301, S302, S304, S316, S335, S339, S350, S383, S388, S400, S406, S438, S448, S452, S453, S472, S474, S483, S488, S494, S500, S519, S526, S538, S544, S547, S553, S554, S563, S565 \\ \midrule
  \textbf{Cellular \newline(not-specified)} & S018, S021, S027, S029, S033, S042, S048, S050, S059, S071, S074, S076, S077, S086, S090, S091, S097, S101, S105, S117, S119, S147, S157, S166, S170, S171, S179, S201, S202, S204, S211, S223, S235, S237, S247, S248, S256, S258, S265, S268, S272, S273, S274, S285, S296, S303, S317, S319, S328, S344, S354, S372, S374, S391, S395, S402, S409, S416, S418, S428, S431, S440, S454, S464, S467, S469, S470, S482, S501, S504, S506, S513, S524, S546, S548, S550, S552 \\ \bottomrule
\end{tabularx}
\caption{Cellular communication technologies classification.}% \textcolor{red}{Add WIFI, Bluetooth and other non-cellular technologies}} 
\label{tab:cell_class}
\end{subtable}
\vspace{0.2cm}

\begin{subtable}[h]{\textwidth}
\centering
% \begin{tabularx}{\columnwidth}{>{\vskip-0.2cm\hsize=2.2cm\color{blue}}X|>{\vskip-0.2cm\color{blue}}X}
\begin{tabularx}{\columnwidth}{>{\hsize=2.2cm\color{blue}}X|>{\color{blue}}X}
  \toprule
  \textbf{Bluetooh} & S030, S043, S115 \\ \midrule
  \textbf{RFID} & S087, S103, S242, S266, S283, S334, S558 \\ \midrule
  \textbf{Satellite} & S034, S066, S199, S217, S234, S246, S264, S293, S337, S429, S496, S503, S539 \\ \midrule
  \textbf{WiFi} & S007, S011, S013, S023, S024, S025, S035, S044, S045, S051, S058, S063, S064, S068, S070, S073, S088, S089, S092, S096, S098, S100, S110, S112, S116, S121, S122, S125, S126, S128, S132, S135, S138, S144, S152, S174, S177, S182, S192, S218, S221, S224, S243, S244, S249, S260, S290, S305, S307, S309, S311, S313, S315, S325, S329, S338, S340, S345, S364, S369, S371, S380, S382, S389, S392, S394, S414, S420, S421, S427, S441, S442, S447, S451, S456, S458, S459, S461, S466, S475, S476, S480, S493, S517, S522, S527, S545, S560 \\ \midrule
  \textbf{Zigbee} & S384, S460, S492, S525 \\ \midrule
  \textbf{Non-Cellular\newline(not-specified)} & S003, S014, S016, S017, S019, S020, S028, S057, S061, S065, S069, S072, S078, S085, S093, S095, S106, S107, S113, S118, S127, S129, S130, S131, S137, S140, S151, S172, S175, S176, S178, S181, S188, S191, S193, S194, S196, S197, S203, S212, S216, S222, S226, S230, S232, S239, S250, S251, S253, S262, S263, S267, S282, S288, S289, S299, S308, S314, S318, S321, S322, S331, S341, S342, S346, S349, S352, S361, S362, S365, S367, S368, S381, S385, S386, S398, S401, S411, S417, S423, S435, S439, S463, S477, S479, S495, S498, S531, S542, S549, S556 \\ \bottomrule
\end{tabularx}
\caption{Non-cellular communication technologies classification.}% \textcolor{red}{Add WIFI, Bluetooth and other non-cellular technologies}} 
\label{tab:non_cell_class}
\end{subtable}

\vspace{0.2cm}
\begin{subtable}[h]{\textwidth}
\centering
% \begin{tabularx}{\columnwidth}{>{\color{blue}}l|>{\vskip-0.2cm\color{blue}}X}
\begin{tabularx}{\columnwidth}{>{\color{blue}}l|>{\color{blue}}X}
\toprule
\textbf{Not-Mentioned} &
  S002, S004, S005, S006, S008, S010, S015, S026, S031, S032, S036, S037, S038, S039, S041, S047, S049, S052, S053, S054, S055, S062, S067, S079, S080, S082, S083, S084, S102, S108, S109, S114, S120, S123, S133, S134, S136, S139, S142, S145, S146, S148, S149, S153, S155, S156, S159, S160, S161, S162, S163, S164, S165, S167, S168, S169, S173, S180, S183, S184, S187, S189, S190, S195, S198, S200, S205, S206, S207, S210, S213, S214, S215, S219, S220, S225, S227, S229, S231, S233, S238, S240, S241, S245, S252, S255, S270, S271, S279, S280, S284, S286, S287, S291, S292, S294, S295, S297, S298, S306, S310, S312, S320, S323, S324, S326, S327, S330, S333, S336, S343, S347, S348, S351, S353, S355, S356, S357, S358, S363, S370, S373, S376, S378, S387, S390, S393, S397, S399, S403, S404, S405, S407, S408, S410, S412, S413, S415, S419, S422, S424, S425, S426, S430, S433, S434, S436, S443, S445, S446, S449, S450, S455, S457, S462, S468, S471, S473, S478, S484, S489, S490, S491, S497, S499, S502, S505, S507, S508, S509, S512, S515, S518, S520, S521, S523, S528, S529, S530, S533, S534, S535, S536, S537, S540, S541, S543, S551, S555, S557, S559, S561, S562, S564 \\ \bottomrule
\end{tabularx}

\caption{Studies that do not explicitly mention the communication technology used.}% \textcolor{red}{Add WIFI, Bluetooth and other non-cellular technologies}} 
\label{tab:not_mentioned_class}
\end{subtable}

% }
\caption{Primary studies regarding the communication technologies classification.}% \textcolor{red}{Add WIFI, Bluetooth and other non-cellular technologies}} 
\label{tab:comm_class}
\end{table*}

\begin{table*}[!b]
\centering

% \begin{tabularx}{\columnwidth}{>{\vskip-0.2cm\hsize=2.2cm\color{blue}}X|>{\vskip-0.2cm\color{blue}}X}
\begin{tabularx}{\columnwidth}{>{\hsize=2.2cm\color{blue}}X|>{\color{blue}}X}
\toprule
\textbf{Clustering} &
  S009, S011, S025, S045, S057, S059, S071, S073, S078, S088, S089, S100, S116, S127, S133, S135, S137, S152, S161, S164, S169, S172, S185, S188, S189, S193, S224, S244, S258, S262, S310, S325, S335, S340, S348, S371, S383, S386, S401, S411, S427, S442, S463, S469, S480, S485, S486, S487, S498, S513, S516, S517, S529, S537 \\ \midrule
\textbf{Cognitive \newline Architecture} &
  S042, S388, S499 \\ \midrule
\textbf{Computer\newline Vision} &
  S001, S014, S017, S019, S030, S033, S039, S043, S056, S066, S072, S074, S077, S080, S083, S087, S090, S102, S107, S110, S111, S112, S115, S121, S159, S160, S181, S187, S192, S211, S216, S223, S226, S231, S234, S236, S237, S239, S259, S261, S264, S267, S271, S280, S289, S291, S302, S311, S316, S317, S318, S320, S326, S328, S338, S347, S355, S356, S361, S362, S365, S370, S376, S379, S391, S396, S397, S398, S403, S404, S408, S410, S420, S423, S443, S444, S449, S451, S461, S462, S468, S472, S477, S482, S483, S484, S488, S494, S495, S497, S503, S504, S509, S510, S515, S525, S528, S533, S539, S543, S545, S560 \\ \midrule
\textbf{Fuzzy Logic} &
  S010, S013, S022, S028, S037, S051, S063, S065, S068, S151, S176, S179, S194, S207, S215, S221, S288, S321, S322, S330, S381, S382, S415, S438, S440, S456, S466, S489, S493, S527, S544, S553 \\ \midrule
\textbf{Natural \newline Language \newline Processing} &
  S004, S173, S229, S270, S274, S333, S336, S360, S366, S372, S413, S416 \\ \midrule
 \textbf{Neural\newline Networks} &
  S002, S005, S020, S023, S024, S027, S029, S031, S032, S035, S040, S052, S054, S064, S067, S082, S084, S085, S093, S094, S095, S097, S103, S113, S117, S118, S119, S120, S126, S129, S130, S134, S143, S144, S147, S163, S166, S171, S174, S175, S183, S184, S198, S200, S205, S210, S212, S214, S217, S218, S219, S225, S227, S230, S233, S240, S241, S248, S250, S252, S257, S260, S266, S275, S276, S277, S286, S287, S290, S292, S293, S294, S297, S298, S301, S303, S306, S312, S314, S324, S327, S331, S332, S341, S342, S344, S352, S357, S364, S373, S374, S375, S384, S395, S414, S426, S432, S436, S439, S441, S447, S453, S457, S460, S465, S467, S473, S474, S475, S476, S481, S500, S501, S505, S506, S508, S511, S518, S519, S523, S524, S530, S538, S546, S547, S552, S555, S556, S557, S561, S564, S565 \\ \midrule
\textbf{Ontology} &
  S061, S079, S138, S139, S140, S149, S155, S300, S430, S464, S471, S479, S534 \\ \midrule
\textbf{Optimization} &
  S021, S041, S062, S157, S238, S253, S392, S409 \\ \midrule
\textbf{Regression \newline Analysis} &
  S099, S109, S136, S146, S178, S313, S319, S459 \\ \midrule
\textbf{Reinforcement \newline Learning} &
  S026, S098, S101, S106, S108, S154, S182, S204, S247, S279, S282, S284, S295, S296, S305, S343, S351, S353, S367, S369, S390, S394, S399, S406, S412, S419, S422, S437, S502, S512, S521, S540, S541, S548, S550, S562 \\ \midrule
\textbf{State Machine} &
  S044, S081, S086, S158, S197, S209, S255, S531, S551 \\ \midrule
\textbf{Not-specified} &
  S003, S006, S007, S008, S012, S015, S016, S018, S034, S036, S038, S046, S047, S048, S049, S050, S053, S055, S058, S060, S070, S075, S076, S091, S092, S096, S104, S105, S114, S123, S124, S125, S131, S132, S141, S142, S145, S148, S150, S153, S156, S162, S168, S170, S177, S180, S186, S190, S191, S195, S196, S199, S201, S202, S203, S208, S213, S220, S222, S228, S232, S242, S243, S245, S246, S249, S251, S254, S256, S263, S265, S268, S269, S272, S273, S278, S281, S283, S285, S299, S304, S307, S308, S309, S315, S329, S334, S337, S339, S345, S346, S349, S350, S354, S358, S359, S368, S377, S378, S385, S389, S393, S400, S402, S405, S407, S417, S418, S421, S424, S425, S428, S429, S431, S433, S434, S435, S445, S446, S448, S450, S452, S454, S455, S458, S470, S478, S492, S496, S507, S514, S520, S522, S526, S532, S535, S536, S542, S549, S554, S558, S559, S563 \\ \midrule
\textbf{Other} &
  S069, S122, S128, S165, S167, S206, S235, S323, S363, S380, S387, S490, S491 \\ \bottomrule
\end{tabularx}

\caption{Primary studies regarding the AI algorithm classification.}
\label{tab:ai_class}
\end{table*}

\begin{table*}[t]
\centering

% \begin{tabularx}{\columnwidth}{>{\color{blue}}l|>{\vskip-0.2cm\color{blue}}X}
\begin{tabularx}{\columnwidth}{>{\color{blue}}l|>{\color{blue}}X}
\toprule
\textbf{Agriculture} &
  S038, S122\\ \midrule
\textbf{Air Space} &
  S004, S005, S010, S015, S020, S055, S070, S142, S173, S197, S217, S241, S246, S264, S303, S304, S306, S347, S357, S426, S431, S446, S449, S464, S495, S496, S503, S527, S537, S542, S560 \\ \midrule
\textbf{Automotive} &
  S001, S002, S003, S007, S008, S009, S011, S013, S014, S016, S021, S022, S024, S025, S026, S029, S030, S031, S032, S035, S036, S037, S039, S041, S044, S045, S046, S047, S048, S051, S052, S053, S054, S057, S058, S059, S060, S061, S062, S063, S065, S067, S068, S069, S071, S072, S073, S077, S078, S080, S084, S085, S088, S089, S090, S091, S092, S093, S095, S097, S098, S099, S100, S102, S104, S106, S108, S110, S111, S112, S115, S116, S121, S127, S128, S131, S135, S136, S138, S139, S140, S141, S144, S145, S146, S149, S151, S152, S154, S155, S160, S161, S162, S163, S168, S169, S170, S172, S174, S175, S176, S178, S179, S182, S183, S184, S185, S186, S187, S188, S189, S190, S193, S194, S202, S207, S209, S211, S212, S215, S218, S219, S221, S222, S224, S226, S230, S232, S234, S235, S238, S239, S240, S244, S249, S251, S253, S258, S262, S263, S268, S269, S273, S279, S281, S283, S284, S285, S287, S288, S289, S290, S293, S295, S296, S298, S305, S307, S308, S309, S310, S311, S312, S313, S314, S315, S316, S317, S321, S325, S326, S328, S330, S338, S339, S340, S341, S344, S346, S348, S349, S350, S351, S352, S353, S355, S358, S362, S363, S366, S367, S368, S369, S371, S373, S374, S375, S380, S381, S382, S383, S385, S388, S389, S391, S392, S393, S394, S396, S397, S399, S400, S401, S404, S405, S406, S408, S409, S411, S413, S417, S418, S420, S422, S423, S424, S427, S432, S433, S434, S435, S436, S437, S442, S445, S447, S454, S456, S457, S458, S459, S462, S463, S468, S470, S471, S474, S476, S477, S478, S479, S480, S481, S483, S485, S486, S487, S490, S491, S493, S498, S500, S501, S502, S504, S507, S508, S512, S513, S514, S515, S516, S517, S520, S521, S522, S523, S532, S533, S534, S538, S540, S541, S543, S545, S548, S549, S550, S552, S553, S554, S555, S556, S557, S563 \\ \midrule
\textbf{Construction} &
  S083, S109, S113, S129, S192, S233, S415, S443 \\ \midrule
\textbf{Education} &
  S416 \\ \midrule
\textbf{Factory} &
  S006, S017, S019, S421, S482, S526, S551, S558 \\ \midrule
\textbf{Health Care} &
  S033, S040, S043, S049, S056, S079, S094, S103, S107, S124, S153, S159, S196, S201, S210, S225, S236, S255, S266, S271, S272, S274, S300, S318, S322, S323, S333, S334, S345, S360, S372, S376, S384, S430, S439, S444, S455, S473, S492, S499, S531, S535, S547, S562 \\ \midrule
\textbf{Marine} &
  S034, S082, S087, S096, S133, S181, S191, S199, S205, S213, S220, S229, S286, S337, S387, S403, S529, S539 \\ \midrule
\textbf{Mining} &
  S117, S118, S119, S120, S126, S132, S148, S297, S329, S331, S460, S511 \\ \midrule
\textbf{Robotics} &
  S157, S245, S270, S320, S324, S343, S361, S559, S561\\ \midrule
\textbf{Surveillance} &
  S018, S028, S064, S130, S134, S180, S200, S214, S216, S223, S228, S231, S237, S248, S250, S260, S265, S267, S280, S291, S301, S336, S356, S370, S377, S379, S390, S395, S402, S407, S410, S419, S428, S441, S450, S451, S452, S461, S467, S497, S505, S506, S510, S518, S524, S525, S536, S546\\ \midrule
\textbf{Telecommunication} &
  S042, S074, S076, S081, S101, S105, S114, S125, S143, S147, S150, S158, S166, S171, S177, S204, S208, S227, S242, S243, S247, S252, S254, S256, S257, S259, S275, S277, S319, S335, S342, S354, S412, S429, S438, S440, S453, S465, S488, S519, S544, S565 \\ \midrule
\textbf{Other} &
  S012, S023, S027, S050, S066, S075, S086, S123, S137, S156, S164, S165, S167, S195, S198, S203, S206, S261, S276, S278, S282, S292, S294, S299, S302, S327, S332, S359, S364, S365, S378, S386, S398, S414, S425, S448, S466, S469, S472, S475, S484, S489, S494, S509, S528, S530, S564 \\ \bottomrule

\end{tabularx}

\caption{Primary studies regarding the domain classification.}
\label{tab:dom_class}
\end{table*}

\end{appendices}

\end{document}